%% file: ms.tex
\def\withSI{}
\documentclass{article}
\usepackage{arxiv}
\usepackage[T1]{fontenc}
\usepackage{amsmath,amssymb}
\usepackage{graphicx}
\usepackage{booktabs}
\usepackage{enumitem}
\usepackage{xcolor}
\usepackage[numbers,sort&compress]{natbib}
\usepackage[font=small,labelfont=bf,labelsep=period]{caption}
\usepackage{siunitx}
\usepackage[colorlinks=true,linkcolor=blue!50!black,citecolor=blue!50!black,urlcolor=blue!50!black]{hyperref}

\newcommand{\DET}{\textsc{det}}
\newcommand{\REAL}{\textsc{real}}

\renewcommand{\headeright}{Preprint}
\renewcommand{\undertitle}{}
\renewcommand{\shorttitle}{Targeted search and worst-case error in a simulated wave-based neural operator}

\newcommand{\papertitle}{Targeted search shows that random-device testing underestimates worst-case error in a
simulated wave-based neural operator}
\title{\papertitle}
\hypersetup{pdftitle={Targeted search shows that random-device testing underestimates worst-case error in a simulated wave-based neural operator},
  pdfauthor={Samrendra Roy, Jason Yoo, Souvik Chakraborty, Syed Bahauddin Alam}}
\author{%
  Samrendra Roy$^{1}$
  \And
  Jason Yoo$^{1}$
  \And
  Souvik Chakraborty$^{1,2,3}$
  \And
  Syed Bahauddin Alam$^{1,4,}$\thanks{Corresponding author: \texttt{alams@illinois.edu}}
  \AND
  \parbox{\dimexpr\textwidth-2\tabcolsep\relax}{\centering\normalfont\small
  $^{1}$Department of Nuclear, Plasma, and Radiological Engineering,\break
  University of Illinois Urbana-Champaign, Urbana, IL 61801, USA\endgraf
  $^{2}$Department of Applied Mechanics, Indian Institute of Technology Delhi, Hauz Khas, New Delhi 110016, India\endgraf
  $^{3}$Yardi School of Artificial Intelligence (ScAI), Indian Institute of Technology Delhi, Hauz Khas,
  New Delhi 110016, India\endgraf
  $^{4}$National Center for Supercomputing Applications (NCSA),\break
  University of Illinois Urbana-Champaign, Urbana, IL 61801, USA}%
}

\date{\mbox{}}%
\begin{document}
\maketitle

\begin{abstract}
\noindent Wave-based processors promise fast, energy-efficient Fourier layers for neural operators. They are usually
validated on randomly sampled devices, but using them requires knowing how large their error can become under
fabrication and alignment variation. In a stylised numerical case study, a hybrid Fourier neural operator runs its
four spectral layers on simulated coherent 4f processors with 32 toleranced knobs, whose half-widths are
representative rather than calibrated. For 120 models (four tasks, six training methods, five seeds), we compared the
worst of $N$ random in-spec devices with a searched one. On a deterministic simulator with one frozen draw of the
random static errors, the searched device's held-out error was 1.08--3.10 times the maximum over 200 Monte Carlo
devices and 1.06--2.71 times that over 1000. With 20 fresh static draws, it still exceeded the maximum over 200
random devices in 116 of 120 models. Under uniform sampling, the probability of drawing such a device is at most
0.37\% per model (two-sided 95\% Clopper--Pearson), which says nothing about how large its error is. The gap
persisted with uniform or Sobol' sampling at the search's budget, shared knobs, a second crosstalk model, box scales
of 0.25--2 and a pixel-level device model. Models trained only with random static errors reached 3.7--39.9 times
their nominal error on searched devices, and fine-tuning on random and gradient-searched devices gave the lowest
searched error of the six in all 20 task--seed pairs. For two heat-exchanger quantities, a search targeted at each
exceeded the worst of 1000 random devices in all 39 models, and hence the Wilks 95/95 limit (worst of 59). For the
mean pressure of 11 models, no random device exceeded a 1\% error threshold, but the searched device did. Random testing
estimates how often errors exceed a threshold; worst-device search gives a lower bound on how large they can be.
\end{abstract}

\keywords{wave-based computing; physical neural networks; neural operators; worst-case robustness; energy-efficient AI
hardware; nuclear energy systems}

\section{Introduction}\label{sec:introduction}
The energy cost of deep learning and of large-scale scientific computing has renewed interest in wave-based
processors. These compute with light or sound as it propagates, and they promise high throughput at low energy
per operation~\citep{wetzstein2020inference,mcmahon2023physics,momeni2025training}. Optical versions range from
coherent nanophotonic meshes and diffractive networks to free-space Fourier-optical systems and large-scale
optoelectronic processors~\citep{shen2017deep,lin2018all,chang2018hybrid,hamerly2019large,zhou2021large}.
Physical neural networks have also been trained on mechanical and acoustic systems~\citep{wright2022deep}. Neural
operators learn solution maps of partial differential equations
(PDEs)~\citep{li2021fourier,kovachki2023neural,lu2021deeponet}, and they suit this hardware well. The Fourier
neural operator (FNO) is built on a spectral convolution, and a 4f optical system computes such a convolution (input
plane, lens, Fourier plane, lens and output plane, each one focal length apart). A lens Fourier-transforms the field,
a mask weights the spatial frequencies and a second lens transforms back~\citep{goodman2005fourier}. Such Fourier-plane masks have served as trained layers of optical neural
networks~\citep{miscuglio2020massively,yan2019fourier}, and a recent optical neural engine realised this spectral
convolution in free-space optics for Darcy, Navier--Stokes and Maxwell problems~\citep{tang2025one}. Diffractive and
integrated photonic PDE solvers have followed~\citep{jing2026diodes,yuan2025microcomb}.

Energy and nuclear engineering are exploring fast surrogates of this kind, from foundation models for dynamic nuclear
systems~\citep{roy2025ansfoundation} and reconstruction of global radiation fields from sparse sensor
sequences~\citep{kobayashi2025proxies} to plant monitoring~\citep{roy2026neuromorphicmonitoring} and reactor
control~\citep{lee2025agentic}. Monitoring, control, design studies and digital twins all call for fast, repeated
inference. The energy cost of neural-operator inference has been measured for virtual
sensing~\citep{yoo2026virtualsensing}, and neuromorphic hardware has been explored for nuclear PDE
solving~\citep{roy2026scno}. Wave-based processors are a further candidate, but science and engineering will adopt
them only if their error under real fabrication and alignment variation can be bounded.

Physical devices differ from their designs. Spatial light modulators (SLMs) suffer pixel crosstalk (fringing
fields)~\citep{apter2004fringing,persson2012reducing} and have a finite extinction ratio. Fourier-plane masks are
registered only to within a fraction of a pixel, lenses carry residual aberrations, and fabrication adds random per-pixel
errors. In integrated photonic meshes, small component errors accumulate across the circuit, and the architecture
and post-fabrication correction set how much accuracy they cost~\citep{fang2019design,bandyopadhyay2021hardware}. A
common response is hardware-aware or noise-aware training, the hardware counterpart of domain
randomisation~\citep{tobin2017domain}. The model sees randomly perturbed devices during
training~\citep{joshi2020accurate,rasch2023hardware,stutz2021bit,wright2022deep,mengu2020misalignment,xu2026sat}
and is then validated on a sample of random devices or in Monte Carlo fault trials~\citep{xu2026faultfree}. Random
(Monte Carlo) testing estimates the behaviour of a typical device and, with confidence bounds, the probability that a
device exceeds an error threshold~\citep{clopper1934use}.
Probabilistic guarantees from random samples, such as the scenario
approach~\citep{calafiore2006scenario,campi2008exact}, bound the fraction of failing devices. They do not bound
the error of the worst one. Robust optimisation, design centring and worst-case circuit
analysis~\citep{bental2009robust,director1977simplicial,antreich1994circuit} read a tolerance specification
differently, as a set of admissible devices, each of which must meet the requirement.

Machine learning draws the same line. Statistical methods estimate the probability that a random input perturbation
makes a network fail~\citep{webb2019statistical}, whereas adversarial robustness asks for the worst admissible
perturbation~\citep{szegedy2014intriguing,goodfellow2015explaining,madry2018towards}. Gradient-free differential
evolution (DE)~\citep{storn1997differential} has been used to find admissible input perturbations that break
neural networks~\citep{su2019one}. It has recently been applied to neural-operator surrogates of nuclear
thermal-hydraulics, where sparse perturbations raised the error from about 1.5\% to
37--63\%~\citep{roy2026adversarial}. We adopt that DE-based search over an admissible set and do not claim it as
new. That work perturbs the inputs of digital surrogates, whereas here the inputs are clean and the perturbation is
the hardware itself: a device drawn from the tolerance box of the wave-based layers. The closest precedent is
compute-in-memory accelerators, where the worst-case weight perturbation within the device-variation bounds proved
far more damaging than average-case evaluation suggests~\citep{yan2022iccad,yan2023atrice}. Certified
bounds exist for weight~\citep{weng2020weight} and input perturbations~\citep{zhang2018crown,gowal2018ibp}. We search
a different space, the low-dimensional, physically defined tolerance box of a wave-based neural operator, and compare
random testing with search within it, including at matched budgets.

We make this comparison in simulation. All hardware is simulated with a differentiable scalar Fourier-optics model,
and no device was built or measured. The model gives each of the 576 ($24\times24$) input--output channel paths per layer its own
programmable complex mask and does not specify how these paths would be multiplexed. The work is therefore a stylised
numerical case study. We do not assess the tolerances of a realisable processor, and we measure neither energy
nor speed. Because a targeted search is expected to find at least as high an error as random sampling, the questions
are how much higher and whether the difference matters. For this simulator and tolerance box, we draw four
conclusions.
\begin{enumerate}[label=(\roman*),leftmargin=2.2em,topsep=2pt,itemsep=1pt,parsep=0pt]
\item \emph{Search finds in-spec devices whose errors lie well above the worst found by random testing.} In all 120
models on four tasks, the searched device's held-out error exceeded that of each of 1000 random devices, by up to
about three times the worst of 200, with one frozen draw of the random static errors; with fresh draws the gap held
in nearly all models (Section~\ref{sec:endpoint}). Search only bounds the worst case from below, so random testing
underestimates it by at least this factor. The gap remained for random sampling at the search's
budget, on 100,000 uniform devices and in every tolerance set, crosstalk model and device model tried
(Sections~\ref{sec:v3} and~\ref{sec:tol}).
\item \emph{For models trained on random box devices, a short gradient search from a jittered box centre is a cheap,
strong check, given a differentiable device model.} For these box-trained Darcy, Navier--Stokes (NS) and
heat-exchanger (HX) models, 30 steps of projected signed-gradient ascent (PGD) through the differentiable simulator
were the strongest single method (Section~\ref{sec:endpoint}). Random box corners, even at the search's budget, came
within 10\% of the searched error in only 8 of 24 models (Section~\ref{sec:v3}).
\item \emph{Random testing can miss threshold exceedances that search finds.} For two HX quantities, a search
targeted at each exceeded the largest error among 1000 random devices in all 39 HX models, and hence also the Wilks
95/95 limit of nuclear safety analysis (the largest error among 59). For the slice-mean pressure of more than a
quarter of these models, no random device had an error above 1\% of the mean, yet the searched device did
(Section~\ref{sec:hx}).
\item \emph{Fine-tuning on a mixture of random and searched devices gave the lowest searched error.} It was the
best of the six training methods for every task and seed, even against attacks unseen in training
(Section~\ref{sec:arms}).
\end{enumerate}
These results suggest qualifying wave-based hardware by pairing random testing with worst-device search
(Section~\ref{sec:discussion}).

\section{Results}\label{sec:results}
\subsection{A hybrid wave-based neural operator with a 32-dimensional tolerance box}\label{sec:setup}
\begin{figure}[tb]
\centering
\includegraphics[width=\linewidth]{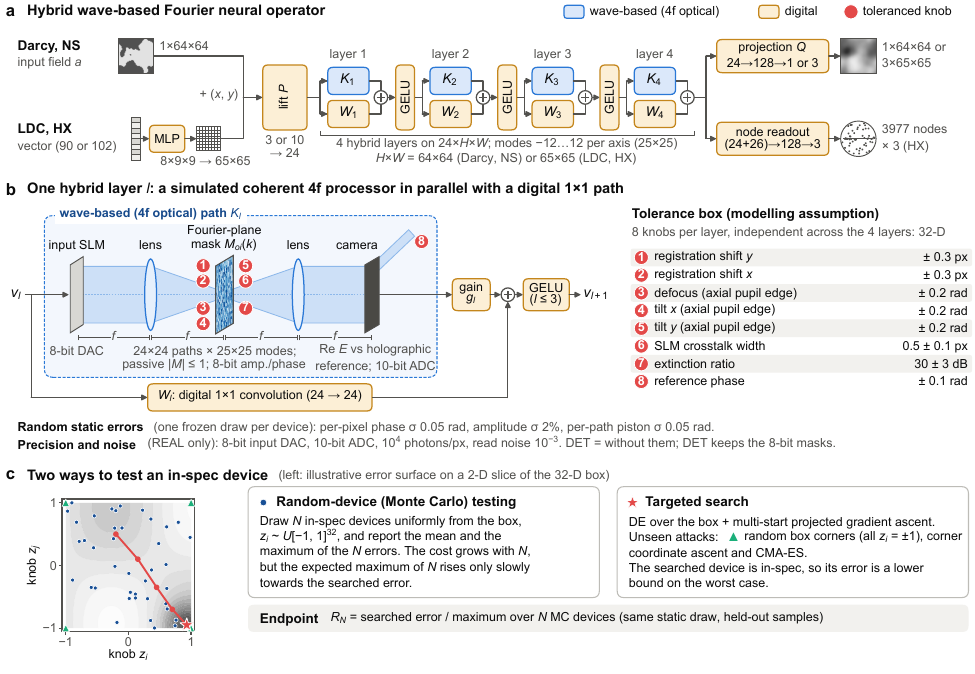}
\caption{\textbf{Simulated hybrid wave-based neural operator, tolerance box and testing approaches.}
\textbf{a},~Model pipeline with tensor shapes (channels\,$\times$\,height\,$\times$\,width). Blue marks the
wave-based (4f optical) spectral paths $K_l$ and amber the digital parts. Darcy and Navier--Stokes (NS) take an
input field. The inputs of the lid-driven cavity (LDC; 90-dimensional lid-history encoding) and the heat exchanger
(HX; 102-dimensional condition vector) pass through a digital multilayer perceptron (MLP) to a coarse latent grid, which is upsampled bilinearly. Two grid coordinates are appended before the
lift, and GELU follows the first three hybrid layers. HX reads its output at the 3977 mesh nodes with a digital
readout. \textbf{b},~One hybrid layer: a simulated coherent 4f processor in parallel with a digital $1\times1$
convolution, with an idealised phase-sensitive camera readout against a holographic reference beam. Red circles mark the eight
toleranced knobs per layer where they act; they are independent across layers in the primary 32-D
box. The table gives the representative half-widths. The two lines at the bottom of the panel list the random
static errors, present in both configurations, and the converters and detection noise, which the deterministic
configuration (\DET) omits. \textbf{c},~Random testing samples Monte
Carlo (MC) devices uniformly from the box, whereas targeted search seeks a high-error device, whose error is a lower
bound on the worst case. Random corners belong to the unseen attacks of the search, not to random testing. The
error surface is illustrative.}
\label{fig:schematic}
\end{figure}

We simulate a four-layer hybrid FNO with 24 channels and Fourier modes $-12,\dots,12$ along each axis
(Fig.~\ref{fig:schematic}a). Each layer adds a wave-based spectral path to a digital $1\times1$ convolution, and a
Gaussian error linear unit (GELU) nonlinearity follows the first three layers. The spectral path is a simulated coherent 4f processor
(Fig.~\ref{fig:schematic}b). An input SLM encodes the 24 channels, a lens forms the spectrum and a finite pupil
keeps $25\times25$ modes. Every input--output channel path is weighted by a passive complex mask with 8-bit
amplitude ($\le1$) and 8-bit phase. A second lens transforms back to real space, where a camera reads the real part
of the field against a reference beam, with local-oscillator shot noise. Lifting, the pointwise ($1\times1$) path,
the nonlinearities and the output projection are digital, as in hybrid optical--electronic designs~\citep{chang2018hybrid,tang2025one}.

A device is one realisation of the wave-based layers: a setting of the systematic knobs (below) and one frozen draw
of random static errors (per-pixel phase and amplitude errors and a per-path piston). Each layer has eight
systematic knobs, each known only to within a tolerance. Five knobs describe the optical alignment:
the Fourier-plane registration shift in $y$ and in $x$ ($\pm0.3$ pixel), the defocus and the tilt in $x$ and in
$y$ (each $\pm0.2$\,rad at the axial pupil edge). Three describe the modulator and the readout: the SLM crosstalk
width ($0.5\pm0.1$ pixel), the extinction ratio ($30\pm3$\,dB) and the reference phase ($\pm0.1$\,rad). Over four
layers the knobs span a 32-dimensional (32-D) box, and a device is in-spec when all of its knobs lie inside it.
Because we model crosstalk as a blur of the complex mask field $\alpha\,e^{i\varphi}$, the blurred mask does not
change when the stored phase wraps by $2\pi$ (Methods). The half-widths are representative values, neither a
manufacturer specification nor measurements of one platform, so Section~\ref{sec:tol} also scales the box and ties
the knobs across layers.

We use four tasks. The Darcy-flow pressure and the NS vorticity fields are defined on $64\times64$ grids, and the
lid-driven cavity (LDC) fields~\citep{kobayashi2025conformalized} on a $65\times65$ grid. The outputs of the HX
task~\citep{roy2026adversarial} are defined on an unstructured mesh. On each task we train six arms (training
methods) with five seeds each (Methods): Random-error-aware, Box, Smooth-mask box and three fine-tunes (FT) of each
Box checkpoint, namely Box FT, Mixed-PGD FT (random and gradient-searched devices) and SAM FT (sharpness-aware
minimisation, SAM). We call the five arms trained on the box (all but Random-error-aware) box-trained. Box and
Smooth-mask box name arms; otherwise lower-case box means the tolerance set. A digital FNO of the same size gives an
accuracy reference.

We use two simulator configurations. The deterministic configuration, \DET, keeps the 8-bit masks and the static
errors but removes the input and output converters and the detection noise; the realistic configuration, \REAL,
includes them. We test every model on \DET\ with one static draw, $s_0$, which all testing methods share, so the
comparison concerns the systematic knobs only. We compare two kinds of test. Monte Carlo (MC) testing draws 1000 devices $z_i$ uniformly from the box, where $z$
denotes the normalised knob coordinates. Search uses two attack families (Fig.~\ref{fig:schematic}c). The DE/PGD
family combines gradient-free DE (three seeds)~\citep{storn1997differential}, which follows the admissible-set search
of ref.~\citep{roy2026adversarial}, with PGD through the differentiable simulator from eight starts. The second
family, the unseen attacks, is never used in training: 200 random box corners, coordinate ascent over corners (CA)
and four runs of the covariance matrix adaptation evolution strategy (CMA-ES)~\citep{hansen2001cmaes}. Because a
searched device only bounds the worst case from below, we report the budget and strength of every method
(Supplementary Note~2).

Each test set is split in two. Every search selects its device on the first 64 samples (the development set), and
errors are reported on the remaining samples (the held-out set) unless stated otherwise. The searched device,
$z_{\text{search}}$, is the device with the highest development error over all searches. The primary endpoint
\begin{equation}
R_N=\frac{e_{\DET,\text{hold}}(z_{\text{search}},s_0)}{\max_{i=1,\dots,N}e_{\DET,\text{hold}}(z_i,s_0)},
\qquad z_i\sim U[-1,1]^{32},
\label{eq:RN}
\end{equation}
divides its held-out error by the largest held-out error among the first $N$ of the 1000 MC devices, where $e$ is
the mean relative $L_2$ error over the held-out samples. We report $R_{200}$ and $R_{1000}$.

The comparison favours MC, because the MC maximum is taken directly on the held-out samples, whereas the searched
device must be chosen on the development samples. For the same reason, the highest-error device of a single attack
family occasionally scored higher on the held-out samples than $z_{\text{search}}$ (3 of 120 models; Supplementary
Note~2 and Tables~S1 and S2). We fixed the endpoint and protocol before the main experimental round (Supplementary
Note~1). A secondary endpoint uses the \REAL\ configuration with independent static draws.

\subsection{Search finds in-spec devices that 1000 random devices miss}\label{sec:endpoint}
\begin{figure}[tb]
\centering
\includegraphics[width=\linewidth]{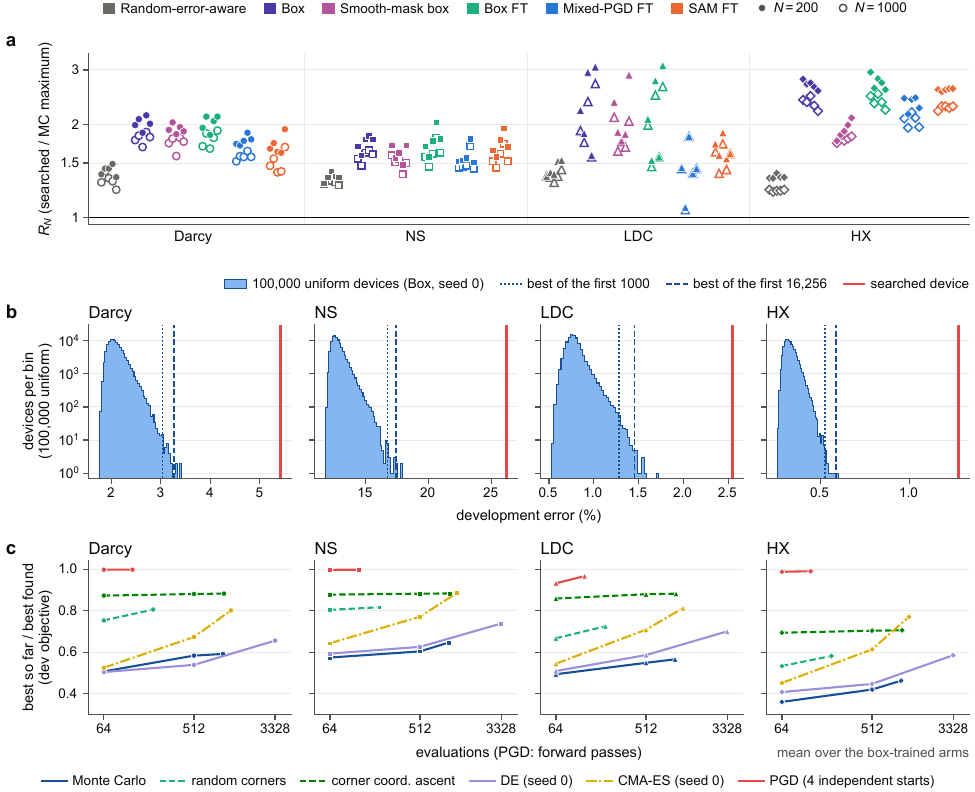}
\caption{\textbf{Search finds in-spec devices that random testing misses.}
\textbf{a},~Primary endpoint $R_N$ (equation~\eqref{eq:RN}) for all 120 models on \DET\ and held-out samples, for
$N=200$ (filled) and $N=1000$ (open), five seeds per arm. All values exceed 1. LDC errors are effectively pressure
errors (Methods).
\textbf{b},~Development errors of 100,000 uniformly drawn devices for the Box seed-0 model of each task
(histogram, logarithmic counts). The dotted and dashed lines mark the highest error among the first 1000
devices and among the first 16,256, the budget of the search, and the red line marks the searched device. No device reached it
(Section~\ref{sec:v3}).
\textbf{c},~Highest development error reached so far, divided by the highest found over all methods, after 64, 512 and
all evaluations, mean over the five box-trained arms. The curves show budget progress, not convergence. DE and
CMA-ES (seed 0) are counted in forward evaluations, and the PGD curve covers only its four independent starts (124
forward passes, most with a backward pass; Methods and Supplementary Note~2).}
\label{fig:endpoint}
\end{figure}

In every one of the 120 models, search found an in-spec device with a larger held-out error than any of 1000 random
devices. The searched error was 1.08--3.10 times the worst of 200 MC devices ($R_{200}$; median 1.79;
Fig.~\ref{fig:endpoint}a; Supplementary Fig.~S1) and 1.06--2.71 times the worst of 1000 ($R_{1000}$; median 1.63).
By task, $R_{200}$ spanned 1.35--2.14 (Darcy), 1.29--2.03 (NS), 1.08--3.10 (LDC, whose
errors are effectively pressure errors; Methods and Supplementary Table~S16) and 1.34--2.95 (HX). The searched LDC
pressure errors were typically dominated by a uniform offset, which carries no physics in a closed cavity. With the
spatial mean of the pressure removed, $R_{200}$ was 1.07--2.15, above 1 in all 30 LDC models (Supplementary Note~10
and Table~S19); the original metric stays primary because the endpoint was fixed in advance. The
smallest margin was for LDC Mixed-PGD FT seed~1, where the searched device reached 0.541\%, compared with 0.500\% for
the worst of 200 MC devices ($R_{200}=1.08$, $R_{1000}=1.06$). The unseen attacks alone also exceeded the MC maximum
in every model, at 1.06--2.89 times the maximum of 200 and 1.04--2.37 times the maximum of 1000 (Supplementary
Tables~S1 and S2). The result therefore does not depend on PGD, the gradient attack that Mixed-PGD FT uses in
training.

The secondary endpoint, on \REAL, asks whether the gap survives the random static errors. We re-scored the searched
knob setting on 20 fresh static draws and compared its mean error with the maximum over 200 \REAL\ MC devices, each
with its own static draw. This ratio was 0.67--2.70 and was at least 1 in 116 of 120 models. The four exceptions are
LDC Mixed-PGD FT models (0.67 to just below 1). Their \DET\ margins were not unusually small (Supplementary
Table~S2). Their searched knob setting, however, had been found with static draw $s_0$ only, and on \REAL\ its error
varied across the 20 fresh draws about as much as the knobs had raised it above the nominal error (Supplementary
Table~S4). For these models, the static errors mattered as much as the systematic knobs. A nested search, which
maximised the mean error over eight common static draws and was re-scored on 20 fresh draws, reached 0.77--2.71
times the same \REAL\ MC maximum and fell below it in the same four models (Supplementary Tables~S3 and S4).

No MC device reached the searched error. For each model, the probability that a uniformly drawn device does so is
therefore at most 0.37\% (upper end of the two-sided 95\% Clopper--Pearson interval~\citep{clopper1934use}), and a
one-sided 95\% bound that holds simultaneously for all 120 models is 0.78\% (Supplementary Fig.~S2; Methods gives
the conditions). Among the 100 box-trained models, between 0 and 3 of 1000 devices exceeded
twice the nominal error, and none did in 97 models; under uniform sampling such errors were rare. The searched
device, however, exceeded twice the nominal error in 72 of the 100. Its error was 1.08--3.06 times the
order-statistic 95\% upper confidence bound on the 0.99 quantile of the MC error (Supplementary Tables~S3 and S4).

Random testing closed the gap slowly; gradient search closed it fastest. For the box-trained arms (means over seeds),
the worst of 1000 MC devices
reached only 0.42--0.71 of the highest error found on the held-out samples (Supplementary Fig.~S3). On the
development samples, its running maximum grew little between 64 and 1000 evaluations (Fig.~\ref{fig:endpoint}c).
CMA-ES reached 0.73--0.90 of the highest error after 1200 evaluations (development samples;
Fig.~\ref{fig:endpoint}c) and was still rising, whereas CA stalled early. PGD from the jittered box centre, 30
signed-gradient steps, was the strongest single method for the box-trained Darcy, NS and HX models (0.97--1.00 of
the highest error), and the highest-error device came from DE or PGD in 96 of the 100 box-trained models. For the
Random-error-aware models, CA, one of the unseen attacks, found it in 13 of 20 (Supplementary Fig.~S3, Tables~S1
and S2).

\subsection{The gap persists at a matched budget and on a pixel-level device model}\label{sec:v3}
Uniform and Sobol' sampling at the search's own budget did not close the gap, nor did 100,000 uniform devices, and
the gap remained on a pixel-level device model. The search spends 16,256 development evaluations per model, compared with 1000 for MC. We gave this budget to four
samplers (uniform, scrambled Sobol'~\citep{owen1998scrambling}, random box corners and the mixed sampler, which alternates corners with
uniform devices; Methods) on the 24 seed-0 models. Each selects its device by development error, and $R_{\rm match}$ divides the searched held-out error by
that device's held-out error. The searched error exceeded that of the uniform device in all 24 models
($R_{\rm match}$ 1.21--2.26) and that of the Sobol' device in all 24 (1.19--2.16; Supplementary Table~S5 and
Fig.~S4). Random corners came closer ($R_{\rm match}$ 1.00--1.90) and were within 10\% of the searched error
($R_{\rm match}\le1.1$) in 8 of 24 models, including all four Random-error-aware ones (1.00--1.03). For the HX Box model and its three fine-tunes, however, the
searched error was still 1.52--1.90 times the corner device's. The mixed sampler behaved like the corner sampler
(within 0.06; Supplementary Note~3).

For the Box seed-0 model of each task, none of 100,000 uniform devices reached the searched development error, and
the upper end of the two-sided 95\% Clopper--Pearson interval for the probability of reaching it fell to
0.0037\%. The highest of these errors reached 0.47--0.68 of the searched one (Fig.~\ref{fig:endpoint}b), and on the held-out samples the searched error was
1.45--2.12 times that of the device selected from 100,000. Generalised
Pareto tail fits are consistent with such errors being extremely rare or unreachable under uniform sampling, but for
NS and LDC
the tail fitted above the 0.99 quantile ends below an error that the search reaches. Tail extrapolation is therefore unreliable here, and
we rely on the empirical result (Supplementary Note~3 and Table~S6).

In the device model used so far, which we call mode-level, one SLM pixel is one Fourier mode. Crosstalk and
registration therefore blur neighbouring modes directly, and the crosstalk and extinction tolerances are two-sided,
with the nominal value at the box centre. The pixel-level device (Supplementary Note~3) is more detailed. Each mode spans
$4\times4$ SLM pixels, and fringing (pixel-level crosstalk) acts between pixels. The registration shift is passive
and given by the exact cell integral of the displaced pixels. The fringing (0.5--0.6 pixel) and extinction-ratio
(27--30\,dB) tolerances are one-sided, with the nominal device at the sharp, high-extinction end. We retrained the
Random-error-aware, Box and Mixed-PGD FT arms on this device for Darcy and HX (seeds 0--2) and repeated the full
protocol.

The gap remained on the pixel-level device (Supplementary Table~S7): no MC device reached the searched error, and
$R_{200}$ was 1.59--2.42 on HX and 1.33--1.65 on Darcy. At the matched budget, $R_{\rm match}$ was 1.27--1.73 for
uniform sampling and 1.01--1.44 for the corner sampler. With a band-limited (sinc) shift, the gap also remained on Darcy ($R_{200}$ 1.25--1.69). Of the three retrained
arms, Mixed-PGD FT had the lowest searched error in all six task--seed pairs, as it had on the mode-level device
(Section~\ref{sec:arms} below). Random-error-aware models degraded less than on the mode-level device
(Section~\ref{sec:nominal} below): their searched error was 2.30--2.89 times the nominal error on Darcy and 9.85--11.2
times on HX. Models trained on the mode-level device
and evaluated on the pixel-level device (transferred models) lost accuracy (nominal error of Darcy Box from 2.46\% to
5.97\%, of HX Box from 0.353\% to 1.17\%), and their searched error still exceeded the MC maximum ($R_{200}$
1.18--1.49).

\subsection{Nominal error hides degradation, and the worst device need not be the most degraded}\label{sec:nominal}
A low nominal error hid strong degradation of the Random-error-aware models on searched devices, and the
highest-error devices were often not the most physically degraded ones. These models form a hardware-aware baseline
that sees only random static errors, with the knobs at the box centre. On the searched devices their errors
were 3.7--39.9 times the nominal errors: 5.5--9.2 times on Darcy, 3.7--5.9 on NS, 17.5--39.9 on LDC and 18.7--20.8
on HX (Table~\ref{tab:arms}; Supplementary Tables~S1 and S2). Random testing already exposes much of this
degradation: on LDC the MC mean was 9.1--13.6\%, compared with a nominal 0.70--1.64\%. Search therefore adds less
for these models than for the box-trained ones, although $R_{200}$ was still 1.29--1.53.

The searched devices give clues to how the box-trained models fail. For the Box and Mixed-PGD FT models of seed~0 on
Darcy, NS and HX, every layer of the searched device had the sharpest crosstalk and the lowest extinction ratio in
the box. Their registration shifts were within 0.05 of zero in normalised coordinates for all 16 shift coordinates on
Darcy and NS and for 14 of 16 on HX (Fig.~\ref{fig:gap}e). Because PGD
starts from a jittered centre, where the shift gradient does not vanish, and takes signed steps of fixed size (Methods;
Supplementary Note~1), these near-zero shifts are unlikely to be an artefact of the start point: the highest-error
device found was sharp and well aligned. In the mode-level model, however, bilinear resampling also blurs the mask
at any non-zero shift, which couples sharpness and alignment. The models may have co-adapted to
the imperfections they were trained with, but the coordinates were optimised jointly and no intervention isolates
sharpness from alignment.

On the pixel-level device the pattern held for all six HX Box and Mixed-PGD FT models and for all four transferred
models (every layer at the sharpest fringing and the lowest extinction ratio, with 6--8 of 8 shift coordinates
within 0.05 of zero). On Darcy it held for only one seed of three per arm, for none of the nine models with the sinc
shift and never for Random-error-aware models (Supplementary Table~S7).

On the mode-level device, the box centre had a higher error than the
average random device for all 25 box-trained models on NS (ratio of nominal error to MC mean 1.02--1.15;
Supplementary Tables~S1 and S2) and on HX (1.06--1.16), and for 21 of 25 on Darcy and 13 of 25 on LDC. On the pixel-level device, whose nominal setting is the sharp, high-extinction end of the box,
the nominal device was worse than the average random device for all Box models with the cell-integral shift
(1.51--1.58 on HX, 1.03--1.08 on Darcy). The most physically degraded device is therefore not necessarily the one
with the highest error.

\subsection{The gap depends on the tolerance set but persists}\label{sec:tol}
\begin{figure}[tb]
\centering
\includegraphics[width=\linewidth]{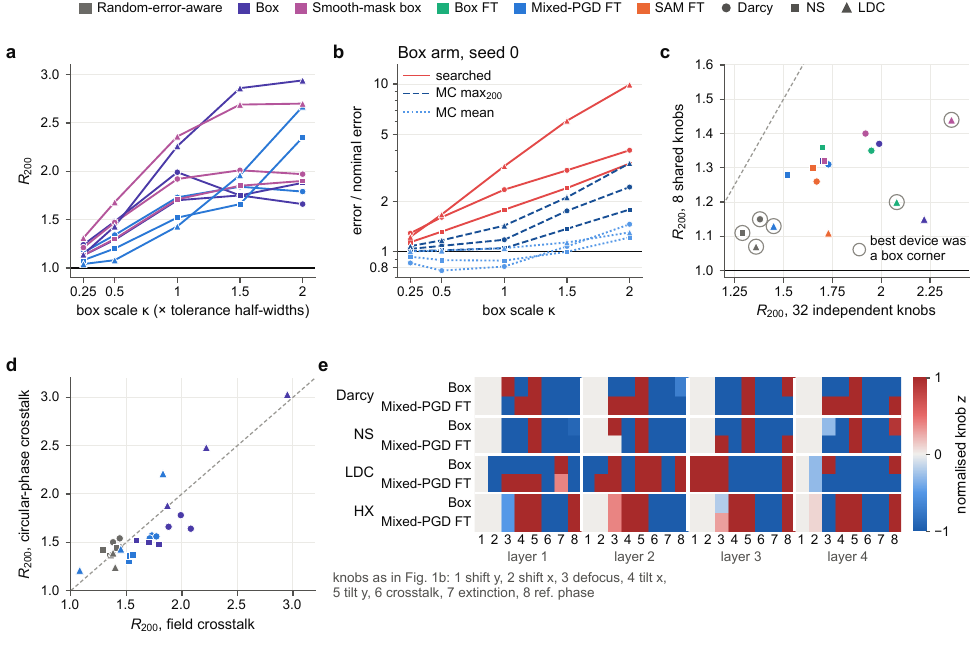}
\caption{\textbf{The gap under other tolerance sets and crosstalk models, and the searched devices.}
\textbf{a},~$R_{200}$ against the box scale $\kappa$ (all half-widths scaled) for the seed-0 Box, Smooth-mask box and Mixed-PGD FT models on Darcy, NS and LDC
(Supplementary Table~S9).
\textbf{b},~Box arm only, same sweep: searched error, MC maximum of 200 and MC mean, each divided by the nominal
error. \textbf{c},~$R_{200}$ in the shared-knob 8-D box (the same eight knobs in all layers) against the 32-D box
for the same seed-0 models (Supplementary Table~S8). Colour gives the arm and marker shape the task (legend at the
top). The dashed line marks equality, and grey rings mark models whose highest-error device was a box corner. \textbf{d},~$R_{200}$ with
circular-phase crosstalk against field crosstalk (seeds 0--2; Supplementary Table~S10). The circular-phase models were retrained, so each
point compares complete training-and-simulator configurations rather than isolating the crosstalk law.
\textbf{e},~Searched devices $z_{\text{search}}$ (Box and Mixed-PGD FT, seed 0) in normalised coordinates, where,
for knobs 6 and 7, $z=-1$ is the sharpest crosstalk and the lowest extinction ratio. Each row is one model.}
\label{fig:gap}
\end{figure}

The size of the gap depended on the assumed tolerances, but the searched error remained above the MC maximum in
every tolerance set and crosstalk model we tried (Fig.~\ref{fig:gap}a--c; Supplementary Note~4 and Tables~S8 and S9). Scaling all half-widths by $\kappa$ for
the Box, Smooth-mask box and Mixed-PGD FT models of seed~0 raised $R_{200}$ from 1.04--1.31 at $\kappa=0.25$ to 1.43--2.36 at
$\kappa=1$. On NS and LDC it kept growing up to $\kappa=2$, whereas on Darcy it peaked at $\kappa=1$--1.5; at $\kappa=2$ it was
1.66--2.94 over all nine sweeps. From $\kappa=0.25$ to 2 the searched error rose 2.6--8.1-fold, while the
MC mean rose only 1.04--1.75-fold. At $\kappa=0.5$ the MC mean was even below the nominal error in 8 of 9 sweeps. Judged by
the mean alone, looser tolerances would look harmless while the searched error grows.

Errors in different layers may be correlated. Tying corresponding knobs across the four layers gives an 8-D box,
whose 256 corners we enumerated. In this 8-D box, $R_{200}$ was smaller, 1.07--1.44
(Fig.~\ref{fig:gap}c), and a box corner was the highest-error device in 6 of 18 models, including
all three Random-error-aware ones. On Darcy and NS the searched 8-D device reached 0.98--1.02 times the 32-D
searched error. This is expected, because the knob settings of the 32-D searched devices were already similar
across layers. The smaller
$R_{200}$ there comes from a higher MC maximum (1.14--1.48 times its 32-D value). On LDC the 8-D
searched error was 0.55--0.86 times the 32-D value. Tying the knobs also changes the admissible set and its
correlations, so this experiment does not isolate the effect of dimension.

Correlation can also be partial: layers built with the same SLM type share its crosstalk and extinction ratio but
not their alignment. In a correlated 26-D box that shares only these two knobs (Methods), $R_{200}$ was 1.16--2.51 for the 24 seed-0
models, and its median per task lay between those of the same models in the 8-D box (not run for HX) and the 32-D
box (Supplementary Note~9 and Table~S17). In 19 of the 20 box-trained models, the searched device had the sharpest
crosstalk (0.4 pixel) and the lowest extinction ratio (27\,dB); the Random-error-aware ones, and the one exception, took the opposite corner.

We also retrained the Random-error-aware, Box and Mixed-PGD FT arms (seeds 0--2) on Darcy, NS and LDC under a second,
circular-phase crosstalk model (Methods). The gap persisted: in all
27 models $R_{200}$ was 1.21--3.03 (Fig.~\ref{fig:gap}d; Supplementary Table~S10), and Mixed-PGD FT gave a lower
searched error than Box in 9 of 9 paired seeds.

\subsection{Mixed-PGD fine-tuning gives the lowest searched error among the tested methods}\label{sec:arms}
\begin{figure}[tb]
\centering
\includegraphics[width=\linewidth]{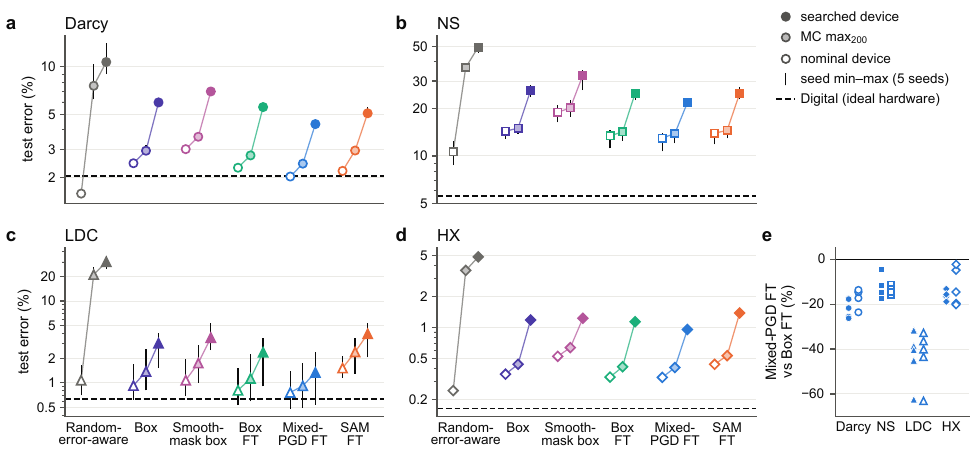}
\caption{\textbf{Training arms on four tasks.} \textbf{a--d},~Nominal device (open), maximum over 200 MC devices
(light) and searched device (filled) for each arm, all on \DET\ and held-out samples. Each symbol is the mean over
five seeds, with a vertical line from the seed minimum to the seed maximum. The dashed line is the digital FNO on
ideal hardware (three seeds), and the $y$-axes are logarithmic. Supplementary Tables~S1 and S2 list all models.
\textbf{e},~Change of the searched (filled) and unseen-attack (open) error of Mixed-PGD FT relative to Box FT with
the same seed. Supplementary Table~S11 lists the paired values.}
\label{fig:arms}
\end{figure}

By searched error, fine-tuning on a mixture of random and searched devices was the best of the six training methods
in all 20 task--seed pairs. We compared five ways of training on the box (Methods). Two arms train from scratch on
random box devices: Box (100 epochs) and Smooth-mask box, whose Fourier masks are bilinear interpolants of a coarse
$7\times7$ grid. Three arms fine-tune each Box checkpoint for 30 epochs. Box FT continues Box training. On half of
the batches, Mixed-PGD FT trains on a device found by two signed-gradient steps instead of a random box device, a
form of adversarial training over the device knobs. SAM FT applies SAM over the
weights~\citep{foret2021sharpness,xu2026sat}. The arms are not compute-matched (Methods). Because fine-tunes of the same seed share their
initial checkpoint and data order, we compare them paired by seed (Fig.~\ref{fig:arms}, Table~\ref{tab:arms}).

Mixed-PGD FT also gave the lowest error under the unseen attacks, which it never saw in training, again in 20 of 20.
Its searched error was $4.35\pm0.19$\% on
Darcy, $21.9\pm0.5$\% on NS, $1.34\pm0.70$\% on LDC and $0.952\pm0.066$\% on HX (Table~\ref{tab:arms}). Paired by seed, it was 22\%,
13\%, 44\% and 16\% below Box FT on average on the four tasks (Supplementary Table~S11). With five seeds per task, a
5-of-5 paired result has a two-sided sign-test $p$-value of 0.0625 per task. We report the 20-of-20 result descriptively, because no combined test
was prespecified. Mixed-PGD FT also had the lowest mean nominal error among the box-trained arms on all four tasks.
On Darcy its nominal error (2.03\%) matched the digital FNO (2.04\%), and on LDC and HX it was 1.20 and 2.01 times
the digital error.

Smooth masks did not reduce the mean searched error relative to Box. The Smooth-mask box arm's searched error was 13--22\%
above Box for every seed on Darcy and 4--49\% above on NS, while on LDC and HX individual seeds were mixed (Supplementary
Tables~S1 and S2). Its nominal error was
higher than Box's in all seeds except two on LDC. Smooth masks lowered $R_{200}$ on HX (1.77--2.10, compared with
2.57--2.80 for Box), but they did so by raising the random-device errors.

On NS, the hybrid hardware and box training cost nominal accuracy. The box-trained arms had mean nominal
errors of 13.0--19.0\%, against 10.6\% for Random-error-aware and 5.55\% for the digital FNO. The width (24 channels)
was not tuned.

\subsection{On a heat-exchanger mesh, search finds exceedances that random testing misses}\label{sec:hx}
The heat exchanger had the largest median gap of the four tasks, and for two engineering quantities a targeted search
exceeded the worst of 1000 random devices in every model. The HX task, a generic thermal-hydraulic surrogate rather
than a model of a nuclear component, extends the comparison to outputs on an unstructured mesh. A 102-dimensional condition vector maps to the
pressure and the two in-plane velocity components at the 3977 nodes of a tube cross-section (Methods).
Interpolating the true fields to a $65\times65$ grid and back already incurs a 5.3\% relative $L_2$ error, so the
model samples the last layer's features at the nodes and applies a per-node digital multilayer perceptron (MLP) with
a coordinate encoding (Supplementary Note~5). The metric is the mean over the three channels of the per-channel
relative $L_2$ error. Without this averaging, the pressure, of order $10^3$--$10^4$\,Pa, would dominate a joint
error.

The digital FNO reached 0.162\%. Random-error-aware models reached 0.237--0.258\% at the nominal device but
4.6--4.9\% on the searched device. Among the box-trained arms, $R_{200}$ was 2.57--2.80 for
Box and 2.16--2.44 for Mixed-PGD FT. Averaged over seeds, pressure had the largest
channel error for every arm and device, and it did so for every model--device pair but one (Supplementary Note~5 and
Table~S12).

\begin{figure}[tb]
\centering
\includegraphics[width=\linewidth]{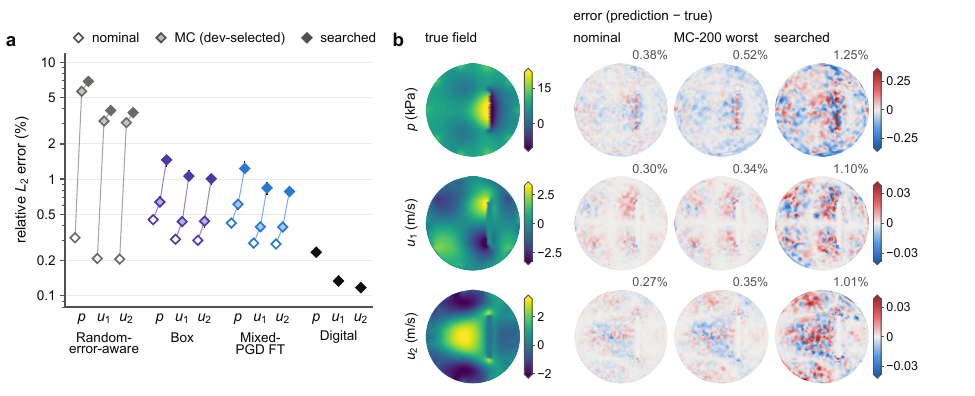}
\caption{\textbf{Heat exchanger (3977-node mesh).} \textbf{a},~Per-channel relative $L_2$ error (pressure $p$,
in-plane $z$- and $y$-velocities $u_1$ and $u_2$) on \DET\ for three arms (Random-error-aware, Box and Mixed-PGD
FT). Devices: nominal, the MC device selected by development
error from all 1000 (not the held-out MC maximum used in $R_N$) and the searched device. Values are means over five seeds, with vertical lines from the minimum to the maximum over seeds. Black marks the digital FNO (three seeds), and the $y$-axis is
logarithmic. Supplementary Table~S12 lists the per-channel and joint errors. \textbf{b},~Box seed~0, held-out
sample with the median nominal error: true fields and the errors (prediction minus true field) on the nominal
device, on the worst of the first 200 MC devices (selected by development error) and on the searched device. The
error maps of a channel share a symmetric colour scale, clipped at the 99.5th percentile of the absolute error, and
true fields span their 0.5th to 99.5th percentiles. Error colour bars use the row label's units. Each map gives that channel's relative
$L_2$ error (primary metric 0.31\%, 0.40\% and 1.12\%).}
\label{fig:hx}
\end{figure}

For the held-out sample with the median nominal error of the Box seed-0 model, the searched device raised the error
in every channel across the cross-section (Fig.~\ref{fig:hx}b; Supplementary Figs~S6 and S7 show other models and
tasks).

Engineers act on quantities derived from these fields. On the HX slice we computed the area-weighted mean gauge
pressure $Q_p$ (6.19\,kPa on average) and the root-mean-square in-plane (secondary-flow) speed $Q_s$
(1.63\,m\,s$^{-1}$). A larger field error need not mean a larger error in a given quantity: the field-searched
device had stayed below the MC maximum of $Q_p$ in 6 of the 30 main-round models (Supplementary Note~9, Table~S18
and Fig.~S13). We
therefore searched for the in-spec device that maximises the error of each quantity (Methods). For both
quantities, its error exceeded the largest over all 1000 MC devices in all 30 main-round and all nine pixel-level
models (Supplementary Note~10 and Table~S20). It also exceeded the error of the field-searched device in all but 1
of 78 model--quantity pairs.

In nuclear safety analysis, the largest of 59 random code runs is a widely used one-sided 95/95 tolerance
limit~\citep{wilks1941determination,glaeser2008grs} (Discussion). Here this limit, $W_1$, is the largest error among
the first 59 MC devices, so a search that exceeds the maximum of all 1000 necessarily exceeds it. The informative
comparison is with an error threshold; we
used illustrative thresholds of 1, 2 and 5\% of the mean quantity, set before this search, and the counts depend on
them. At 1\% of the mean $Q_p$, 11 box-trained models passed the Wilks test and none of their 1000 MC devices
exceeded the threshold, yet their searched devices did; the same held for $Q_s$ at 2\% in one pixel-level model.
Here random testing gave no warning. In the other such splits, it did: at 5\% of the mean $Q_p$, the five
Random-error-aware main-round models passed the Wilks test ($W_1$ 4.51--4.92\%) and their searched devices reached
6.42--6.87\%, but 0.1--0.8\% of their MC devices also exceeded 5\%, as did 0.4--2.6\% at 2\% for the three
pixel-level ones. There the Wilks test worked as designed, bounding the 95th percentile with the
worst case above it (Supplementary Note~10, Table~S20 and Fig.~S14).

\begin{table}[tb]
\centering
\caption{\textbf{Training arms} (\DET, held-out samples, relative $L_2$ error in \%; HX: mean of the per-channel
errors; LDC: joint error over $p$, $k$ and $|\mathbf{u}|$, dominated by $p$; see Methods). Values are means over five seeds (digital FNO: three seeds). Unseen-attack best: error of the highest-error
device found by corners, CA or CMA-ES. Searched:
mean $\pm$ s.d.\ over seeds. $R_{200}$: range over seeds. $\Delta$: change of the searched error relative to Box FT
with the same seed, mean [min, max]. Digital FNO: ideal hardware.}
\label{tab:arms}
\scriptsize
\setlength{\tabcolsep}{4pt}
\input{tables/tab_arms.tex}
\end{table}

\subsection{Certification does not reach the full box}\label{sec:cert}
A Taylor-model bound on the worst case was tight only on very small boxes and far too loose at the full box. We
computed this enclosure of the worst case over the device knobs~\citep{makino2003taylor,moore2009interval}
for two Darcy models, on ten test inputs each (Supplementary Note~6); it is sound only under exact arithmetic. The
bound was 1.01--1.09 times the highest error found
for box scales up to $\kappa=0.005$ and 2.03--2.46 times at $\kappa=0.01$. At the full box it was
$6\times10^8$--$8.5\times10^8$ times (Supplementary Table~S13), useless for qualifying a device at these tolerances. The
experiment does not separate genuine nonlinearity of the device-to-output map from conservatism in the enclosure.

\section{Discussion}\label{sec:discussion}
A search is expected to find higher errors than random sampling. The contribution here is to measure the gap across
120 models, four tasks and several tolerance sets, to show that uniform or Sobol' sampling with the search's own
budget did not close it in the 24 models tested and to work out what it implies for training and for engineering
quantities. A fabricated device is one fixed point in the box, and where it sits is unknown rather than random. For
such interval uncertainty, a tolerance specification read as a guarantee has to cover the worst admissible device,
whose error search bounds from below.

For the simulated model and box studied here, we would therefore qualify hardware with both: random testing to
estimate how often a device exceeds an error threshold, and worst-device search to give a lower bound on the error
of the worst in-spec device. The search need not be expensive. Random box corners need no gradients and make a
simple first screen; for the Random-error-aware models they came close to the searched error. Corners alone are not
enough. The box has $2^{32}\approx4.3\times10^9$ corners, and even at the search's budget the searched error
exceeded the corner sampler's by more than 10\% for 16 of the 20 box-trained seed-0 models (Section~\ref{sec:v3}).
Several searched devices also lay off the corners, with registration shifts near zero (Fig.~\ref{fig:gap}e). For
box-trained models, a short gradient search from the jittered centre is a cheap first check. Tests should cover the favourable ends of the tolerances as well as the degraded
ones, because the highest-error devices found were often not the most degraded ones.

The distinction matters most where a surrogate's output feeds an engineering decision. A thermal-hydraulic surrogate
in a plant monitoring system or a digital twin is verified over its intended operating range. On wave-based
hardware, the device tolerances become part of what must be verified, which calls for a bound on the error of the
worst in-spec device as well as a failure rate. Search supplies an empirical lower bound on that error, whereas no
rigorous upper bound is yet available at the full box (Section~\ref{sec:cert}).
Nuclear safety analysis makes the same distinction. Best-estimate-plus-uncertainty analyses
propagate input uncertainties through a system code by random sampling and use Wilks' order statistics to set the
number of runs. The largest of 59 runs is then a one-sided upper tolerance limit that covers the 95th percentile of
the output with 95\% confidence~\citep{wilks1941determination,glaeser2008grs,nutt2004evaluation}. Such a 95/95
statement and a worst-device search answer different questions: whether the 95th percentile of the error lies below
a criterion, and whether every in-spec device does. On the HX task, the informative cases were those in which no
random device crossed an error threshold that the searched device crossed (Section~\ref{sec:hx}). If a surrogate on wave-based hardware entered such
an analysis, its device tolerances would be one more uncertain input, and a 95/95 statement over them would leave the
error of the worst in-spec device unknown.

Circuit and optical design have long made this distinction. Monte Carlo yield analysis estimates or bounds a failure
probability, as our Clopper--Pearson bound does. Importance sampling and statistical blockade make rare failures cheaper to
estimate~\citep{kanj2006mixture,singhee2009statistical}, the latter with extreme-value tail fits like ours. Corner and
worst-case-distance analysis seek the worst parameters within the
tolerances~\citep{nassif2001modeling,antreich1994circuit}, and optical tolerancing predicts performance loss from
sensitivities~\citep{youngworth2000cost}. Robust topology optimisation treats under- and over-etching of micro- and
nanostructures as a worst-case design problem~\citep{sigmund2009manufacturing,wang2011projection}. We borrow this
worst-case view for wave-based neural-operator hardware. In such hardware the error landscape is learned and differs from model to model, and a
local linear estimate of the searched error proved unreliable (0.27--1.79 times the searched error; Supplementary
Fig.~S8). On the training side, Mixed-PGD fine-tuning does for the device knobs what adversarial weight
perturbation does for the weights~\citep{wu2020awp}, and its mixture of random and searched devices lies between
average- and worst-case training~\citep{robey2022probabilistic}. Uniform sampling has similar limits on the data
side, where targeted sampling of training inputs makes surrogates more
reliable~\citep{roy2026beyonduniform,roy2026ansreliability}.

A simple heuristic suggests why the gap appears with many knobs and closes slowly with more samples. Suppose the
error were additive in the $D$ knobs and approximately Gaussian under uniform sampling. With equal contributions from
all knobs, the excess of the worst device over the mean error would then be about $\sqrt{3D/(2\ln N)}$ times the
expected excess of the maximum of $N$ uniform devices; unequal contributions give a smaller factor. For $D=32$ the
factor is 3.01 at $N=200$ and 2.64 at $N=1000$ (Supplementary Note~2). It grows with the number of knobs but falls
only as $1/\sqrt{\ln N}$, so more samples help little. Several observed excess ratios exceed it (2.0--5.4 for the
box-trained arms at $N=1000$). Because the factor rests on a leading-order approximation and each observed maximum is
a single realisation, we offer the heuristic only as intuition. Tying the knobs across layers ($D=8$) shrank the gap,
as the heuristic suggests, although tying also changed the admissible set.

Extreme-value theory describes how sample maxima grow with $N$~\citep{coles2001introduction}, and the upper tail can be
extrapolated from extreme order statistics~\citep{pickands1975statistical}, as the CLEVER score does for the
robustness of neural networks~\citep{weng2018clever}. Such an extrapolation, like our tail
fits to 100,000 devices, still describes the sampling law and does not bound the supremum over the box. A certified
bound would, but full-box certification remains open: the enclosure was tight on small boxes but vacuous on the full
box. Interval dependence, norm propagation through globally mixing spectral layers and conservative remainders
can all inflate such bounds, and we cannot say which dominates or whether tighter methods or architectural changes
would give useful full-box bounds.

\paragraph{Limitations.} Hardware for optical neural operators is still at an early stage, with few
demonstrations~\citep{tang2025one,jing2026diodes,yuan2025microcomb}, and whether these conclusions transfer to a
calibrated physical platform is untested. The optics are modelled as scalar, paraxial and coherent, with idealised
lenses. The model omits finite wavelength, focal length and pixel pitch, partial coherence, polarisation, SLM phase--amplitude
coupling, fill factor, zero-order leakage and aliasing. It also omits lens errors beyond defocus and tilt,
Fourier-plane magnification and rotation errors, vignetting, drift and detector non-idealities such as saturation,
fixed-pattern noise and dark current. We do not specify the multiplexing architecture that would realise the 576
separately programmable channel paths per layer, nor the errors it would add, such as fan-out and fan-in loss and
phase drift between paths. The output
analogue-to-digital converter (ADC) of the \REAL\ configuration uses an idealised per-sample automatic gain that
knows the exact output range of every sample. A real detector chain would have to obtain this range separately. The
gain makes 10-bit quantisation benign, and the primary \DET\ endpoint omits the ADC altogether. Both crosstalk
models are invariant to $2\pi$ phase wraps, whereas fringing on liquid-crystal SLMs acts on the wrapped phase and
distorts it near each wrap (phase flyback)~\citep{apter2004fringing,persson2012reducing}; a calibrated
voltage-to-transmission model~\citep{moser2019model} may change the error surface. The gap persisted under the
circular-phase crosstalk model and on the pixel-level device, but that device still averages the field over each
mode's cell instead of modelling the full coupling between modes, and we tested it on two tasks with three seeds.

The half-widths are not calibrated to a platform, and uniform sampling of the box is a
reference sampling law that does not model fabrication. The reliability bounds hold only under that law (Methods). Real
tolerances are partly correlated across layers, and we tested no correlation structure other than the shared-knob
and correlated boxes (Section~\ref{sec:tol}). The budget-matched and 100,000-device runs used seed-0 models only. The searched errors
are lower bounds with a finite budget (Fig.~\ref{fig:endpoint}c; Supplementary Note~2), and DE and CMA-ES were
still improving at the end. All models were tested at one static draw on \DET, and a static seed is not the same
physical draw on different GPUs (Supplementary Note~1). Absolute NS errors are high (Section~\ref{sec:arms}). For
LDC and HX the optical layers act on a latent grid produced by a digital MLP, not on a physical input field. Relative $L_2$ error may hide local PDE failures and, for LDC, the $k$ and $|\mathbf{u}|$ errors (Methods);
averaging the HX quantity errors over operating conditions may hide exceedances at single conditions.

\paragraph{Outlook.} Worst-device search could move from a final check into the design loop, with the tolerance
specification a design variable, traded against fabrication cost and co-designed with the model, as Mixed-PGD
fine-tuning begins to do. A shared operator library serving many energy-system tasks would let one qualification of
a wave-based accelerator cover all of them (Supplementary Note~8 reports an exploratory frozen-library pilot). The
protocol needs only a bounded set of physical knobs and a way to evaluate the error, so it applies to other
wave-based or analogue processors, including acoustic ones. On a calibrated testbed, measured device spread would set
the tolerance box, and the same search could run on a simulator fitted to that device. A free-space 4f system with
SLMs is a natural first platform: registration, defocus, tilt and reference phase can be programmed on the
modulators, whereas crosstalk and extinction would have to be selected across devices or emulated, so a searched
worst device could be set up and measured on the bench. Metasurface or integrated-photonic Fourier layers cannot be
retuned after fabrication, so finding their worst device beforehand matters more. In situ
training, with the forward pass on hardware and gradients from the fitted simulator, would test whether Mixed-PGD
fine-tuning keeps its advantage on a real device, and measuring energy per inference at a fixed accuracy target
against measured digital baselines would tie robustness to cost.

\section{Methods}\label{sec:methods}
\subsection{Simulator}
We use \texttt{wavesim}, a differentiable PyTorch~\citep{paszke2019pytorch} simulator written for this project.
Supplementary Note~1 records its version history, configuration hashes and unit tests. An optical spectral layer encodes each input channel on an SLM with a per-sample,
per-channel scale, an 8-bit magnitude and the sign as a 0/$\pi$ phase. It then takes a centred, orthonormal 2-D fast Fourier transform
(FFT; first lens) and keeps the $(2m+1)\times(2m+1)$ central modes ($m=12$, the finite pupil). Each input--output
channel pair has its own complex mask $M_{oi}(k)$, where $k$ is the Fourier coordinate, and the masked spectra are
summed coherently over inputs. The result is placed back in the spectrum and inverse-transformed (second lens).
Detection takes the real part of the output field $E$ against a reference beam. It adds local-oscillator-limited
shot noise of standard deviation $\eta_{\rm LO}/\sqrt{N_{\rm ph}}$ of full scale ($N_{\rm ph}=10^4$ photons per
pixel, $\eta_{\rm LO}=1/\sqrt2$) and read noise of $10^{-3}$ of full scale. A 10-bit ADC with per-sample automatic gain
and a learned digital gain follow. The automatic gain sets the ADC full scale to the maximum $|\mathrm{Re}\,E|$ of
each sample and output channel, an idealisation discussed under Limitations.

The mask amplitude $\alpha$ is the sigmoid of a learned logit ($\le1$, passive). It is quantised to 8 bits and then mapped onto
$[\alpha_{\min},1]$, with the extinction floor $\alpha_{\min}=10^{-\varepsilon/20}$ and $\varepsilon$ the extinction
ratio in dB, through $\alpha\mapsto\alpha_{\min}+(1-\alpha_{\min})[\alpha]_8$, where $[\cdot]_8$ denotes 8-bit
rounding. This map is continuous in $\varepsilon$. The phase is
also quantised to 8 bits. Quantisation uses a straight-through gradient (rounding forward, identity backward). The input SLM and the output ADC use the same
rule. Static errors are i.i.d.\ Gaussian per pixel (phase $\sigma=0.05$\,rad, relative amplitude $\sigma=0.02$) and
per path (piston $\sigma=0.05$\,rad; Supplementary Note~7 reports a pilot search over the per-pixel errors). The
amplitude error multiplies the amplitude by $1+0.02\zeta$, with $\zeta$ standard
normal. It is not clipped, so a pixel near full transmission can slightly exceed $|M|=1$.

Crosstalk is a Gaussian blur of standard deviation $\gamma$ pixels, applied to each channel path's $25\times25$
mask. The one-dimensional kernel $\exp(-\delta^2/(2\gamma^2))$ is sampled on the integer offsets $|\delta|\le r$,
with $r=\max(1,\lceil 3\gamma\rceil)$ ($r=2$ throughout the box), and normalised to unit sum. It is applied separably along $x$ and then $y$
after replicate padding by $r$ pixels. The field model blurs the real and imaginary parts of $\alpha\,e^{i\varphi}$. The
circular-phase model blurs the amplitude and takes the argument of the blurred $e^{i\varphi}$. Both are invariant to
$2\pi$ wraps of the stored phase. Per mask, quantisation and the extinction floor come
first, then the static errors and crosstalk, and last the defocus, tilt, per-path piston and reference-phase terms
and the registration shift.

Defocus and tilt add $d\,|k|^2$ and $t\cdot k$ phases over the pupil, with $k$ normalised to $[-1,1]$ along each
axis. Defocus $d$ therefore gives a phase of $d$ at the axial pupil edge and $2d$ at the corners. The reference phase
adds a global piston per layer. Registration shifts the mask by bilinear resampling with zero padding. The mask is
zero outside the programmed $25\times25$ aperture, meaning that no light passes there, so a shift attenuates the
edge modes. This edge loss is part of the registration model. With all non-idealities off, the layer reproduces an
FNO spectral convolution.

\subsection{Tolerance box and devices}
For each layer $l$, $\theta_l\in\mathbb{R}^8$ collects the registration shift ($y$, $x$), defocus, tilt ($x$,
$y$), crosstalk width, extinction ratio and reference phase. The box has centre $\theta^0$ and half-widths $h$
(Fig.~\ref{fig:schematic}b), and $\theta=\theta^0+\kappa\,z\odot h$, with $z\in[-1,1]^{32}$, $\odot$ the
elementwise product and $\kappa=1$ unless stated otherwise. The shared-knob variant uses the same $z\in[-1,1]^8$ in every
layer. The correlated box shares only the crosstalk and extinction coordinates across layers, $z\in[-1,1]^{26}$. A device is $(\theta,s)$, where the seed $s$ indexes one frozen static-error draw. \DET\ and \REAL\ differ only in the input
digital-to-analogue converter (DAC), the output ADC and the detection noise (Section~\ref{sec:setup}).

\subsection{Tasks and data}
\emph{Darcy.} The equation is $-\nabla\cdot(a\nabla u)=1$ on a $64\times64$ grid with zero boundary values, and
$a\in\{3,12\}$ comes from a thresholded Gaussian random field (FNO benchmark convention~\citep{li2021fourier}). We
generated 1000 training and 200 test samples independently for this work (\texttt{wavesim.data.make\_darcy}).
\emph{Navier--Stokes.} The map is vorticity $\omega_0\mapsto\omega(T=2)$ on the periodic $64\times64$ torus, with
viscosity $10^{-3}$ and the forcing of the FNO benchmark~\citep{li2021fourier}. The initial vorticity is three times the FNO benchmark amplitude, so that
advection matters. Time stepping is pseudo-spectral Crank--Nicolson/Heun with $\Delta t=10^{-3}$, and there are 1000
training and 200 test samples.

\emph{Lid-driven cavity.} The dataset is from ref.~\citep{kobayashi2025conformalized}, and we use its partition,
with one boundary-condition realisation as the splitting unit: 3949 training, 493 validation and 495 test samples. A
90-dimensional encoding of the lid-velocity history (six interpolation points at 15 snapshots) maps to the pressure
$p$, the turbulent kinetic energy $k$ and the velocity magnitude $|\mathbf{u}|$ of the source's Reynolds-averaged
simulations, in that order, on a $65\times65$ grid. The LDC error is the relative $L_2$ error over the three fields
jointly. The pressure is of order $10^2$--$10^3$ in the source's units, whereas $k$ is below 1 and $|\mathbf{u}|$ at
most about 10, so the pressure dominates this error: on the held-out set the joint error equals the pressure error to
within 0.001 percentage points (Supplementary Table~S16). All models, including the digital FNO ($k$:
13.2\nobreakdash--14.2\%; $|\mathbf{u}|$: 7.7\nobreakdash--8.6\% over its three seeds), predict $k$ and
$|\mathbf{u}|$ far less accurately, so the LDC results concern the pressure. The searched device also raised the $k$
and $|\mathbf{u}|$ errors above those of the MC device with the largest joint error in 26 and 24 of 30 models.
Section~\ref{sec:endpoint} and Supplementary Note~10 also report results with the spatial mean of the pressure
removed from each predicted and true field.

\emph{Heat exchanger.} The source campaign, generated by our group for a companion
study~\citep{roy2026adversarial}, is a Reynolds-averaged Navier--Stokes simulation of a heat-exchanger tube with a
zero-thickness wavy tape insert. It comprises 1546 cases with inlet velocity sampled uniformly on
$[4,5]$\,m\,s$^{-1}$, inlet temperature on $[263,323]$\,K and an imposed wall heat-flux profile with peak values up to
about 22\,kW\,m$^{-2}$. A 102-dimensional
condition vector (inlet velocity, inlet temperature and a 100-point wall heat-flux profile) maps to pressure and two
in-plane velocities on 3977 nodes of a 2-D unstructured mesh of the tube cross-section. The split used here is 1200
training, 173 validation and 173 test cases. The channels are, in order, gauge pressure (Pa) and the in-plane $z$-
and $y$-velocities (m\,s$^{-1}$) on the $(y,z)$ cross-section at $x=0.79$\,m. The axial velocity normal to the slice is not
used. Node coordinates are scaled by the tube radius.

\emph{Splits and input conditioning.} The validation splits were not used for training, model selection or the
search. Every test set is split into the first 64 samples (development) and the rest (held-out: 136 for Darcy and
NS, 431 for LDC, 109 for HX). The participation ratio $(\sum_i\lambda_i)^2/\sum_i\lambda_i^2$
of the covariance eigenvalues $\lambda_i$ of the normalised training inputs is 20.8 for Darcy (90\% of the variance
in 187 components) and 11.3 for NS (19 components). It is 2.84 for LDC (5 components) and 1.04 for HX, where one
component holds 90\% of the variance, so the HX inputs form almost a one-parameter family.

\subsection{Models}
\texttt{HybridFNO2d} lifts the input and two grid coordinates linearly to 24 channels. Four layers follow,
$v\leftarrow\sigma_l(K_lv+W_lv)$, where $\sigma_l$ is a GELU~\citep{hendrycks2016gelu} for $l\le3$ and the
identity for $l=4$. Here $K_l$ is the optical spectral convolution ($24\times24$ masks over $25\times25$ modes,
Fourier indices $-12,\dots,12$ per axis) and $W_l$ a digital $1\times1$ convolution. A projection $24\to128\to1$
(Darcy, NS) or $24\to128\to3$ (LDC), with a GELU after the first map, gives the output. For LDC, a digital MLP
($90\to256\to256\to8\times9\times9$) produces a latent field that is bilinearly upsampled to $65\times65$. For HX,
an MLP of the same form ($102\to256\to256\to8\times9\times9$) feeds a $65\times65$ latent grid over $[-1,1]^2$. The
last layer's 24 feature channels are sampled bilinearly at the 3977 node coordinates and concatenated with a
26-dimensional coordinate encoding (the two scaled in-plane coordinates $c_1$, $c_2$ and $\sin$, $\cos$ of
$\pi jc_1$ and $\pi jc_2$ for $j=1,2,4,\dots,32$).
A per-node digital MLP ($50\to128\to3$) maps them to the outputs (Supplementary Fig.~S9). Training-split statistics
normalise the inputs and targets. The loss is the relative $L_2$ error of the normalised outputs, and the HX metric
is the mean of the per-channel relative $L_2$ errors in physical units. The digital FNO uses ideal hardware with the
same width, modes and layers.

\subsection{Training arms}
All runs use AdamW~\citep{loshchilov2019adamw} (weight decay $10^{-4}$) with a one-cycle
schedule~\citep{smith2019superconvergence}, batch size 20 and seeds 0--4. Scratch runs train for 100 epochs at
learning rate $2\times10^{-3}$. Every hardware arm trains on the \REAL\ configuration, with converters and detection noise.
\emph{Random-error-aware} training uses fresh static errors on every forward pass with the knobs at the box centre. \emph{Box} training draws
a new $\theta\sim U(\text{box})$ and a new static draw for every batch. \emph{Smooth-mask box} is Box training with the
amplitude logit and phase on a $7\times7$ grid per channel pair, bilinearly upsampled to the $25\times25$ pupil.
Fine-tunes start from the Box checkpoint of the same seed and run for 30 epochs at learning rate $10^{-3}$, seeded
with the run seed${}+10$. \emph{Box FT} continues Box training. \emph{Mixed-PGD FT} uses, on a random half of the batches, a
device found by two signed-gradient steps of 0.5 in normalised coordinates ($0.5h$) from a uniform random start with
the static draw frozen. It uses a random box device otherwise. Each searched batch therefore needs two extra
forward--backward passes. \emph{SAM FT} applies SAM to the
weights~\citep{foret2021sharpness} ($\rho=0.05$, not tuned) under random box devices, with the same device for
both SAM passes, so each batch needs one extra forward--backward pass.

Each run uses three random-number streams: the global PyTorch generator (run seed; initialisation and Mixed-PGD
FT's PGD start points), a private generator (seed${}+200$; data order) and a NumPy generator (seed${}+100$; box
devices and static-draw seeds). Arms with the same seed therefore share the data order, scratch arms other than
Smooth-mask box share the initialisation, and Box FT and SAM FT draw the same devices. Mixed-PGD FT consumes extra
device-stream draws (searched-or-random choices and static draws), so its devices are not aligned with theirs.

\subsection{Evaluation and search}
All searches maximise the mean error over the 64 development samples on \DET\ with static draw $s_0=0$, and
`best' always means highest error.
\emph{MC}: 1000 devices $z_i\sim U[-1,1]^{32}$ from a fixed generator. \emph{Corners}: 200 random corners.
\emph{CA}: starting from the best corner, coordinate ascent flips one sign at a time and, after a sweep without
gain, restarts from a random corner (1024 evaluations). \emph{DE}: SciPy \texttt{differential\_evolution}~\citep{virtanen2020scipy}, with
population 8 per dimension, 12 generations and three seeds (3328 evaluations each). \emph{PGD}: 30 signed-gradient
steps (0.25 for 15 steps, then 0.08), clipped to the box, from eight starts. Four starts are seeded by other
methods: the best DE optimum, the CA optimum, the best corner and the best MC device. The other four are
independent. One is the box centre jittered by $U[-0.05,0.05]$, because the shift gradient vanishes at zero
shift, and three are random points. \emph{CMA-ES}: $(\mu/\mu_w,\lambda)$ with $\lambda=16$, $\sigma_0=0.5$
and reflection into the box. Three seeds start from the jittered centre and one from the CA optimum (1200
evaluations each).

Each method's device is selected by its development error, and $z_{\text{search}}$ is the best of these. The
unseen attacks (corners, CA, CMA-ES) share no code path with the training attack. The local linear estimate is
$e_0+\sum_j\max(g_j^+,-g_j^-,0)$ on the full test set. Here $g_j^{\pm}=\pm[e(\pm0.05\,\hat u_j)-e_0]/0.05$ are
one-sided difference quotients along the $j$-th unit vector $\hat u_j$. \emph{Secondary endpoint}: on \REAL, we report the nominal error over 20 static
draws, 200 MC devices with independent static draws and $z_{\text{search}}$ re-scored on 20 static draws. A nested
search $\max_z\text{mean}_s$ runs PGD over eight common static draws and is re-scored on 20 fresh draws. The
shared-knob box (all 256 corners) and the correlated box (26-D; 14,400 development evaluations; no \REAL\ endpoint
or nested search) use the same protocol. The tolerance sweep uses a reduced protocol (Supplementary Note~4).

\emph{HX quantities.} Each node receives a third of the area of its triangles in a Delaunay triangulation of the
scaled nodes (triangles failing either filter of Supplementary Note~9 would be dropped; none were), and the weights
$w_n$ sum to 1. Then $Q_p=\sum_nw_np_n$ and $Q_s=(\sum_nw_n|\mathbf{u}_n|^2)^{1/2}$ with $\mathbf{u}=(u_1,u_2)$
(Supplementary Note~9 also uses the 99th percentile of $|\mathbf{u}|$). $Q_p$ is the pressure drop to the outlet only
if the outlet is at gauge zero, which we assume. A quantity's error is $|Q_{\rm pred}-Q_{\rm true}|$ divided by the
test-set mean of $|Q_{\rm true}|$, averaged over the held-out samples. The quantity-targeted search maximises the
development mean of this error for $Q_p$ or $Q_s$ with a reduced budget of 7000 evaluations per quantity: one DE run
(3328), CA (1024), PGD from eight starts chosen as above (248 evaluations) and two CMA-ES runs of 1200, from the
jittered centre and from the CA optimum. Its device is the best of these by development error and is scored on the
held-out samples, as are the nominal, MC, corner and field-searched devices.

\subsection{Budget-matched sampling and the pixel-level device}
The search budget is $N_{\rm match}=16{,}256$ development evaluations, summed over corners, DE, CA, PGD and CMA-ES.
The 1000 MC devices that seed one PGD start are not counted. The uniform and corner samplers continue the MC generator and the 200 corners, the Sobol'
sampler uses scrambled points~\citep{owen1998scrambling} from SciPy~\citep{virtanen2020scipy}, and the mixed sampler alternates uniform and corner
devices. For the Box seed-0 models the uniform stream continues to 100,000 devices. Supplementary Note~3 gives the
tail fits, a reproduction of the earlier MC and corner values and the pixel-level device. Quantisation, the extinction floor, static errors and piston act per mode, fringing uses the same kernel between
pixels, and each mode's transfer is the mean field over its cell. The shift half-width is now in SLM pixels, and the defocus, tilt and
reference-phase half-widths are unchanged. The sinc ablation uses a band-limited, non-passive shift, and
Random-error-aware training uses the nominal device.

\subsection{Statistics}
For an exceedance count $n_{\rm exc}$ of $N$ MC devices, we use the two-sided (equal-tailed) 95\%
Clopper--Pearson interval $[B_{0.025}(n_{\rm exc},N-n_{\rm exc}+1),B_{0.975}(n_{\rm exc}+1,N-n_{\rm exc})]$~\citep{clopper1934use},
with $B_q(a,b)$ the $q$ quantile of a beta distribution (endpoints 0 at $n_{\rm exc}=0$ and 1 at $n_{\rm exc}=N$). For
$n_{\rm exc}=0$ and $N=1000$ its upper end is 0.37\%, a pointwise bound for each model at 95\% confidence,
equivalent to a one-sided 97.5\% bound. The simultaneous bound
over the 120 models for $n_{\rm exc}=0$ is a one-sided 95\% bound, $B_{1-0.05/120}(1,N)=0.78\%$ (Bonferroni). Both bounds are conditional on the held-out inputs
and on uniform sampling of the box, and neither includes the finite-input-sample uncertainty. They bound
an exceedance probability and say nothing about the severity of rarer devices. They also assume a threshold set independently of
the MC draws, which does not strictly hold for the PGD start seeded at the best MC device
(Pm in Supplementary Tables~S1 and S2). The 95\% upper confidence bound on the 0.99 quantile is the order statistic $X_{(r)}$ with the smallest $r$ such that
$P(\mathrm{Bin}(N,0.99)\le r-1)\ge0.95$. Wilks' first-order one-sided 95/95 tolerance limit $W_1$ is the largest
of the first 59 MC errors, which exceeds the 95th percentile of the error with probability
$1-0.95^{59}=0.952$~\citep{wilks1941determination}. The second-order limit $W_2$ is the second largest of the first 93
(probability 0.950). We also computed $W_1$ for each of the 16 disjoint sets of 59 among the 1000 MC devices.
The illustrative HX thresholds are 1, 2 and 5\% of the test-set mean. Arms are compared paired by seed (seed-level values, means,
standard deviations and per-task sign tests, with no multiple-comparison correction).

\subsection{Compute}
On NVIDIA GH200 nodes, with three jobs sharing each GPU, the main experimental round (312 jobs, including the
sweep, shared knobs, circular-phase set and certificate) used about 8.8 allocated GPU-hours and the HX round (65 jobs)
about 2.5. The budget-matched, 100,000-device, HX pixel-level and transfer runs (56 jobs) used about 3.6, and the
correlated-box and HX-quantity runs (29 jobs) about 0.9. The mean-removed LDC check and the quantity-targeted HX
search (six jobs) took 3.1 job-hours (run time summed over jobs) on the same nodes, and the 36 pixel-level Darcy jobs
(both shift models) took 12.0 job-hours on one RTX 5070.

\section*{Data availability}
The lid-driven cavity dataset is from the conformalised neural-operator study of Kobayashi
et al.~\citep{kobayashi2025conformalized}, and the heat-exchanger dataset was generated by our group for a companion
study~\citep{roy2026adversarial}. The corresponding author can provide the Darcy and NS data, the data splits
used here, the raw evaluation outputs, the checksummed result exports and the model checkpoints upon reasonable
request.

\section*{Code availability}
The \texttt{wavesim} simulator and the training, evaluation, search, certificate and figure code are available from
the corresponding author upon reasonable request. One command runs the figure scripts, which regenerate every figure
and table and a ledger of every quoted number with its source.

\section*{Use of large language models}
Large language models were used to assist with editing the text of the manuscript and with debugging analysis code.
The authors designed the experiments, interpreted the results and reviewed and revised all manuscript text. No
large language model is an author of this work.

\section*{Acknowledgements}
This work used the Delta and DeltaAI systems at the National Center for Supercomputing Applications (awards OAC 2005572
and OAC 2320345) through allocation CIS240093 from the Advanced Cyberinfrastructure Coordination Ecosystem: Services \&
Support (ACCESS) program, which is supported by National Science Foundation grants \#2138259, \#2138286, \#2138307,
\#2137603 and \#2138296.

\section*{Author contributions}
S.R. conceived the study, designed and performed the experiments and analysed the results. J.Y. helped with the
writing. S.C. and S.B.A. guided and supervised the work. All authors contributed to writing and revising the manuscript.

\section*{Competing interests}
The authors declare no competing interests.

\setlength{\bibsep}{0pt plus 0.5pt}\renewcommand{\bibfont}{\small}
\bibliographystyle{unsrtnat}
\bibliography{references}

\ifdefined\withSI
\clearpage
\begingroup
\setcounter{figure}{0}\setcounter{table}{0}\setcounter{equation}{0}
\renewcommand{\thefigure}{S\arabic{figure}}\renewcommand{\thetable}{S\arabic{table}}%
\renewcommand{\theequation}{S\arabic{equation}}%
\renewcommand{\theHfigure}{S\arabic{figure}}\renewcommand{\theHtable}{S\arabic{table}}%
\renewcommand{\theHequation}{S\arabic{equation}}%
\makeatletter\setlength{\@fptop}{0pt}\makeatother
\gdef\shorttitle{Supplementary Information}%
\begin{center}
{\Large\bfseries Supplementary Information}\\[6pt]
{\large \papertitle}\\[6pt]
{\normalsize Samrendra Roy, Jason Yoo, Souvik Chakraborty, Syed Bahauddin Alam}
\end{center}
\vspace{4pt}
\noindent\textbf{Contents}\par
\setcounter{tocdepth}{1}%
\makeatletter\@starttoc{sic}\makeatother
\let\SIaddcontentsline\addcontentsline
\def\SItoc{toc}%
\renewcommand{\addcontentsline}[3]{\SIaddcontentsline{#1}{#2}{#3}\def\SIext{#1}\ifx\SIext\SItoc
  \SIaddcontentsline{sic}{#2}{#3}\fi}%
\def\maketitle{}\def\tableofcontents{}\def\bibliographystyle#1{}\def\bibliography#1{}%

\maketitle
\tableofcontents

\section*{Abbreviations}
\addcontentsline{toc}{section}{Abbreviations}
ADC, analogue-to-digital converter; CA, coordinate ascent over box corners; CMA-ES, covariance matrix adaptation
evolution strategy; DAC, digital-to-analogue converter; DE, differential evolution; \DET, deterministic simulator
configuration (no converters or detection noise); FFT, fast Fourier transform; FNO, Fourier neural operator; FT,
fine-tune; GELU, Gaussian error linear unit; HX, heat exchanger; LDC, lid-driven cavity; LO, local oscillator; MC,
Monte Carlo; MLP, multilayer perceptron; NS, Navier--Stokes; PDE, partial differential equation; PGD, projected
signed-gradient ascent; pp, percentage points; \REAL, simulator configuration with converters and detection noise;
RMS, root mean square; SAM, sharpness-aware minimisation; SLM, spatial light modulator; TKE, turbulent kinetic energy; UCB, upper confidence bound.

\section*{Supplementary Note 1: Simulator versions, devices and known artefacts}
\addcontentsline{toc}{section}{Supplementary Note 1: Simulator versions, devices and known artefacts}
All optical hardware in this work is simulated with \texttt{wavesim}. No device was built or measured.

\paragraph{Simulator versions.} All main-text results use \texttt{wavesim} v2 (the mode-level device), except the
results on the pixel-level device (Supplementary Note~3), which use \texttt{wavesim} v3. A first round, the ``v1 pilot''
below, used an earlier configuration with two defects. Its crosstalk blurred the amplitude and the \emph{unwrapped}
stored phase separately, so the result depended on the arbitrary $2\pi$ representation of each pixel. Its 8-bit
amplitude quantiser also acted after the extinction floor, which made the mask discontinuous in the extinction knob.
Version 2 blurs the complex mask $\alpha\,e^{i\varphi}$ (field model) or the amplitude and $e^{i\varphi}$
(circular-phase model), and both are invariant to $2\pi$ wraps. It quantises before applying the extinction floor and adds
local-oscillator-limited shot noise to coherent detection. With its legacy options selected, v2 reproduces v1 bit
for bit, but the experiments use the revised options. A hash of the full configuration is stored with every
result. We fixed the primary endpoint and the evaluation protocol before the main (v2) experimental round. No statistic mixes the two
versions, and v1 numbers appear only in Supplementary Notes~6, 7 and 8, where they are labelled.

\paragraph{Unit tests.} Fourteen unit tests passed on the compute nodes, and a fifteenth, which compares the code
with a saved copy of the v1 package, passed in the local smoke test. They check
bit-for-bit agreement with the v1 defaults, invariance to $2\pi$ phase wraps, gradients, the noise model, the box
projection, the quantisation order and deterministic scoring. The v3 code adds 17 tests, which also passed on the
compute nodes (Supplementary Note~3).

\paragraph{Static draws depend on the GPU.} Per-device static errors are drawn from a seeded generator on the
model's GPU, so a static seed gives a different draw on different GPUs. All mode-level (v2) results, including HX, come from GH200
nodes, and so do the v3 results except the Darcy runs on the pixel-level device, which ran on an RTX 5070.

\paragraph{Shift gradient at zero.} Bilinear sub-pixel resampling returns the mask unchanged at exactly zero shift,
where the gradient with respect to the shift knobs vanishes. In the v1 pilot, the zero registration of the searched
devices was therefore partly an artefact of this vanishing gradient at the box centre. In the v2 protocol, the
gradient search that starts at the centre begins from a jittered centre ($z\sim U[-0.05,0.05]$), where the gradient does not vanish. The searched
devices of the box-trained Darcy, NS and HX models still had registration shifts within 0.05 of zero (main text).
The exports contain only the final coordinates of each search, not its trajectory.

\paragraph{Detection, crosstalk and registration.} Coherent detection returns the real part of the output field,
as an ideal holographic readout would. The local-oscillator shot noise has a standard deviation of
$\eta_{\rm LO}/\sqrt{N_{\rm ph}}$ of full scale ($\eta_{\rm LO}=1/\sqrt2$, $N_{\rm ph}=10^4$ photons), and read noise is added.
The per-sample automatic gain of the ADC is an oracle-like idealisation (main-text Methods and Limitations), and the
primary \DET\ endpoint has no ADC. The main-text Methods give the crosstalk kernel, the order of operations per mask
and the zero-padded registration model.

\section*{Supplementary Note 2: Random sampling versus search}
\addcontentsline{toc}{section}{Supplementary Note 2: Random sampling versus search}

\paragraph{Endpoint and protocol.} The endpoint $R_N$ (main-text equation~(1)) compares two held-out errors on \DET\
with static draw $s_0=0$. The numerator is the error of the searched device $z_{\text{search}}$, the device with
the highest development error over all searches. The denominator is the largest error among the first $N$ of
1000 MC devices. Throughout, `best' means highest error, the objective of every search. In 3 of 120 models, the best device of the unseen attacks, also selected on development error,
scored higher on the held-out samples than $z_{\text{search}}$. These are Darcy Random-error-aware seed~0, Darcy
SAM FT seed~0 and LDC SAM FT seed~1. We also scored each method's development-selected device on the full test set.
The full-test error of the searched device was 0.975--1.038 times its held-out error. Supplementary
Tables~\ref{tab:models_a}--\ref{tab:real_b} list every model, and Supplementary Fig.~\ref{fig:s1} shows their errors.

\paragraph{Reliability.} For every model we record the number $n_{\rm exc}$ of the 1000 MC devices whose error exceeds $c$
times the nominal error ($c=1.1,1.25,1.5,2$), and the number whose error exceeds the searched error. The latter was 0
in all 120 models. This gives the bounds of 0.37\% (pointwise; upper end of the two-sided 95\% Clopper--Pearson
interval) and 0.78\% (simultaneous over the 120 models; one-sided 95\%, Bonferroni), under the conditions stated in
the main-text Methods. Supplementary Fig.~\ref{fig:exc} shows the exceedance curves. For the 100 box-trained models,
$n_{\rm exc}$ at $c=2$ was 0--3.
The searched error was 1.08--3.06 times the 95\% upper confidence bound on the 0.99 quantile.

\paragraph{An additive heuristic.} Suppose the error were additive in the normalised knobs,
$e(z)=e_0+\sum_{j=1}^D g_j z_j$, with $z_j$ independent and uniform on $[-1,1]$. Then the mean is $e_0$, the worst
case is $e_0+\|g\|_1$ and the variance is $\|g\|_2^2/3$, and these three statements are exact for the additive model.
For many knobs, $e$ is approximately Gaussian, and the expected maximum of $N$ devices is approximately
$e_0+\|g\|_2\sqrt{2\ln N/3}$. The ratio of the worst-case excess to the max-of-$N$ excess is then approximately
\begin{equation}
\frac{\|g\|_1}{\|g\|_2}\sqrt{\frac{3}{2\ln N}}\;\le\;\sqrt{\frac{3D}{2\ln N}}.
\end{equation}
The inequality is Cauchy--Schwarz, with equality when all knobs contribute equally. The Gaussian and expected-maximum
steps are approximations (the leading-order $\sqrt{2\ln N}$ overstates the expected Gaussian maximum at these $N$),
so the result is a heuristic, not a finite-$N$ theorem. For $D=32$ the equal-contribution
value is 3.01 ($N=200$) and 2.64 ($N=1000$). The observed ratios $(e(z_{\rm search})-\bar e)/(\max_{1000}e-\bar
e)$, with $\bar e$ the MC mean, were 2.1--4.0 (Darcy), 3.2--5.4 (NS), 2.0--5.4 (LDC) and 3.4--5.2 (HX) for the
box-trained arms. They were 1.5--2.2 for the Random-error-aware arm. Where they exceed the equal-contribution value,
the Gaussian heuristic does not account for the excess. The denominator is one realised sample maximum
rather than its expectation, so the comparison is not a test of additivity. The heuristic only illustrates why more
toleranced knobs can widen the gap. Extreme-value theory~\citep{coles2001introduction} and tail estimation from
extreme order statistics~\citep{pickands1975statistical} describe the maximum of $N$ samples under the sampling law,
not the supremum over the box (Supplementary Note~3).

\paragraph{Search strength and budgets.} Supplementary Fig.~\ref{fig:s2} shows, per task and arm, each method's best
held-out error relative to the best over all methods. PGD from the jittered centre was strongest for the box-trained
Darcy, NS and HX models. For LDC no single method dominated, and for the Random-error-aware models CA was strongest.
The best-so-far curves (main-text Fig.~2c) show that DE and CMA-ES were still improving at the
end of their budgets, so the searched errors are lower bounds on the worst-case error, not estimates of it, and may
lie well below it. The PGD curve counts only the four
independent starts, at 31 forward passes each (124 in total). The four seeded starts begin from the results of DE,
CA, the corners and MC, and their inherited cost is not counted.

\paragraph{Local linear estimate.} The estimate $e_0+\sum_j\max(g_j^+,-g_j^-,0)$ (main-text Methods) was
0.27--1.79 times the searched error across models, with both evaluated on the full test set (Supplementary
Fig.~\ref{fig:s3}). It is not a reliable substitute for search.

\section*{Supplementary Note 3: Budget-matched baselines and a pixel-level device model}
\addcontentsline{toc}{section}{Supplementary Note 3: Budget-matched baselines and a pixel-level device model}
This note asks whether random testing falls short only because it evaluates fewer devices than the search, and whether
the gap depends on the mode-level device model. The runs reuse the trained models
and the evaluation code of the main round, and all numbers come from checksummed exports of the result files.

\paragraph{Budget and samplers.} The search spends $N_{\rm match}=16{,}256$ development evaluations per model: 200
random corners, three DE runs of 3328 evaluations, 1024 for CA, eight PGD starts of 31 evaluations each (30 of
them with a gradient) and four CMA-ES runs of 1200. The 1000 MC devices only seed one PGD start and are
not counted, and with them the total would be 17,256. On the 24 seed-0 models (mode-level device, \DET, static draw 0),
four samplers each draw $N_{\rm match}$ devices. The uniform sampler continues the MC generator, so its first 1000
devices are the MC devices of the main protocol. The Sobol' sampler takes the first 16,256 points of a scrambled
Sobol' sequence in 32 dimensions. The corner sampler continues the corner generator, whose first 200 devices are the
main corners. In the mixed sampler, device $2i$ is the $i$-th uniform device and device $2i+1$ the $i$-th corner. The
mixed sampler reuses the uniform and corner evaluations, so its devices are not independent of the other two. As for the search, a sampler selects the device with the highest
development error among its first $n$ devices and scores it on the held-out samples. We define $R_{\rm match}$ as
the held-out error of $z_{\rm search}$ divided by that of the sampler's device at $n=N_{\rm match}$.

\paragraph{Reproduction check.} Each run first re-scored the first 1000 uniform and 200 corner devices, the nominal
device and $z_{\rm search}$. The MC and corner errors matched the main evaluation device by device, $z_{\rm search}$
matched to a relative difference of $9.7\times10^{-6}$, and the recomputed nominal error, MC mean and MC maxima equalled the values
of Supplementary Tables~\ref{tab:models_a} and \ref{tab:models_b} in all 24 models.

\paragraph{Budget-matched results.} Supplementary Table~\ref{tab:v3e1} and Fig.~\ref{fig:v3budget} give the
results. For uniform sampling $R_{\rm match}$ was 1.21--2.26 (median 1.53), for Sobol' points 1.19--2.16 (median
1.50) and for corners 1.00--1.90 (median 1.16), and the mixed sampler stayed within 0.06 of the corners. Corners
came within 10\% of the searched error in 8 of 24 models: the four Random-error-aware models, Darcy Mixed-PGD FT
and SAM FT, and LDC Box and Mixed-PGD FT. For the HX Box model and its three fine-tunes, $R_{\rm match}$ for corners
stayed at 1.52--1.90. Uniform
sampling improved slowly. Between 1000 and 16,256 devices, the held-out error of its selected device changed by a
factor of 0.97--1.14, because a higher development error does not always mean a higher held-out error.

\paragraph{100,000 devices and tail fits.} For the Box seed-0 model of each task the uniform stream continued to
100,000 devices (Supplementary Table~\ref{tab:v3mc} and main-text Fig.~2b). No device reached the searched
development error. The best of 100,000 reached 0.47--0.68 of it, and the device selected from 100,000 by
development error gave a held-out ratio $R$ (at $10^5$ devices) of 1.45--2.12. We fitted a generalised Pareto
distribution by maximum likelihood to the 1000 excesses over the 0.99 quantile of the development
errors~\citep{pickands1975statistical,coles2001introduction}. We define $N^*$ as the number of devices at which the
expected maximum under the fitted tail reaches the searched development error, with the threshold at the 0.98, 0.99
or 0.995 quantile. It is infinite when the shape $\xi$ is negative and the fitted upper endpoint lies below that
error, as for LDC at all three thresholds, NS at 0.98 and 0.99 and HX at 0.98. The shape was $-0.10$ to $+0.01$, and
every finite $N^*$ was $2\times10^{10}$ or larger. In 200 bootstrap resamples at the 0.99 quantile, the 5th percentile of $N^*$ was
$9.5\times10^{9}$ or larger, and 17.5--92.0\% of the resamples gave an infinite $N^*$. The fitted expected maximum of
100,000 devices was 0.94--1.01 times the observed one. The fits imply that errors as large as the searched one are
extremely rare or unreachable under uniform sampling. For NS and LDC, however, the fitted endpoint at the 0.99
quantile lies below the searched development error (21.6\% against 26.2\% for NS, 2.06\% against 2.54\% for LDC),
an error the search reaches inside the box. The fitted law thus under-represents the far tail, tail extrapolation is
unreliable for locating the worst device, and we rely on the empirical result. Block-maxima (generalised extreme value) fits with only 100
blocks of 1000 devices were unstable, and their extrapolation conflicted with the zero exceedances observed among
the 100,000 devices, so we do not use them.

\paragraph{Pixel-level device.} In the mode-level device of the main text, one SLM pixel is one Fourier mode.
Crosstalk and the bilinear registration shift therefore mix neighbouring modes directly, and a 0.3-mode shift puts
30\% of a mode's weight on its neighbour. The pixel-level device (\texttt{wavesim} v3) changes this as follows.
\begin{itemize}
\item Each mode spans $4\times4$ SLM pixels, so the pupil has $100\times100$ pixels. Passivity, the 8-bit
amplitude with the extinction floor, the 8-bit phase, the static errors, the per-path piston and the reference
phase act per mode as before. The static errors keep their draw and strength and are uniform over a mode's block.
\item Fringing is a Gaussian blur of the complex field at the pixel level, with the kernel of the main text, width
$\gamma$ in SLM pixels and replicate padding.
\item The registration shift moves the SLM and the aperture that bounds it by $\Delta$ pixels. The field in each
pixel is the exact integral of the displaced flat-top pixel profile over that pixel, which is linear interpolation
between neighbouring pixel values, with zeros outside the aperture. The weights are non-negative and sum to at most
1, so the shift is passive. As before, its gradient vanishes at $\Delta=0$, and PGD starts from a jittered centre.
\item Defocus and tilt are fixed to the optical axis and act at the pixel centres after the shift.
\item The transfer of each mode is the mean of the realised complex field over its $4\times4$ cell, the top-hat
(cell-average) approximation of how the optical system resolves the Fourier plane. It is exact for masks that are constant over each cell,
where it reduces to the mode-level transfer, and it is passive.
\end{itemize}
All steps are separable, so the transfer is computed exactly with $25\times25$ operators per axis. A 0.3-pixel shift
puts 7.5\% of a mode's weight on its neighbour. The box keeps the shift ($\pm0.3$ pixel, where one pixel is now a
quarter of a mode), defocus, tilt and reference-phase half-widths. Fringing spans 0.5--0.6 pixel and the extinction ratio 27--30\,dB.
The nominal device has zero shift, defocus, tilt and reference phase, 0.5-pixel fringing and 30\,dB, so it is not
the box centre (0.55 pixel, 28.5\,dB). In the ablation, the shift is band-limited: a phase ramp in the conjugate
domain, evaluated as a sinc matrix. It is the identity at $\Delta=0$ and exact for integer shifts, but it treats the
pixel values as samples of a band-limited field, rings at block and aperture edges and is not passive (a fully
transmitting mask reaches $|T|=1.04$). The 17 unit tests compare the cell transfer with a brute-force fine-grid
integral (agreement to $2.5\times10^{-8}$), check passivity and the reduction to the mode-level model for one pixel per mode,
check the gradients for all 32 knobs, and test the samplers, the budget count and the tail fits.

\paragraph{Runs on the pixel-level device.} We retrained the Random-error-aware, Box and Mixed-PGD FT arms on Darcy
and HX (seeds 0--2) with the main settings and the new box. Random-error-aware training uses the nominal device.
Each model received the full evaluation protocol on \DET\ of the new device, the \REAL\ secondary endpoint and the
uniform, corner and mixed samplers to $N_{\rm match}$. HX ran on GH200 nodes with the cell integral. Darcy ran on an
RTX 5070 with both shift models. For the transfer evaluations, the Box and Mixed-PGD FT seed-0 models of Darcy and
HX trained on the mode-level device were evaluated on the new device (cell integral).

\paragraph{Results on the pixel-level device.} Supplementary Table~\ref{tab:v3e2} lists every model. $R_{200}$ was
1.59--2.42 on HX, 1.33--1.65 on Darcy, 1.25--1.69 on Darcy with the sinc shift and 1.18--1.49 for the transferred
models; no MC device reached the searched error. Mixed-PGD FT had the lowest searched error in every seed under both
shift models. The Random-error-aware models reached 9.85--11.2 (HX) and 2.30--2.89 (Darcy) times their nominal error
on the searched device, and 1.96--2.42 on Darcy with the sinc shift. For the transferred models the corner sampler
was weaker than uniform sampling ($R_{\rm match}$ 1.26--1.63, compared with 1.12--1.42). The nominal device was worse than
the MC mean for every HX Box model (1.51--1.58) and Darcy Box model (1.03--1.08), but slightly better than it for
the Darcy Box models with the sinc shift (0.95--0.98). The last column of Supplementary Table~\ref{tab:v3e2}
describes the searched device. With the cell integral, all four layers were at the sharpest fringing and the lowest
extinction ratio, with 6--8 of the 8 shift coordinates within 0.05 of zero, for all six HX Box and Mixed-PGD FT
models and all four transferred models. On Darcy this held for seed~2 of Box and Mixed-PGD FT only. Under the sinc
shift it held for none of the nine models, although one searched device had all layers at the sharpest fringing and
lowest extinction ratio, with only 4 of 8 shifts near zero. The other Darcy searched devices sat at the blurred, high-extinction end with
shifts at the box edges.

\paragraph{Field snapshots.} Main-text Fig.~5b and Supplementary Figs~\ref{fig:fieldsldc} and \ref{fig:fieldsother}
show one held-out input for each of HX, LDC and Darcy: the sample with the median error under the nominal device of the task's Box
seed-0 model, so every model of a task shows the same input. The predictions are those of the nominal device, the
worst of the first 200 MC devices (selected by development error) and $z_{\rm search}$, on \DET\ with static draw
0. LDC and HX use the mode-level device and Darcy the pixel-level device. The single-sample nominal errors agree with
the batched protocol scores to $1.7\times10^{-5}$ (relative). The LDC fields are, in stored order, the pressure $p$,
the turbulent kinetic energy $k$ and the velocity magnitude $|\mathbf{u}|$ of the source's Reynolds-averaged
simulations~\citep{kobayashi2025conformalized}. The fields match this order: $|\mathbf{u}|$ vanishes on the three
fixed walls and equals the final lid speed along the lid, and $k$ is non-negative. The joint relative $L_2$ error, the LDC metric,
equals the pressure error to $6\times10^{-5}$ for this sample, whereas $k$ and $|\mathbf{u}|$ have relative errors
of 7--17\% on all three devices. Supplementary Table~\ref{tab:ldcfields} gives the per-field errors of every LDC arm
(means over five seeds) on the full held-out set.

\section*{Supplementary Note 4: Other tolerance sets and crosstalk models}
\addcontentsline{toc}{section}{Supplementary Note 4: Other tolerance sets and crosstalk models}
In the shared-knob box (Supplementary Table~\ref{tab:shared}), all four layers use the same eight knob values. It uses the
full search protocol and all 256 corners. The tolerance sweep (Supplementary Table~\ref{tab:sweep}) scales all
half-widths by $\kappa\in\{0.25,0.5,1,1.5,2\}$ with a reduced protocol: 200 MC devices,
100 corners, one DE seed (10 generations), CA (512 evaluations) and four PGD starts. Its $\kappa=1$ values therefore differ slightly from the main
protocol. The circular-phase set (Supplementary Table~\ref{tab:cp}) retrains three arms (Random-error-aware, Box and
Mixed-PGD FT) with circular-phase crosstalk on seeds 0--2 for Darcy, NS and LDC, giving 27 models and no HX. Its
search uses one DE seed and one CMA-ES seed from the centre. Because the models were retrained, each comparison with
field crosstalk is between two complete training-and-simulator configurations.

\section*{Supplementary Note 5: Heat-exchanger task}
\addcontentsline{toc}{section}{Supplementary Note 5: Heat-exchanger task}
\paragraph{Data.} The main-text Methods describe the heat-exchanger task, its data from
ref.~\citep{roy2026adversarial} and the 102-dimensional input. The 3977 node coordinates lie in $[-1,1]^2$. The splits
are 1200 training, 173 validation (not used) and 173 test cases, of which 64 are development and 109 held-out. The
channels are, in order, gauge pressure (Pa), then the in-plane $z$- and
$y$-velocities (m\,s$^{-1}$) on the $(y,z)$ cross-section at $x=0.79$\,m. The input participation ratio is 1.0 (1.04 after
normalisation), and one principal component holds 90\% of the input variance.

\paragraph{Readout.} The two faces of the tape insert carry nearly coincident nodes, and the mesh is strongly refined.
Interpolating the true fields to a $G\times G$ grid and back to the nodes therefore costs 5.3\%, 3.4\% and 1.9\% at
$G=65$, 97 and 129. The model instead predicts the fields at the mesh nodes from interpolated latent features. It
samples the 24 features of the last hybrid layer at the node coordinates, appends a 26-dimensional coordinate encoding and
applies a per-node digital MLP. The optical layers act on a $65\times65$ latent grid, exactly as for the grid tasks,
so the tolerance box is unchanged. Supplementary Fig.~\ref{fig:arch} shows this architecture next to the LDC variant,
with the tensor shape at each stage.

\paragraph{Metric.} The primary metric is the mean over the three channels of the per-channel relative $L_2$ error.
Each channel error is computed per sample over the nodes in physical units and then averaged over samples. The joint
relative $L_2$ error over all channels, averaged over samples, equals the pressure-channel mean error to the
reported precision, because pressure is about $10^3$ times larger than the velocities. The median of the joint error
over held-out samples is not equal to the pressure-channel mean (digital FNO: 0.211\% median, 0.236\% pressure-channel
mean). Supplementary Table~\ref{tab:hx} gives both, and Supplementary Fig.~\ref{fig:hxmetric} plots them for every
model. Averaged over seeds, pressure had the largest of the three channel errors for every arm and device. One
individual model--device pair (SAM FT seed~2, searched device) was an exception.

\section*{Supplementary Note 6: A Taylor-model certificate (negative feasibility study)}
\addcontentsline{toc}{section}{Supplementary Note 6: A Taylor-model certificate (negative feasibility study)}
\paragraph{Bound and computation.} For the deterministic device with one frozen static draw and one Darcy test input
$a$ with target $u$, the certificate computes a number $B$ with
$\sup_{\theta\in\Theta_\kappa}\|u_\theta(a)-u\|/\|u\|\le B$, where $\Theta_\kappa$ is the box scaled by $\kappa$. Per-input bounds are averaged over ten test inputs. This is a
separate per-input calculation, not the primary common-device endpoint. The mask of each layer is expanded to second
order in ten smooth variables per layer, with exact coefficients from forward-mode differentiation. These are the
signed and absolute registration shift per axis, defocus, two tilts, crosstalk, extinction and reference phase. The
third-order remainder is bounded entrywise with interval arithmetic~\citep{moore2009interval}. The network is
decomposed exactly into a linear part plus quadratic, bilinear and remainder sources. Remainder norms are propagated
with the exact block-diagonal operator norm of each spectral layer~\citep{makino2003taylor}.

The bound is sound only conditional on exact real arithmetic for the simulator as specified. No directed rounding is
used, so it is not a machine-checked floating-point certificate. A functional re-implementation matches the
simulator to $4.1\times10^{-5}$ relative error, and a second run adds a float32 slack of $10^{-5}$ per mask entry.
For every input, each row of the $\kappa$ scan also evaluates eight random interior devices and eight random
corners. No
sampled device exceeded the bound in any of the 48 rows of the $\kappa$ scans, including those with the float32 slack. This is a sampled validation, not a proof.

\paragraph{Expansion point.} The Taylor model is expanded about the centre of its relaxed variables, which is not the
nominal device. For boxes that contain zero shift, the absolute shift $|\Delta|$ sits at the midpoint of its
bounding interval $[0,w]$ ($w>0$), which corresponds to a slightly blurred mask. The extinction-dependent amplitude
is replaced by its secant across the extinction interval. The error at this expansion point therefore moves with
$\kappa$ (Supplementary Table~\ref{tab:cert}). For the Box model it falls from 2.396\% at $\kappa=0.001$ to
1.710\% at $\kappa=1$. With extinction fixed, only the $|\Delta|$ effect remains (1.668\% at $\kappa=1$).

\paragraph{Result (v2 simulator).} Supplementary Table~\ref{tab:cert} and Fig.~\ref{fig:s4} give the scan. For the
Darcy Box and Mixed-PGD FT models of seed~0 with all 32 knobs, the bound was 1.01--1.09 times the highest error found
up to $\kappa=0.005$. It was 2.03--2.46 times at $\kappa=0.01$, 127--160 times at $\kappa=0.03$ and
$6\times10^8$--$8.5\times10^8$ times at the full box. With the extinction knob fixed, the values were 1.05--1.06 at
$\kappa=0.005$ and 1.71--2.00 at $\kappa=0.01$. With the float32 mask slack, the bound was already 1.72--2.02 times
the highest error found at $\kappa=0.001$. Quantising before the extinction floor removed the discontinuity that
made the v1 enclosure vacuous at every $\kappa$. It did not remove the growth of the bound at large $\kappa$, whose
cause the experiment cannot isolate (main-text Section~2.8).

\paragraph{Mechanism (v1 pilot, legacy simulator).} The following decomposition comes from the v1 pilot. We did not
recompute it under v2, so it is a historical record and does not explain the v2 result. In the v1 pilot the
device-to-output map was strongly nonlinear at the stated tolerances. Along the worst direction found for the
Mixed-PGD FT model, the Taylor terms of orders one to four were 0.025, 0.022, 0.018 and 0.007 times $\|u\|$. The
second-order mass $\tfrac12\sum_{jk}\|\partial^2u/\partial x_j\partial x_k\|\,r_jr_k/\|u\|$, with $r_j$ the
half-width of variable $j$, was 0.27, about twice the first-order mass (0.13). Nonlinearity was distributed across
many knob variables, with per-layer amplification, which limits Taylor-model certificates to small boxes. In detail,
80\% of the second-order mass sat in 22 of 40 variables (13 for the Random-error-aware model), and 48\% of it (71\%
for the Random-error-aware model) coupled different layers. Splitting the box along a few knobs therefore could not
help. Norm bounds through globally mixing
spectral layers also cannot see the downstream cancellations that kept the box-trained models' output change small.

\section*{Supplementary Note 7: v1 pilot under the legacy simulator}
\addcontentsline{toc}{section}{Supplementary Note 7: v1 pilot under the legacy simulator}
The v1 pilot (Darcy, local RTX 5070, legacy unwrapped-phase crosstalk and extinction-then-quantise order) informed
the v2 design. Its crosstalk model depends on the phase representation, so its effect decomposition is retained only
as a record. Applied alone to three passive-ideal models, 0.5-pixel crosstalk multiplied the error by 9.76 on average,
a $(0.3,0.3)$-pixel registration error by 6.88 and the 30-dB extinction floor by 5.29. Each quantisation step (8-bit masks, input DAC
and output ADC) and each detection-noise source, applied alone, changed it by at most a factor of 1.028
(Supplementary Table~\ref{tab:decomp}); the largest change came from the 8-bit mask phase. The model trained with random
static errors reached 2.15\% on its own device class but 19.7\% on clean passive optics, an early indication of
co-adaptation. The median wrapped phase step between neighbouring Fourier modes was 1.50--1.57\,rad across
nine v1 Darcy checkpoints, close to the $\pi/2$ that independent uniform phases give, so the trained phases were
almost uncorrelated between neighbouring modes. For the model trained with random static errors, we treated its
per-pixel static errors (2.88 million variables) as a box. Twenty random draws inside it changed the error by at
most 0.1 percentage points (2.01\% to at most 2.10\%). A gradient search over the same variables reached 48.9\% (Supplementary Fig.~\ref{fig:s5}). A
coordinated perturbation of millions of variables is not a physical tolerance, and such errors are better specified
statistically.

\section*{Supplementary Note 8: Chip portability (frozen-library pilot)}
\addcontentsline{toc}{section}{Supplementary Note 8: Chip portability (frozen-library pilot)}
A separate exploratory pilot under the v1 simulator studied frozen optical libraries. One bank of four passive 4f
blocks was pretrained jointly on Darcy, a variable-coefficient heat equation and a Helmholtz problem, with a fresh
simulated device on every forward pass, and then frozen. For NS, a task outside the pretraining set, a 6209-parameter
digital adapter was trained on one frozen simulated chip (one simulated device) and tested on ten others. The control replaces the learned
library by frozen random optics. In most settings, the learned library had a smaller cross-chip penalty (main
configuration: +1.16, compared with +1.76 percentage points). It did not with a slim adapter or 16 optical blocks, and with
intensity detection the library failed (57\% error). Device-randomised adapter training removed the penalty
(Supplementary Table~\ref{tab:port}, Fig.~\ref{fig:s6}). These mostly single-seed, mean-over-chip results are
exploratory and do not enter the main worst-case claims.

\section*{Supplementary Note 9: Correlated tolerances and engineering quantities}
\addcontentsline{toc}{section}{Supplementary Note 9: Correlated tolerances and engineering quantities}
Both studies are evaluation only. They reuse the trained models, the evaluation code and the static draw 0 of the
main round, and all numbers come from checksummed exports of the result files.

\paragraph{Correlated tolerance box.} Layers built with the same SLM type share its crosstalk and extinction ratio,
whereas the registration shift, defocus, tilt and reference phase of each layer depend on its own alignment. The
correlated box therefore has one crosstalk and one extinction coordinate for all four layers and six independent
coordinates per layer, $D=2+4\times6=26$, with the centres and half-widths of the main box. It contains the shared
8-D box and lies inside the independent 32-D box. A device $z\in[-1,1]^{26}$ is mapped to the 32 knobs by repeating
the two shared coordinates in every layer, both when devices are scored and in the differentiable path used by PGD.
Ten unit tests on the compute nodes checked that this map reproduces the scores of the main evaluation code bit for
bit, that it reduces to the 32-D and 8-D searches, that the gradient reaches all 26 coordinates and that the budget
is counted correctly. The protocol is that of the main round on \DET\ in 26 dimensions: 1000 MC devices, 200 random
corners, three DE runs of 2704 evaluations, CA (1024), eight PGD starts (248) and four CMA-ES runs of 1204
($\lambda=14$), so the search spends 14,400 development evaluations. The \REAL\ secondary endpoint and the nested
search were not run. The nominal error equalled the main-round value in all 24 models.

\paragraph{Results for the correlated box.} Supplementary Table~\ref{tab:hybrid} lists the 24 seed-0 models.
$R_{200}$ was 1.16--2.51 and $R_{1000}$ 1.14--2.19, so no MC device reached the searched error. $R_{200}$ was below
the 32-D value of the same model in 23 of 24 models (the exception is LDC Random-error-aware, 1.40 against 1.36) and
above the 8-D value in all 18 models with an 8-D run. The smaller gap comes mostly from the MC maximum. The worst of
200 MC devices was 0.90--1.19 times its 32-D value (higher in 21 of 24 models), whereas the searched error was
0.88--1.01 times the 32-D searched error. Every device of the correlated box is also a 32-D device, yet the searched error
in the correlated box exceeded the 32-D one in 12 of 24 models (by at most 1.2\%), so the 32-D searches did not always find the worst
device. In the same way, the searched error in the correlated box was below the 8-D searched error in 7 of 18 models (down to 0.98
times). Both searches give lower bounds. In 19 of the 20 box-trained models, the searched device had the sharpest
crosstalk (0.4 pixel) and the lowest extinction ratio (27\,dB) of the box. The exception, LDC Smooth-mask box, and
all four Random-error-aware models sat at the opposite corner (0.6 pixel, 33\,dB).

\paragraph{Engineering quantities on the heat-exchanger slice.} The HX outputs are the gauge pressure $p$ and the
in-plane velocities $u_1$ and $u_2$ at the 3977 nodes of the $(y,z)$ slice at $x=0.79$\,m, in physical units. Node
areas come from a Delaunay triangulation of the scaled node coordinates; the scaling is isotropic, so area ratios
are physical. A triangle is dropped if its longest edge exceeds three times the 0.99 quantile of all edge lengths
or if its centroid lies outside the tube wall, a circle fitted to the convex hull (maximum radial residual
$5.6\times10^{-4}$ of the radius). Neither rule dropped a triangle on this mesh: all 7807 were kept, and their area
equals that of the convex hull, 99.9\% of the fitted disc. Each node receives one third of the area of every
triangle it belongs to, and the weights $w_n$ are normalised to sum to 1. They span a factor of $7\times10^5$,
because the mesh is strongly refined (Supplementary Note~5). The quantities are
\begin{itemize}
\item $Q_p=\sum_nw_np_n$, the slice-mean gauge pressure. It equals the pressure drop from the slice to the outlet
only if the outlet is at gauge zero. The data description does not state the reference, so this reading is an
assumption.
\item $Q_s=(\sum_nw_n(u_{1,n}^2+u_{2,n}^2))^{1/2}$, the root-mean-square in-plane speed, a measure of the secondary
(swirling) flow that the tape insert is meant to create.
\item $Q_{\max}$, the 99th percentile of the in-plane speed $(u_1^2+u_2^2)^{1/2}$ over the nodes (unweighted), a
robust peak.
\end{itemize}
Over the 173 test samples, $Q_p$ was 6.19\,kPa on average (s.d.\ 0.70\,kPa, range 5.04--7.38\,kPa), $Q_s$
1.63\,m\,s$^{-1}$ (s.d.\ 0.10\,m\,s$^{-1}$, range 1.46--1.79\,m\,s$^{-1}$) and $Q_{\max}$ 3.49\,m\,s$^{-1}$ (s.d.\
0.21\,m\,s$^{-1}$, range 3.14--3.83\,m\,s$^{-1}$). The
error of a quantity for one sample is $|Q_{\rm pred}-Q_{\rm true}|$ divided by the mean of $|Q_{\rm true}|$ over the
test set, so that a small true value cannot inflate it. We report its mean over the 109 held-out samples, in percent.
The 30 main-round HX models (mode-level device) and the nine HX models on the pixel-level device (Supplementary
Note~3) were each scored on the nominal device, on the searched device $z_{\rm search}$ of their evaluation and on
all 1000 MC devices of that evaluation, on \DET\ with static draw 0. The recomputed field errors matched the
evaluation results to $6.7\times10^{-6}$ (relative).

\paragraph{Results for the engineering quantities.} Supplementary Table~\ref{tab:hxqoi} summarises them per arm, and
Supplementary Fig.~\ref{fig:hxqoi} shows every model. Among the 30 main-round models, the searched device's error
exceeded the largest error over all 1000 MC devices in 24 ($Q_p$), 20 ($Q_s$) and 23 ($Q_{\max}$) models, and in all
three quantities at once in 12. On the pixel-level device it did so in 9, 9 and 6 of 9 models. The
Random-error-aware models were affected most. Their $Q_p$ error rose from 0.09--0.16\% at the nominal device to
6.3--6.8\% on the searched device (mode-level device) and from 0.13--0.17\% to 3.2--3.5\% (pixel-level device), 21--74
times the nominal error. For the 31 box-trained models, the largest error of any quantity on any device scored
here was 1.64\% (Smooth-mask box seed~4, $Q_p$), and the largest MC maximum was 0.94\%. The search maximised the
field error, and a device with a larger field error need not have a larger error in a given quantity. Among the 25
box-trained main-round models, the searched device's error was below the MC maximum for $Q_p$ in 6, for $Q_s$ in 10
and for $Q_{\max}$ in 7, down to 0.27 times it. For Box seed~0, 297 of the 1000 MC devices gave a $Q_p$ error at least
as large as the searched device (ratio 0.46), although its field error was 2.4 times the MC maximum. On the pixel-level
device, the searched $Q_{\max}$ error of the three Mixed-PGD FT models was 0.94--0.98 times the MC maximum. A search
that maximises the error of the quantity itself gives a tighter lower bound on its worst case; Supplementary Note~10
reports one for $Q_p$ and $Q_s$.

\section*{Supplementary Note 10: Pressure offset in LDC and quantity-targeted search with Wilks 95/95 limits}
\addcontentsline{toc}{section}{Supplementary Note 10: Pressure offset in LDC and quantity-targeted search with Wilks 95/95 limits}
Both studies are evaluation only. They reuse the trained models, the evaluation code and the static draw 0 of the
main round on \DET, and all numbers come from checksummed exports of the result files. Ten unit tests passed on the
compute nodes. They check that the mean-removed metric is unchanged by pressure offsets and matches a NumPy
reference, that the quantities match those of Supplementary Note~9 exactly, that the search gradients agree with
finite differences, that a simulated Wilks limit has its nominal coverage and that the budget is counted correctly.

\paragraph{Mean-removed LDC metric.} In the closed lid-driven cavity the pressure is defined only up to an additive
constant, so a uniform offset in the predicted pressure carries no physics. For every sample we removed the spatial
mean over the $65\times65$ grid from the predicted and from the true pressure separately, and recomputed the joint
relative $L_2$ error over $(p-\bar p,k,|\mathbf{u}|)$ and the pressure error alone. The offset share of the pressure
error $e_p=p_{\rm pred}-p_{\rm true}$ is $n_{\rm g}\bar e_p^{\,2}/\|e_p\|^2$, with $n_{\rm g}=65^2$ grid points and
$\|e_p\|^2=n_{\rm g}\bar e_p^{\,2}+\|e_p-\bar e_p\|^2$. For the 30 main-round LDC models we re-scored the nominal device, all
1000 MC devices, the 200 random corners and $z_{\rm search}$. The recomputed original joint errors matched the
evaluation results to $5\times10^{-6}$ (relative), and the original $R_{200}$ was reproduced in all 30 models.

\paragraph{Results for LDC.} Supplementary Table~\ref{tab:ldcmeanp} summarises them per arm. The searched device's
pressure error was mostly offset: its offset share was 0.48--0.94 (per-arm medians 0.84--0.94), against 0.30--0.94
for the MC device with the largest joint error among 200. Removing the mean, however, also removes most of the true
pressure's norm, which is dominated by its spatial mean. The searched device's relative error therefore changed only
by a factor of 0.93--2.51 (per-arm medians 0.98--1.32), and it grew most for the Mixed-PGD FT and SAM FT models. On
the nominal device the mean-removed error was 1.02--3.15 times the original one, and for the digital FNO it rose from
0.59--0.68\% to 1.04--1.21\%. On the mean-removed metric, $R_{200}$ was 1.07--2.15 and exceeded 1 in all 30 models,
with per-arm medians of 1.15--1.76, against 1.40--2.22 on the original metric. $R_{1000}$ was 0.96--1.93 and
exceeded 1 in 29 of 30 models; the exception is SAM FT seed~4 (0.96). The best of the 200 random corners exceeded
the searched device on this metric in one model (SAM FT seed~2, 5.206\% against 5.148\%). Because $z_{\rm search}$
was found on the original metric, these ratios are lower bounds on what a search on the mean-removed metric would
find. A PGD re-search on that metric for the six seed-0 models, from the eight starts of the main protocol and from
$z_{\rm search}$ (31 evaluations each), raised $R_{1000}$ in all six, from 1.15--1.80 to 1.25--1.83, and
for SAM FT from 1.17 to 1.51. The LDC gap is therefore not an artefact of the arbitrary pressure level.

\paragraph{Quantity-targeted search.} For $Q_p$ and $Q_s$ separately, the search maximised the mean error of the
quantity over the 64 development samples within the 32-D box, with the stages of the main protocol at a reduced
budget of 7000 evaluations per quantity:
\begin{itemize}
\item one DE run of 3328 evaluations (population 8 per dimension, 12 generations);
\item CA from the best corner (1024);
\item PGD from eight starts chosen as in the main protocol but ranked by the quantity's development error
($8\times31=248$ evaluations, 240 with a gradient, with the quantity error differentiated through the model);
\item two CMA-ES runs of 1200, from the jittered centre and from the CA optimum.
\end{itemize}
The main field search uses 16,256 evaluations. The device with the highest development error over these methods (MC excluded, as in the main
protocol) was scored on the 109 held-out samples. For the same quantities we scored the nominal device, all 1000 MC
devices, the 200 random corners and the field-searched device $z_{\rm search}$ of the 30 main-round and nine
pixel-level HX models. The recomputed field errors matched the evaluation results to $6.7\times10^{-6}$, and the
quantities of the nominal, field-searched and MC-maximum devices matched Supplementary Note~9 to $7\times10^{-16}$
(relative).

\paragraph{Wilks limits and thresholds.} The first-order limit $W_1$ is the largest error among the first 59 MC
devices and the second-order limit $W_2$ the second largest among the first 93 (main-text Methods). Both are
distribution-free one-sided 95/95 upper tolerance limits: with probability $1-0.95^{59}=0.952$ (0.950 for $W_2$),
they exceed the 95th percentile of the error over uniformly sampled devices~\citep{wilks1941determination}. In
best-estimate-plus-uncertainty analyses of nuclear safety, such a limit of a code output is compared with an
acceptance criterion~\citep{glaeser2008grs,nutt2004evaluation}. To show how much the limit depends on which 59
devices are drawn, we also computed $W_1$ for each of the 16 disjoint sets of 59 among the 1000 MC devices (and $W_2$
for the 10 sets of 93). As illustrative thresholds we used 1, 2 and 5\% of the test-set mean of the quantity, the
unit of the error. They were set in the earlier engineering-quantity round, before this search was run, and the
counts below depend on them.

\paragraph{Results for the quantity-targeted search.} Supplementary Table~\ref{tab:hxqoisearch} summarises them
per arm, and Supplementary Fig.~\ref{fig:hxwilks} shows every model. For both quantities, the searched error exceeded
the largest over all 1000 MC devices in all 39 models: $R_{1000}$ was 1.20--3.17 ($Q_p$) and 1.22--3.30 ($Q_s$) on
the main round and 1.52--2.02 and 1.49--2.06 on the pixel-level device. It also exceeded the best of the 200 random
corners in every model, although the corners came within 6\% for the main-round Random-error-aware models (ratio
1.03--1.06) and the ratio fell to 1.00 for one box-trained model. The searched error was 1.40--4.63 ($Q_p$) and
1.39--6.45 ($Q_s$) times $W_1$ on the main round and 1.88--4.22 and 1.78--4.16 times on the pixel-level device, and
it exceeded $W_1$ of all 16 sets in every model. This is implied by the first result: each $W_1$ is the largest
error among a subset of the 1000 MC devices, so a searched error above the MC maximum is necessarily above it. For
Box seed~0, whose field-searched device had a $Q_p$ error of only 0.46 times the MC maximum (Supplementary Note~9),
the $Q_p$-targeted search reached 0.8725\%, 2.45 times the MC maximum and 5.36 times the field-searched device.

The informative comparison is with an error threshold. We count the splits in which the Wilks test passes ($W_1$ at
or below the threshold) while the searched error exceeds the threshold. They are of two kinds. In the first, no
random device exceeded the threshold either. At 1\% of the mean $Q_p$, $W_1$ was 0.289--0.588\% and the searched
error 1.01--1.76\% for 11 box-trained models (Box seed~1, Smooth-mask box seeds 0--4, Box FT seed~4, SAM FT seed~4
and the three pixel-level Box models), and none of their 1000 MC devices exceeded 1\%. At 2\% of the mean $Q_s$, the
same held for seed~2 of the pixel-level Random-error-aware models ($W_1$ 1.129\%, searched 2.01\%; no MC device above
2\%). For these 12 model--quantity pairs, random testing with 1000 devices gave no warning of an exceedance that the
search found. In the second kind, random testing did flag exceedances. At 5\% of the mean $Q_p$, the five
Random-error-aware main-round models passed the Wilks test ($W_1$ 4.510--4.922\%; $W_2$ 4.355--4.922\%), whereas
their $Q_p$-searched devices reached 6.42--6.87\% (398--425\,Pa). Between 0.1\% and 0.8\% of their 1000 MC devices
exceeded 5\% (MC maximum 5.135--5.568\%), consistent with a 95th percentile below 5\%. Over the 16 sets, $W_1$ ranged
from 4.190--4.525\% (lowest set) to 5.135--5.568\% (highest set), so some draws of 59 devices would have failed the
test. At 2\% of the mean $Q_p$, the three pixel-level Random-error-aware models were similar ($W_1$ 1.785--1.889\%,
searched 3.37--3.61\%, 0.4--2.6\% of the MC devices above 2\%). In these eight cases the Wilks test worked as
designed: it bounds the 95th percentile, and the worst case lies above it. In all of the split cases $W_2$ also
passed. The opposite split (the Wilks test fails while the searched error passes) cannot occur, because the
searched error exceeded the MC maximum and hence $W_1$.

The quantity-targeted search found larger errors than the field-searched device in 77 of 78 model--quantity pairs,
with per-arm medians of 1.01--3.00 and per-model ratios up to 9.08. The exception is Box FT seed~2 on $Q_p$ (0.7268\%
against 0.7556\%, ratio 0.96), where the field search, with more than twice the budget, found a device with a
slightly larger $Q_p$ error; both are lower bounds on the worst case. For the main-round Random-error-aware models the
two searches gave nearly the same error (ratio 1.00--1.02), and CA found the quantity-searched device of all eight
Random-error-aware models. Over all 78 pairs, the device came from CA in 29 and from PGD in 49 (24 of them from the
jittered centre), and never from corners, DE or CMA-ES alone. The development error of the searched device was
0.92--1.20 times its held-out error.

\clearpage
\section*{Supplementary Tables}
\addcontentsline{toc}{section}{Supplementary Tables}
\begin{table}[!ht]
\centering
\caption{\textbf{Primary endpoint, Darcy and NS} (\DET, static draw 0, held-out samples, error in \%). MC: 1000
devices, maximum over the first 200 and over all 1000. Unseen: best of corners, CA and CMA-ES. DE/PGD: best of DE
and PGD. Searched: the searched device $z_{\rm search}$, the device with the highest development error over all
searches. Method identifies the search that found it: A = CA, K = corners, Pd/Pc/Pa/Pk/Pm/Pr = PGD from DE,
jittered centre, CA, best corner, best MC device or random start, C0, C1, \dots\ = CMA-ES runs. No searched device in these tables came from DE alone.}
\label{tab:models_a}
\scriptsize\setlength{\tabcolsep}{3.5pt}
\input{tables/tabS_models_a.tex}
\end{table}
\begin{table}[tbp]
\centering
\caption{\textbf{Primary endpoint, LDC and HX}, defined as in Supplementary Table~\ref{tab:models_a}. HX error is the
mean of the per-channel errors.}
\label{tab:models_b}
\scriptsize\setlength{\tabcolsep}{3.5pt}
\input{tables/tabS_models_b.tex}
\end{table}
\begin{table}[tbp]
\centering
\caption{\textbf{Secondary endpoint and reliability, Darcy and NS.} \REAL\ configuration (\%): nominal error (20
static draws), maximum of 200 MC devices with independent static draws, and $z_{\rm search}$ re-scored on 20 static
draws (mean and maximum). Nested: PGD over eight common static draws, re-scored on 20 fresh draws. Reliability on
\DET: number of the 1000 MC devices with error above $c$ times nominal, and the 95\% upper confidence bound on the
0.99 quantile (\%).}
\label{tab:real_a}
\scriptsize\setlength{\tabcolsep}{3.2pt}
\input{tables/tabS_real_a.tex}
\end{table}
\begin{table}[tbp]
\centering
\caption{\textbf{Secondary endpoint and reliability, LDC and HX}, as in Supplementary Table~\ref{tab:real_a}.}
\label{tab:real_b}
\scriptsize\setlength{\tabcolsep}{3.2pt}
\input{tables/tabS_real_b.tex}
\end{table}
\begin{table}[tbp]
\centering
\caption{\textbf{Budget-matched samplers} (seed-0 models, mode-level device, \DET, static draw 0, held-out samples,
error in \%). Each sampler draws $N_{\rm match}=16{,}256$ devices and selects the one with the highest development
error. Error: held-out error of that device. $R_{\rm match}$: searched error divided by it. Corners + uniform: the
mixed sampler, which alternates corner and uniform devices (Supplementary Note~3).}
\label{tab:v3e1}
\scriptsize\setlength{\tabcolsep}{3.5pt}
\input{tables/tabS_v3_e1.tex}
\end{table}
\begin{table}[tbp]
\centering
\caption{\textbf{100,000 uniform devices and generalised Pareto tail fits} (Box seed 0, mode-level device, \DET,
development errors in \%). Best development error among the first $10^3$, $10^4$ and $10^5$ devices; searched:
development error of $z_{\rm search}$; $\ge$ searched: devices of $10^5$ that reached it; $R$ at $10^5$: held-out
error of $z_{\rm search}$ divided by that of the device selected from $10^5$. Tail: shape $\xi$ (threshold at the
0.99 quantile), $N^*$ at three thresholds (infinite when the fitted tail ends below the searched error) and the 5th
percentile of $N^*$ over 200 bootstrap resamples (0.99 quantile). The tail values are extrapolations of a fitted
law (Supplementary Note~3).}
\label{tab:v3mc}
\scriptsize\setlength{\tabcolsep}{3.5pt}
\input{tables/tabS_v3_mc.tex}
\end{table}
\begin{table}[tbp]
\centering
\caption{\textbf{Pixel-level device} (\DET, static draw 0, held-out samples, error in \%; HX: mean of the
per-channel errors). HX and Darcy: retrained models with the cell-integral shift; Darcy, sinc: the same arms and
seeds retrained with the band-limited shift; Transfer: seed-0 models trained on the mode-level device and evaluated
on the pixel-level device (cell integral). Nominal: the nominal device (0.5-pixel fringing, 30\,dB), not the box
centre. $R_{\rm match}$: as in Supplementary Table~\ref{tab:v3e1} for the uniform, corner and mixed samplers. Nom./MC
mean: nominal error divided by the MC mean. Searched device: layers at the sharpest fringing / layers at the lowest
extinction ratio / shift coordinates within 0.05 of zero (of 8).}
\label{tab:v3e2}
\scriptsize\setlength{\tabcolsep}{2.6pt}
\input{tables/tabS_v3_e2.tex}
\end{table}
\begin{table}[tbp]
\centering
\caption{\textbf{Shared-knob 8-D box} (seed 0, the same eight knobs in all four layers, \DET, held-out samples,
error in \%). Method codes as in Supplementary Table~\ref{tab:models_a}.}
\label{tab:shared}
\scriptsize
\input{tables/tabS_shared.tex}
\end{table}
\begin{table}[tbp]
\centering
\caption{\textbf{Tolerance sweep} (seed 0, half-widths scaled by $\kappa$, reduced protocol, \DET, held-out samples,
error in \%). Nominal errors are as in Supplementary Tables~\ref{tab:models_a} and \ref{tab:models_b}.}
\label{tab:sweep}
\scriptsize
\input{tables/tabS_sweep.tex}
\end{table}
\begin{table}[tbp]
\centering
\caption{\textbf{Circular-phase crosstalk set.} Retrained models, seeds 0--2, with a reduced search of one DE seed
and one CMA-ES seed from the centre (\DET, held-out samples, error in \%). Unseen: best of the unseen attacks.}
\label{tab:cp}
\scriptsize
\input{tables/tabS_cp.tex}
\end{table}
\begin{table}[tbp]
\centering
\caption{\textbf{Paired values behind main-text Fig.~4e.} Searched error and best unseen-attack error of Box FT and
Mixed-PGD FT for every task and seed, and the change of Mixed-PGD FT relative to Box FT. All values are on \DET\ and
held-out samples (\%), and the HX error is the mean of the per-channel errors.}
\label{tab:paired}
\scriptsize
\input{tables/tabS_paired.tex}
\end{table}
\begin{table}[tbp]
\centering
\caption{\textbf{Heat exchanger: per-channel and joint errors} (\%, mean over seeds, held-out samples, \DET). MC
(dev-selected): the MC device selected by development error from all 1000 MC devices. Joint median: joint relative
$L_2$ error over all channels, median over samples.}
\label{tab:hx}
\scriptsize
\input{tables/tabS_hx.tex}
\end{table}
\begin{table}[tbp]
\centering
\caption{\textbf{Certificate bound across box scales $\kappa$} (v2 simulator, Darcy seed 0, \DET, static draw 0,
mean over ten test inputs). Expansion-point error: error at the Taylor model's expansion point, which is not the
nominal device and moves with $\kappa$ (Supplementary Note~6). Worst found: per-input search, floored by the sampled
devices.}
\label{tab:cert}
\scriptsize
\input{tables/tabS_cert.tex}
\end{table}
\begin{table}[tbp]
\centering
\caption{\textbf{v1 pilot effect decomposition} (legacy simulator): each non-ideality alone at its magnitude in the
v1 \textsc{realistic} configuration, the v1 counterpart of \REAL\ (error in \%, random effects averaged over 20 frozen devices).}
\label{tab:decomp}
\scriptsize
\input{tables/tabS_decomp.tex}
\end{table}
\begin{table}[tbp]
\centering
\caption{\textbf{Transfer of adapters calibrated on one simulated chip to other simulated chips} (frozen-library
pilot, v1 simulator, NS unless noted, relative $L_2$ error in \%, penalty in percentage points;
$N_\mathrm{train}$: number of training samples). Pre-mix adapter: the adapter adds a digital pre-mix before each optical pass;
slim adapter: a reduced adapter with a diagonal skip and a 32-wide projection; 16 blocks: 16 optical blocks, four per
layer, instead of one shared bank of four; phase-only: phase-only masks.}
\label{tab:port}
\scriptsize\setlength{\tabcolsep}{4pt}
\input{tables/tabS_portability.tex}
\end{table}
\begin{table}[tbp]
\centering
\caption{\textbf{Per-field LDC errors} (\DET, static draw 0, held-out samples, relative $L_2$ error in \%, means over
five seeds; Digital: three seeds, ideal hardware). nom.: nominal device; MC: the MC device with the largest joint
error among 200, that is, the device that sets the maximum in $R_{200}$; search: the searched device. The joint error, the LDC metric,
equals the pressure error to 0.001 percentage points in every model and device. The turbulent kinetic energy (TKE)
$k$ and the speed $|\mathbf{u}|$ are predicted far less accurately by all models, including the digital FNO. The
searched device, selected on the joint error, raised the $k$ error above that of the MC device with the largest
joint error in 26 of 30 models and the $|\mathbf{u}|$ error in 24.}
\label{tab:ldcfields}
\scriptsize\setlength{\tabcolsep}{4pt}
\input{tables/tabS_ldc_fields.tex}

\end{table}
\begin{table}[tbp]
\centering
\caption{\textbf{Correlated tolerance box} (seed 0, mode-level device, \DET, static draw 0, held-out samples, error
in \%). Crosstalk and extinction are shared by the four layers and the other six knobs are independent per layer
($D=26$; Supplementary Note~9). MC: 1000 devices, maximum over the first 200 and over all 1000. Shared at
$z_{\rm search}$: crosstalk width (pixel) and extinction ratio (dB) of the searched device. $R_{200}$, other boxes:
the same model in the independent 32-D box (Supplementary Tables~\ref{tab:models_a} and \ref{tab:models_b}) and in
the shared 8-D box (Supplementary Table~\ref{tab:shared}; not run for HX). Method codes as in Supplementary
Table~\ref{tab:models_a}.}
\label{tab:hybrid}
\scriptsize\setlength{\tabcolsep}{3.5pt}
\input{tables/tabS_hybrid.tex}
\end{table}
\begin{table}[tbp]
\centering
\caption{\textbf{Heat exchanger: engineering quantities} (\DET, static draw 0, held-out samples). $Q_p$: slice-mean
gauge pressure; $Q_s$: root-mean-square in-plane speed; $Q_{\max}$: 99th percentile of the in-plane speed
(Supplementary Note~9). Errors are $|Q_{\rm pred}-Q_{\rm true}|$ in \% of the test-set mean of $|Q_{\rm true}|$,
averaged over the held-out samples, and are means over seeds for the nominal device (nom.), the largest over all
1000 MC devices (MC max) and the searched device. Ratio: searched error divided by the MC maximum, range over seeds
(All: median). $n_>$: models in which the searched error exceeds the MC maximum. The search maximises the field
error, not these quantities.}
\label{tab:hxqoi}
\scriptsize\setlength{\tabcolsep}{2.6pt}
\input{tables/tabS_hx_qoi.tex}
\end{table}
\begin{table}[tbp]
\centering
\caption{\textbf{LDC with the pressure offset removed} (\DET, static draw 0, held-out samples; Supplementary
Note~10). The spatial mean of the pressure is removed from each predicted and true field separately. $R_{200}$:
original metric (Supplementary Table~\ref{tab:models_b}). $R_{200}^{\rm mr}$, $R_{1000}^{\rm mr}$: searched error
divided by the MC maximum of 200 and 1000 devices on the mean-removed joint error; $R_{200}^{p,\rm mr}$: on the
mean-removed pressure error alone. Values are medians over seeds; $>1$: models with the ratio above 1. Offset share:
share of the squared norm of the searched device's pressure error that lies in its spatial mean. $e^{\rm mr}/e$:
mean-removed over original joint error of the searched device. Seed 0: $R_{1000}^{\rm mr}$ of $z_{\rm search}$,
found on the original metric, and of a PGD re-search on the mean-removed metric (start codes as in Supplementary
Table~\ref{tab:models_a}; Pz: from $z_{\rm search}$). Range: minimum to maximum over the 30 models.}
\label{tab:ldcmeanp}
\scriptsize\setlength{\tabcolsep}{3.5pt}
\input{tables/tabS_ldc_meanp.tex}
\end{table}
\begin{table}[tbp]
\centering
\caption{\textbf{Heat exchanger: quantity-targeted search and Wilks 95/95 limits} (\DET, static draw 0, held-out
samples; Supplementary Note~10). The search maximises the development error of $Q_p$ or $Q_s$ (7000 evaluations
each). $R_{1000}$: searched error divided by the largest over all 1000 MC devices; search$/W_1$: divided by the
first-order Wilks limit, the largest of the first 59 MC devices; $R_{\rm field}$: divided by the error of the same
quantity on the field-searched device $z_{\rm search}$. Values are medians over seeds. $n_{>\rm MC}$: models whose
searched error exceeds the MC maximum; $n_{>W_1}$: models whose searched error exceeds $W_1$ of all 16 disjoint sets
of 59 MC devices. 1/2/5\%: models in which $W_1$ is at or below a threshold of 1, 2 or 5\% of the mean quantity while
the searched error exceeds it.}
\label{tab:hxqoisearch}
\scriptsize\setlength{\tabcolsep}{3pt}
\input{tables/tabS_hx_qoi_search.tex}
\end{table}

\clearpage
\section*{Supplementary Figures}
\addcontentsline{toc}{section}{Supplementary Figures}
\begin{figure}[!ht]
\centering
\includegraphics[width=\linewidth]{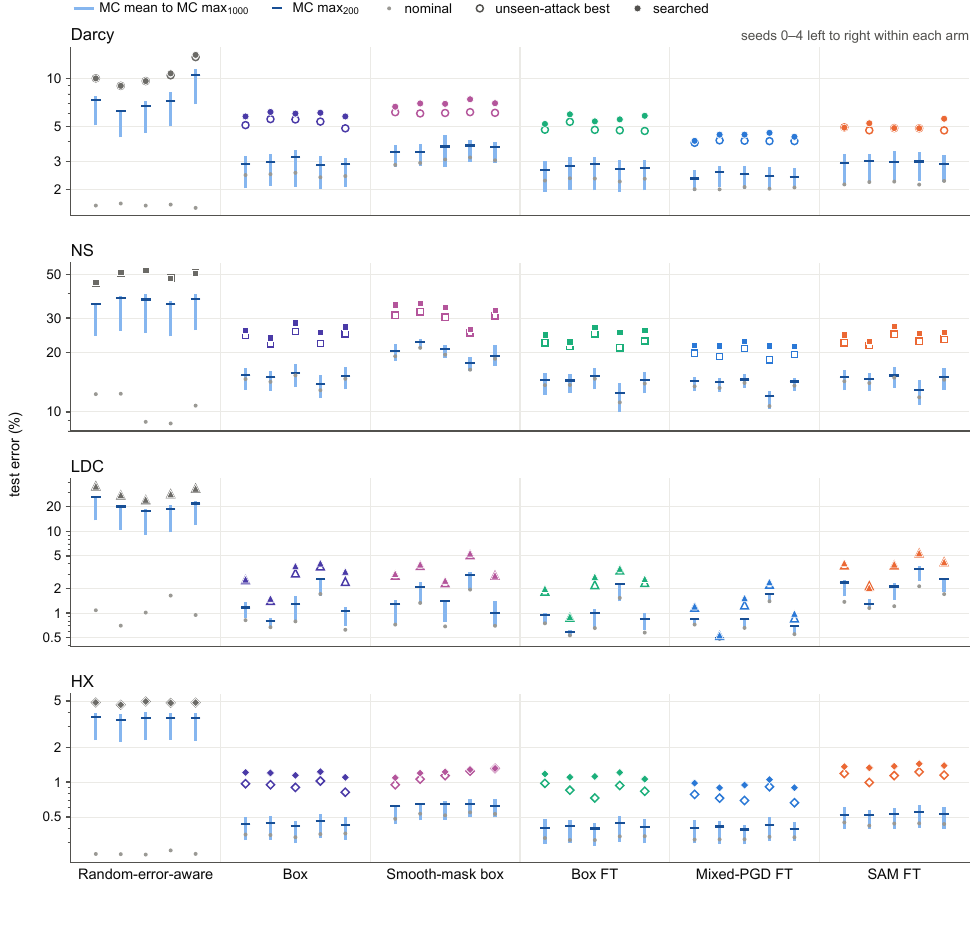}
\caption{\textbf{All 120 main models} (\DET, held-out samples): nominal error, MC mean to MC maximum of 1000 (line),
MC maximum of 200 (bar), best unseen-attack device (open) and searched device (filled).}
\label{fig:s1}
\end{figure}
\begin{figure}[tbp]
\centering
\includegraphics[width=\linewidth]{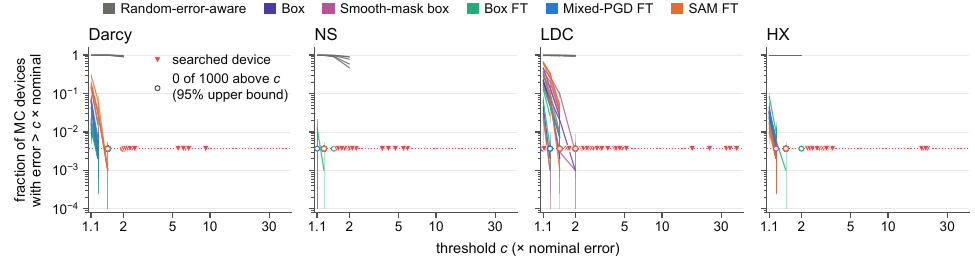}
\caption{\textbf{Exceedance curves of the 1000 MC devices.} Fraction of the 1000 MC devices whose error exceeds $c$
times the nominal error, one line per model (\DET, held-out samples), with pointwise two-sided 95\% Clopper--Pearson
intervals. Open circles mark the upper end of the interval (`95\% upper bound' in the legend) where no device
exceeds the threshold. Red triangles
mark $c$ = searched error / nominal error. They sit at that upper end (0.37\%, dotted line; main-text Methods) because no MC
device reached the searched error in any model. Grey lines at the top are Random-error-aware models.}
\label{fig:exc}
\end{figure}
\begin{figure}[tbp]
\centering
\includegraphics[width=\linewidth]{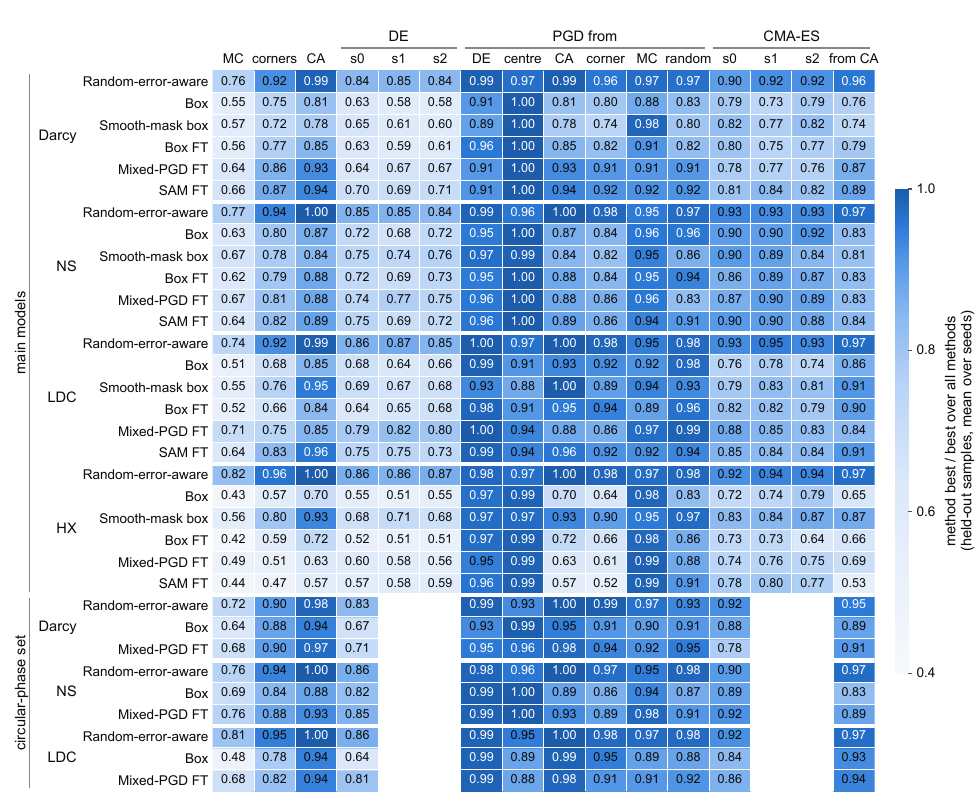}
\caption{\textbf{Error found by each search method relative to the best across methods.} Best held-out error of the
method divided by the best over all methods, mean over seeds, per task and arm (rows), including the circular-phase
set (blank: not run).}
\label{fig:s2}
\end{figure}
\begin{figure}[tbp]
\centering
\includegraphics[width=\linewidth]{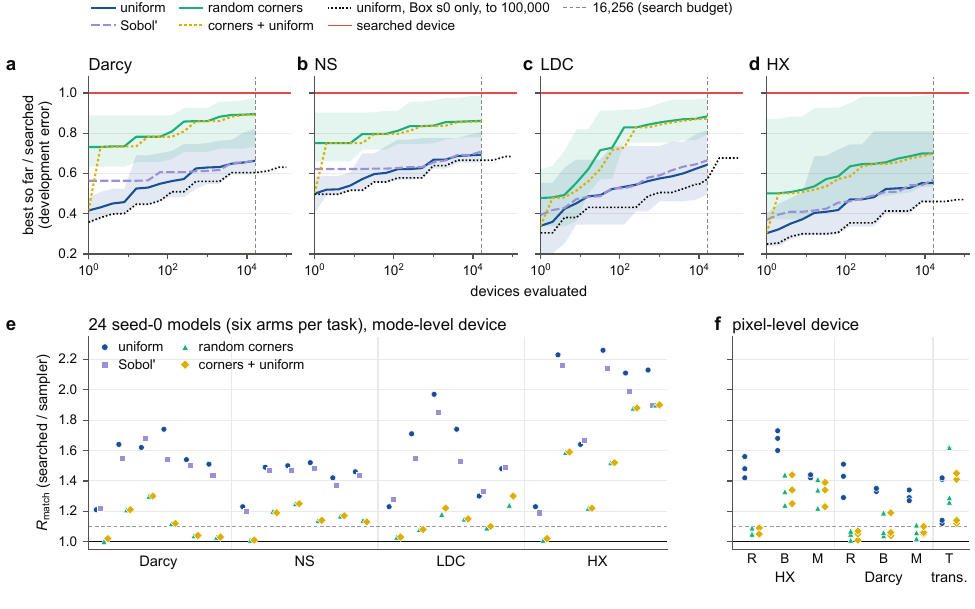}
\caption{\textbf{Budget-matched random sampling.} \textbf{a--d},~Best development error found so far by each
sampler (corners + uniform: the mixed sampler), divided by the searched device's development error, against the number of devices (seed-0 models,
mode-level device). Lines are means over the six arms of a task, and bands the range over arms for the uniform and
corner samplers. The dotted line is the uniform sampler of the Box model alone, continued to 100,000 devices, and
the dashed vertical line marks the search budget. \textbf{e},~$R_{\rm match}$ of each sampler for the 24 models, six arms per
task in the order of main-text Fig.~4. \textbf{f},~$R_{\rm match}$ of the uniform, corner and mixed samplers on the
pixel-level device (R, Random-error-aware; B, Box; M, Mixed-PGD FT; three seeds each) and for the four transferred
models (T). Dashed line: $R_{\rm match}=1.1$.}
\label{fig:v3budget}
\end{figure}
\begin{figure}[tbp]
\centering
\includegraphics[width=0.6\linewidth]{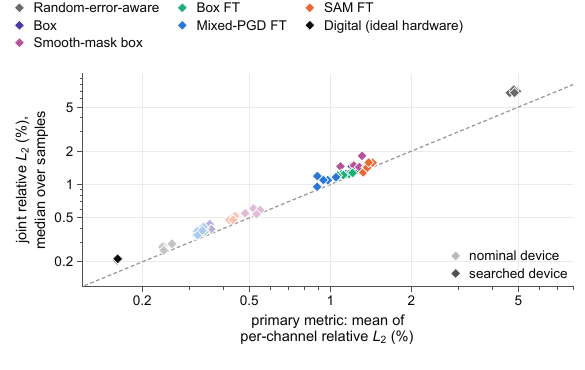}
\caption{\textbf{HX: joint error against the primary metric.} Joint relative $L_2$ error (median over held-out
samples) against the primary metric (mean of the per-channel relative $L_2$ errors) for every HX model on the
nominal (faint) and searched (solid) device, and for the digital FNO on ideal hardware. Both axes are logarithmic, and the dashed line marks equality.}
\label{fig:hxmetric}
\end{figure}
\begin{figure}[tbp]
\centering
\includegraphics[width=\linewidth]{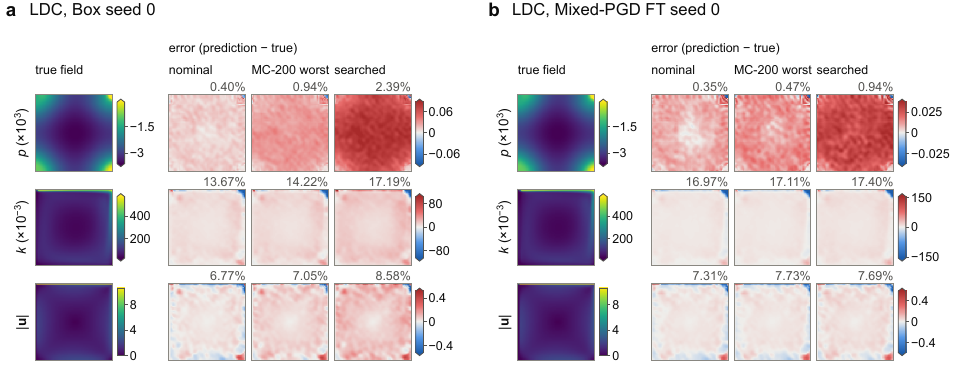}
\caption{\textbf{LDC field snapshot} (mode-level device, \DET, held-out sample with the median nominal error of the
Box seed-0 model). \textbf{a},~Box seed~0. \textbf{b},~Mixed-PGD FT seed~0. True fields and errors (prediction minus
true field) on the nominal device, on the worst of the first 200 MC devices (selected by development error) and on the
searched device; the lid is at the top. Rows: pressure $p$, turbulent kinetic energy $k$ and velocity magnitude
$|\mathbf{u}|$ (Supplementary Note~3). The pressure dominates the joint error used as the LDC metric (joint errors,
equal to the pressure errors shown: 0.40\%, 0.94\% and 2.39\% for Box; 0.35\%, 0.47\% and 0.94\% for Mixed-PGD FT).
Error maps of a field share a symmetric colour scale, clipped at the 99.5th
percentile of the absolute error, and give that field's relative $L_2$ error. True fields span their 0.5th to
99.5th percentiles. Colour-bar values are to be multiplied by the power of ten in the row label.}
\label{fig:fieldsldc}
\end{figure}
\begin{figure}[tbp]
\centering
\includegraphics[width=\linewidth]{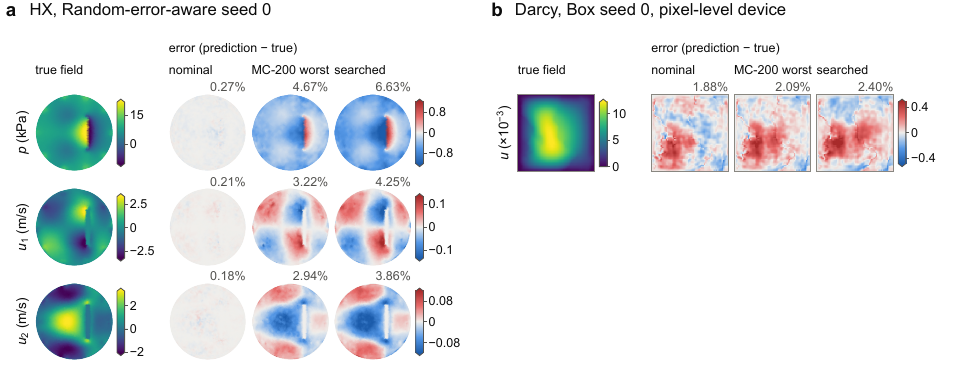}
\caption{\textbf{Further field snapshots}, as in main-text Fig.~5b. \textbf{a},~HX Random-error-aware seed~0
(mode-level device; mean of the per-channel errors 0.22\%, 3.61\% and 4.91\%). Its errors on the MC and searched
devices are smooth and large-scale, unlike the fine-grained errors of the Box model in main-text Fig.~5b.
\textbf{b},~Darcy Box seed~0 on the pixel-level device (cell integral): the error grows in the same region on all
three devices (1.88\%, 2.09\% and 2.40\%). Colour scales as in Supplementary Fig.~\ref{fig:fieldsldc}.}
\label{fig:fieldsother}
\end{figure}
\begin{figure}[tbp]
\centering
\includegraphics[width=0.6\linewidth]{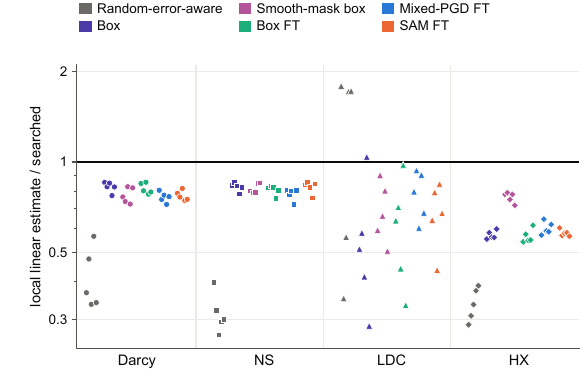}
\caption{\textbf{Ratio of the local linear estimate to the searched error} (full test set, \DET).}
\label{fig:s3}
\end{figure}
\begin{figure}[tbp]
\centering
\includegraphics[width=\linewidth]{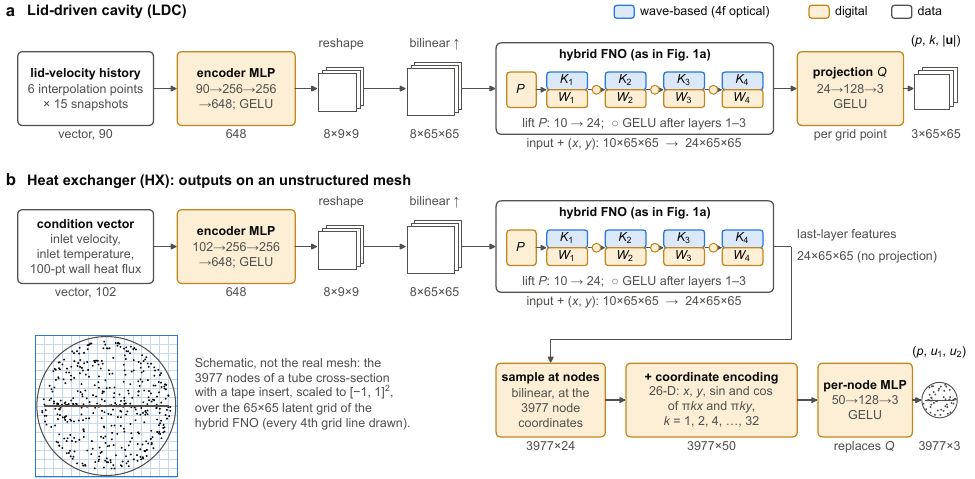}
\caption{\textbf{Architectures of the two tasks with vector inputs.} Tensor shapes are channels\,$\times$\,height\,$\times$\,width
on grids and nodes\,$\times$\,features on the mesh. Blue: wave-based (4f optical) paths; amber: digital.
\textbf{a},~LDC. The 90-dimensional encoding of the lid-velocity history passes through a digital MLP
($90\to256\to256\to8\times9\times9$, GELU after the first two maps), is reshaped to an $8\times9\times9$ latent field
and is upsampled bilinearly to $8\times65\times65$. The hybrid FNO of main-text Fig.~1a appends two grid coordinates,
lifts to 24 channels and applies the four hybrid layers. The projection ($24\to128\to3$) gives $(p,k,|\mathbf{u}|)$ on the
$65\times65$ grid. \textbf{b},~HX. The 102-dimensional condition vector passes through the same encoder onto a
$65\times65$ latent grid over $[-1,1]^2$. The 24 features of the last hybrid layer are sampled bilinearly at the 3977
node coordinates and concatenated with a 26-dimensional coordinate encoding. A per-node digital MLP ($50\to128\to3$)
replaces the projection and gives the pressure and the two in-plane velocities $(p,u_1,u_2)$ at the nodes. The mesh
inset is a schematic, not the real node coordinates.}
\label{fig:arch}
\end{figure}
\begin{figure}[tbp]
\centering
\includegraphics[width=0.6\linewidth]{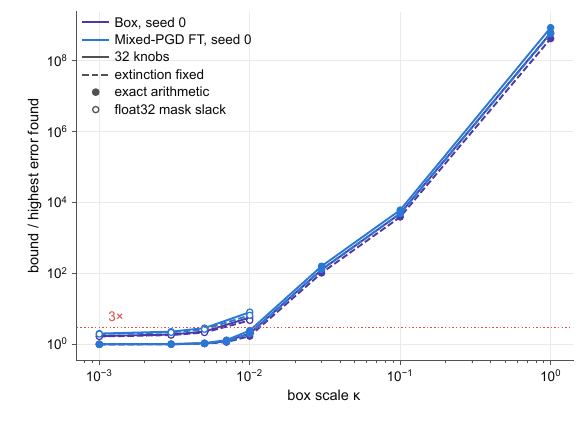}
\caption{\textbf{Certificate bound divided by the highest error found}, against the box scale $\kappa$ (v2
simulator, Darcy seed 0). Line style: 32 knobs or extinction fixed. Markers: exact arithmetic or float32 mask slack.
Dotted line: 3$\times$.}
\label{fig:s4}
\end{figure}
\begin{figure}[tbp]
\centering
\includegraphics[width=0.55\linewidth]{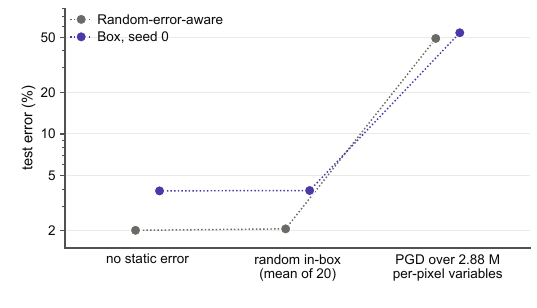}
\caption{\textbf{v1 pilot: per-pixel static errors} (Darcy) for the v1 Random-error-aware and Box seed-0 models.
Error without static errors, mean over 20 random draws inside the per-pixel box, and the error after gradient search
over all 2.88 million per-pixel variables (legacy simulator).}
\label{fig:s5}
\end{figure}
\begin{figure}[tbp]
\centering
\includegraphics[width=0.6\linewidth]{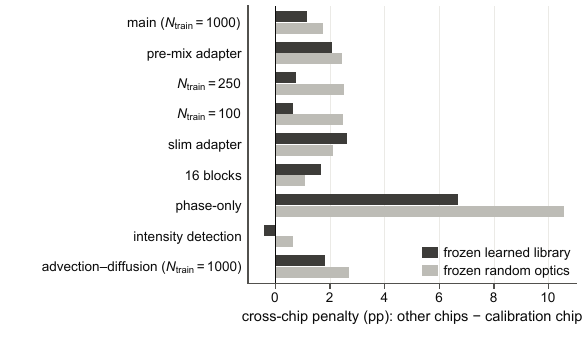}
\caption{\textbf{Cross-chip penalty when adapters calibrated on one simulated chip are used on others}, for a learned
frozen library and for random frozen optics (frozen-library pilot, v1 simulator; $N_\mathrm{train}$: number of
training samples).}
\label{fig:s6}
\end{figure}
\begin{figure}[tbp]
\centering
\includegraphics[width=\linewidth]{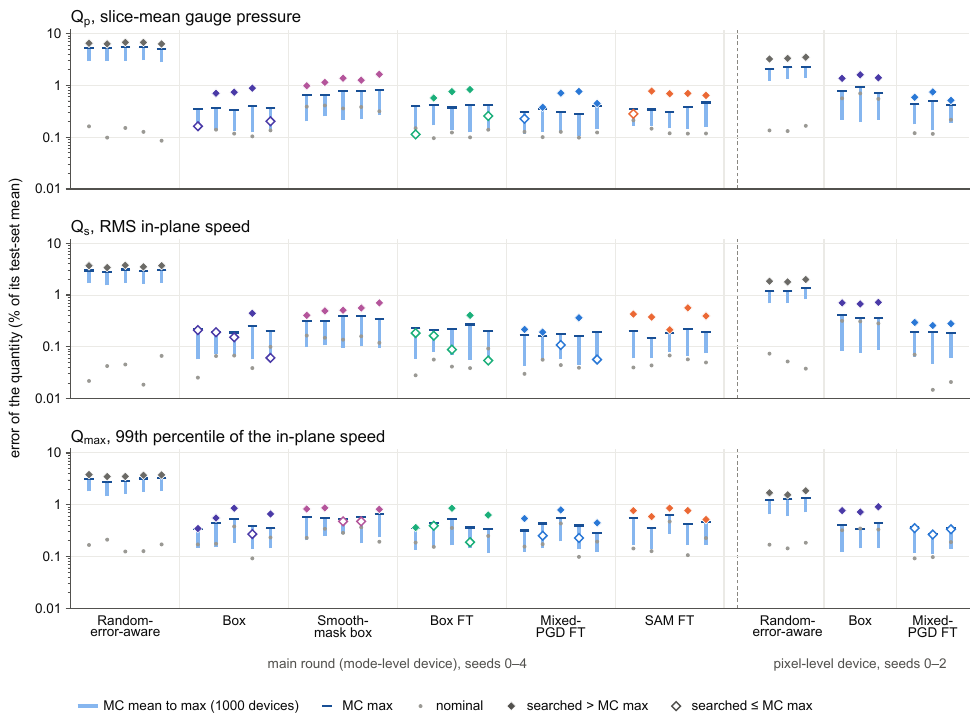}
\caption{\textbf{Heat exchanger: errors of three engineering quantities} for all 39 HX models (\DET, static draw 0,
held-out samples; Supplementary Note~9). Error of the quantity in \% of its test-set mean on the nominal device, from
the mean to the maximum over all 1000 MC devices, and on the searched device, which maximises the field error.
Filled diamonds: the searched error exceeds the MC maximum; open diamonds: it does not. Colour gives the arm. The
$y$-axes are logarithmic.}
\label{fig:hxqoi}
\end{figure}
\begin{figure}[tbp]
\centering
\includegraphics[width=\linewidth]{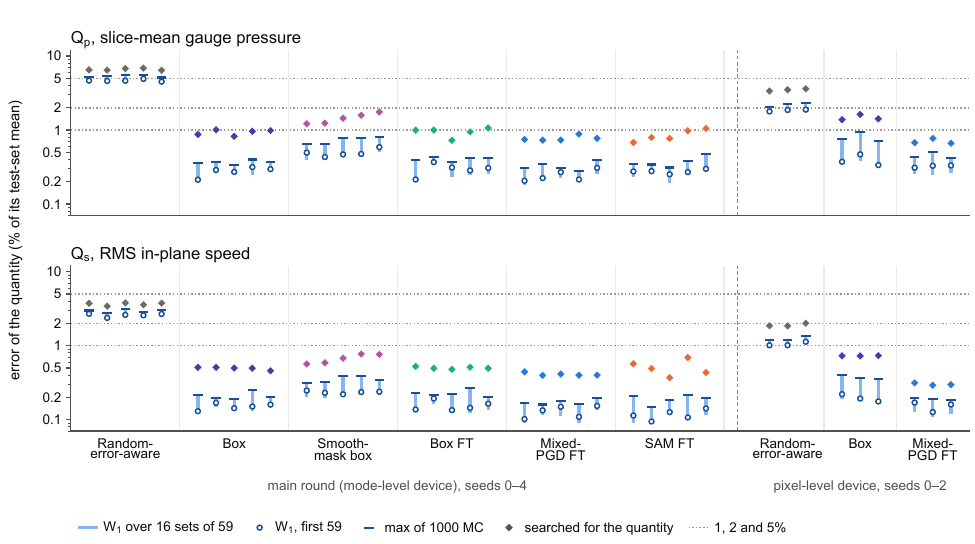}
\caption{\textbf{Heat exchanger: quantity-targeted search against Wilks 95/95 limits} for all 39 HX models (\DET,
static draw 0, held-out samples; Supplementary Note~10). Error of the quantity in \% of its test-set mean: the
first-order Wilks limit $W_1$ from the first 59 MC devices (open circles) and its range over the 16 disjoint sets of
59 (bars), the maximum over all 1000 MC devices (dashes) and the device found by the search for that quantity
(diamonds; colour gives the arm). Dotted lines: thresholds of 1, 2 and 5\%. The $y$-axes are logarithmic.}
\label{fig:hxwilks}
\end{figure}

\clearpage
\bibliographystyle{unsrtnat}
\bibliography{references}

\clearpage
\endgroup
\fi

\end{document}

%% file: tables/tab_arms.tex
\begin{tabular}{llrrrrrl}
\toprule
Task & Arm & Nominal & MC max$_{200}$ & Unseen-attack best & Searched (mean $\pm$ s.d.) & $R_{200}$ & $\Delta$ vs Box FT (\%) \\
\midrule
Darcy & Random-error-aware & 1.58 & 7.58 & 10.6 & 10.7 $\pm$ 2.0 & 1.35–1.49 & +92 [+51, +142] \\
 & Box & 2.46 & 2.95 & 5.27 & 5.96 $\pm$ 0.18 & 1.88–2.14 & +7 [$-$1, +12] \\
 & Smooth-mask box & 3.02 & 3.62 & 6.09 & 6.99 $\pm$ 0.27 & 1.86–2.03 & +26 [+17, +34] \\
 & Box FT & 2.31 & 2.76 & 4.84 & 5.57 $\pm$ 0.32 & 1.85–2.12 & -- \\
 & \textbf{Mixed-PGD FT} & 2.03 & 2.45 & 4.02 & 4.35 $\pm$ 0.19 & 1.71–1.88 & $-$22 [$-$26, $-$18] \\
 & SAM FT & 2.20 & 2.96 & 4.81 & 5.09 $\pm$ 0.32 & 1.62–1.93 & $-$9 [$-$12, $-$4] \\
 & Digital FNO (3 seeds) & 2.04 & -- & -- & -- & -- & -- \\
\addlinespace
NS & Random-error-aware & 10.6 & 36.7 & 49.5 & 49.5 $\pm$ 2.8 & 1.29–1.41 & +98 [+84, +125] \\
 & Box & 14.4 & 15.0 & 23.9 & 26.1 $\pm$ 1.7 & 1.59–1.84 & +4 [0, +6] \\
 & Smooth-mask box & 19.0 & 20.3 & 29.9 & 32.8 $\pm$ 3.8 & 1.48–1.71 & +31 [+4, +57] \\
 & Box FT & 13.4 & 14.2 & 22.6 & 25.1 $\pm$ 1.6 & 1.57–2.03 & -- \\
 & \textbf{Mixed-PGD FT} & 13.0 & 13.9 & 19.6 & 21.9 $\pm$ 0.50 & 1.51–1.80 & $-$13 [$-$17, $-$4] \\
 & SAM FT & 13.9 & 14.6 & 23.1 & 25.1 $\pm$ 1.6 & 1.57–1.94 & 0 [$-$3, +2] \\
 & Digital FNO (3 seeds) & 5.55 & -- & -- & -- & -- & -- \\
\addlinespace
LDC & Random-error-aware & 1.08 & 21.0 & 30.0 & 30.1 $\pm$ 4.5 & 1.36–1.53 & +1470 [+714, +2991] \\
 & Box & 0.925 & 1.38 & 2.65 & 3.04 $\pm$ 1.0 & 1.59–3.07 & +34 [+16, +67] \\
 & Smooth-mask box & 1.08 & 1.74 & 3.42 & 3.56 $\pm$ 1.2 & 1.77–2.89 & +89 [$-$10, +337] \\
 & Box FT & 0.810 & 1.13 & 2.13 & 2.37 $\pm$ 0.98 & 1.53–3.10 & -- \\
 & \textbf{Mixed-PGD FT} & 0.763 & 0.919 & 1.22 & 1.34 $\pm$ 0.70 & 1.08–1.83 & $-$44 [$-$62, $-$32] \\
 & SAM FT & 1.52 & 2.38 & 3.90 & 3.96 $\pm$ 1.2 & 1.55–1.88 & +78 [+42, +126] \\
 & Digital FNO (3 seeds) & 0.634 & -- & -- & -- & -- & -- \\
\addlinespace
HX & Random-error-aware & 0.243 & 3.55 & 4.82 & 4.82 $\pm$ 0.11 & 1.34–1.39 & +326 [+298, +356] \\
 & Box & 0.351 & 0.438 & 0.931 & 1.18 $\pm$ 0.053 & 2.57–2.80 & +4 [+1, +9] \\
 & Smooth-mask box & 0.523 & 0.635 & 1.14 & 1.22 $\pm$ 0.086 & 1.77–2.10 & +8 [$-$7, +24] \\
 & Box FT & 0.329 & 0.414 & 0.865 & 1.13 $\pm$ 0.059 & 2.59–2.95 & -- \\
 & \textbf{Mixed-PGD FT} & 0.327 & 0.406 & 0.756 & 0.952 $\pm$ 0.066 & 2.16–2.44 & $-$16 [$-$19, $-$13] \\
 & SAM FT & 0.437 & 0.531 & 1.14 & 1.38 $\pm$ 0.041 & 2.55–2.61 & +22 [+15, +31] \\
 & Digital FNO (3 seeds) & 0.162 & -- & -- & -- & -- & -- \\
\bottomrule
\end{tabular}

%% file: tables/tabS_models_a.tex
\begin{tabular}{llrrrrrrrrrl}
\toprule
Task & Arm, seed & nominal & MC mean & max$_{200}$ & max$_{1000}$ & unseen & DE/PGD & searched & $R_{200}$ & $R_{1000}$ & method \\
\midrule
Darcy & Random-error-aware s0 & 1.58 & 5.07 & 7.28 & 7.72 & 10.0 & 10.0 & 10.0 & 1.38 & 1.30 & Pd \\
Darcy & Random-error-aware s1 & 1.63 & 4.25 & 6.23 & 6.38 & 9.00 & 9.00 & 9.00 & 1.44 & 1.41 & A \\
Darcy & Random-error-aware s2 & 1.58 & 4.55 & 6.69 & 7.19 & 9.65 & 9.65 & 9.65 & 1.44 & 1.34 & A \\
Darcy & Random-error-aware s3 & 1.60 & 5.02 & 7.25 & 8.25 & 10.5 & 10.8 & 10.8 & 1.49 & 1.31 & Pd \\
Darcy & Random-error-aware s4 & 1.53 & 6.90 & 10.5 & 11.4 & 13.7 & 14.1 & 14.1 & 1.35 & 1.23 & Pd \\
Darcy & Box s0 & 2.46 & 2.03 & 2.90 & 3.24 & 5.09 & 5.77 & 5.77 & 1.99 & 1.78 & Pc \\
Darcy & Box s1 & 2.49 & 2.10 & 2.96 & 3.35 & 5.55 & 6.16 & 6.16 & 2.08 & 1.84 & Pc \\
Darcy & Box s2 & 2.54 & 2.07 & 3.19 & 3.56 & 5.52 & 6.01 & 6.01 & 1.88 & 1.69 & Pc \\
Darcy & Box s3 & 2.38 & 2.02 & 2.84 & 3.23 & 5.35 & 6.08 & 6.08 & 2.14 & 1.89 & Pc \\
Darcy & Box s4 & 2.42 & 2.07 & 2.88 & 3.17 & 4.85 & 5.77 & 5.77 & 2.01 & 1.82 & Pc \\
Darcy & Smooth-mask box s0 & 2.85 & 2.77 & 3.45 & 3.82 & 6.13 & 6.64 & 6.64 & 1.92 & 1.74 & Pc \\
Darcy & Smooth-mask box s1 & 2.94 & 2.81 & 3.44 & 3.87 & 6.03 & 6.97 & 6.97 & 2.03 & 1.80 & Pc \\
Darcy & Smooth-mask box s2 & 3.08 & 2.76 & 3.73 & 4.38 & 6.07 & 6.93 & 6.93 & 1.86 & 1.58 & Pc \\
Darcy & Smooth-mask box s3 & 3.18 & 2.99 & 3.78 & 4.09 & 6.14 & 7.40 & 7.40 & 1.96 & 1.81 & Pc \\
Darcy & Smooth-mask box s4 & 3.05 & 2.95 & 3.69 & 3.96 & 6.08 & 6.99 & 6.99 & 1.90 & 1.76 & Pc \\
Darcy & Box FT s0 & 2.27 & 1.93 & 2.65 & 3.03 & 4.76 & 5.17 & 5.17 & 1.95 & 1.70 & Pc \\
Darcy & Box FT s1 & 2.35 & 1.99 & 2.80 & 3.18 & 5.33 & 5.95 & 5.95 & 2.12 & 1.87 & Pc \\
Darcy & Box FT s2 & 2.34 & 1.97 & 2.90 & 3.21 & 4.75 & 5.37 & 5.37 & 1.85 & 1.67 & Pc \\
Darcy & Box FT s3 & 2.24 & 1.92 & 2.69 & 3.06 & 4.71 & 5.53 & 5.53 & 2.05 & 1.81 & Pc \\
Darcy & Box FT s4 & 2.33 & 1.97 & 2.74 & 3.06 & 4.67 & 5.82 & 5.82 & 2.12 & 1.91 & Pc \\
Darcy & Mixed-PGD FT s0 & 2.00 & 1.96 & 2.34 & 2.66 & 3.93 & 4.05 & 4.05 & 1.73 & 1.52 & Pc \\
Darcy & Mixed-PGD FT s1 & 2.00 & 2.06 & 2.58 & 2.81 & 4.07 & 4.43 & 4.43 & 1.71 & 1.58 & Pc \\
Darcy & Mixed-PGD FT s2 & 2.07 & 2.01 & 2.50 & 2.83 & 4.05 & 4.43 & 4.43 & 1.77 & 1.57 & Pc \\
Darcy & Mixed-PGD FT s3 & 2.02 & 1.97 & 2.41 & 2.76 & 4.03 & 4.55 & 4.55 & 1.88 & 1.64 & Pc \\
Darcy & Mixed-PGD FT s4 & 2.05 & 2.00 & 2.40 & 2.73 & 4.03 & 4.29 & 4.29 & 1.79 & 1.57 & Pc \\
Darcy & SAM FT s0 & 2.15 & 2.19 & 2.93 & 3.34 & 4.92 & 4.90 & 4.90 & 1.67 & 1.47 & Pc \\
Darcy & SAM FT s1 & 2.22 & 2.27 & 3.00 & 3.36 & 4.72 & 5.22 & 5.22 & 1.74 & 1.55 & Pc \\
Darcy & SAM FT s2 & 2.24 & 2.21 & 2.98 & 3.49 & 4.87 & 4.87 & 4.87 & 1.63 & 1.40 & A \\
Darcy & SAM FT s3 & 2.14 & 2.27 & 3.00 & 3.44 & 4.86 & 4.86 & 4.86 & 1.62 & 1.41 & A \\
Darcy & SAM FT s4 & 2.26 & 2.24 & 2.89 & 3.31 & 4.71 & 5.58 & 5.58 & 1.93 & 1.69 & Pc \\
\midrule
NS & Random-error-aware s0 & 12.3 & 24.3 & 35.2 & 35.4 & 45.5 & 45.5 & 45.5 & 1.29 & 1.28 & A \\
NS & Random-error-aware s1 & 12.4 & 25.8 & 38.0 & 38.9 & 51.1 & 51.1 & 51.1 & 1.35 & 1.31 & A \\
NS & Random-error-aware s2 & 8.90 & 25.2 & 37.3 & 39.8 & 52.4 & 52.4 & 52.4 & 1.41 & 1.32 & A \\
NS & Random-error-aware s3 & 8.73 & 24.2 & 35.2 & 36.8 & 47.8 & 47.8 & 47.8 & 1.36 & 1.30 & Pa \\
NS & Random-error-aware s4 & 10.8 & 26.0 & 37.7 & 40.0 & 50.7 & 50.7 & 50.7 & 1.35 & 1.27 & Pa \\
NS & Box s0 & 14.7 & 13.0 & 15.4 & 16.8 & 24.4 & 26.1 & 26.1 & 1.70 & 1.55 & Pc \\
NS & Box s1 & 14.2 & 12.8 & 15.0 & 16.2 & 22.2 & 23.8 & 23.8 & 1.59 & 1.47 & Pc \\
NS & Box s2 & 15.3 & 13.4 & 15.8 & 17.5 & 25.7 & 28.4 & 28.4 & 1.79 & 1.62 & Pc \\
NS & Box s3 & 12.9 & 11.8 & 13.8 & 15.4 & 22.2 & 25.4 & 25.4 & 1.84 & 1.65 & Pc \\
NS & Box s4 & 14.7 & 13.0 & 15.2 & 16.9 & 25.0 & 27.1 & 27.1 & 1.79 & 1.60 & Pr \\
NS & Smooth-mask box s0 & 19.2 & 18.2 & 20.5 & 22.1 & 31.0 & 35.0 & 35.0 & 1.71 & 1.59 & Pm \\
NS & Smooth-mask box s1 & 21.2 & 20.8 & 22.8 & 23.5 & 32.3 & 35.6 & 35.6 & 1.56 & 1.51 & Pd \\
NS & Smooth-mask box s2 & 19.6 & 18.9 & 21.0 & 21.9 & 30.3 & 34.1 & 34.1 & 1.62 & 1.55 & Pc \\
NS & Smooth-mask box s3 & 16.4 & 16.1 & 17.7 & 19.1 & 25.2 & 26.3 & 26.3 & 1.48 & 1.38 & Pc \\
NS & Smooth-mask box s4 & 18.6 & 17.2 & 19.3 & 22.0 & 30.8 & 33.0 & 33.0 & 1.71 & 1.50 & Pm \\
NS & Box FT s0 & 13.7 & 12.2 & 14.5 & 15.7 & 22.5 & 24.6 & 24.6 & 1.70 & 1.57 & Pc \\
NS & Box FT s1 & 13.7 & 12.5 & 14.4 & 15.6 & 21.5 & 22.7 & 22.7 & 1.57 & 1.46 & Pc \\
NS & Box FT s2 & 14.7 & 13.0 & 15.2 & 16.7 & 25.0 & 26.9 & 26.9 & 1.77 & 1.61 & Pc \\
NS & Box FT s3 & 11.2 & 9.98 & 12.5 & 14.0 & 21.2 & 25.4 & 25.4 & 2.03 & 1.81 & Pc \\
NS & Box FT s4 & 13.9 & 12.4 & 14.5 & 16.0 & 22.9 & 26.0 & 26.0 & 1.80 & 1.62 & Pc \\
NS & Mixed-PGD FT s0 & 13.5 & 12.7 & 14.4 & 15.1 & 19.9 & 21.8 & 21.8 & 1.52 & 1.45 & Pc \\
NS & Mixed-PGD FT s1 & 13.2 & 12.7 & 14.1 & 14.9 & 19.1 & 21.7 & 21.7 & 1.53 & 1.46 & Pc \\
NS & Mixed-PGD FT s2 & 14.0 & 13.2 & 14.6 & 15.5 & 21.0 & 22.8 & 22.8 & 1.56 & 1.47 & Pc \\
NS & Mixed-PGD FT s3 & 10.7 & 10.3 & 12.0 & 12.8 & 18.4 & 21.7 & 21.7 & 1.80 & 1.70 & Pc \\
NS & Mixed-PGD FT s4 & 13.6 & 12.8 & 14.3 & 14.9 & 19.6 & 21.5 & 21.5 & 1.51 & 1.44 & Pc \\
NS & SAM FT s0 & 14.3 & 12.9 & 15.0 & 16.3 & 22.5 & 24.7 & 24.7 & 1.65 & 1.52 & Pc \\
NS & SAM FT s1 & 14.0 & 12.8 & 14.6 & 15.8 & 21.7 & 22.9 & 22.9 & 1.57 & 1.45 & Pc \\
NS & SAM FT s2 & 14.9 & 13.2 & 15.3 & 17.0 & 24.8 & 27.3 & 27.3 & 1.78 & 1.61 & Pc \\
NS & SAM FT s3 & 11.9 & 10.9 & 13.0 & 14.5 & 22.9 & 25.1 & 25.1 & 1.94 & 1.73 & Pc \\
NS & SAM FT s4 & 14.6 & 12.9 & 15.0 & 16.6 & 23.4 & 25.4 & 25.4 & 1.69 & 1.52 & Pc \\

\bottomrule
\end{tabular}

%% file: tables/tabS_models_b.tex
\begin{tabular}{llrrrrrrrrrl}
\toprule
Task & Arm, seed & nominal & MC mean & max$_{200}$ & max$_{1000}$ & unseen & DE/PGD & searched & $R_{200}$ & $R_{1000}$ & method \\
\midrule
LDC & Random-error-aware s0 & 1.08 & 13.6 & 26.3 & 26.3 & 35.8 & 35.8 & 35.8 & 1.36 & 1.36 & A \\
LDC & Random-error-aware s1 & 0.702 & 10.4 & 20.1 & 20.7 & 27.6 & 28.1 & 28.1 & 1.40 & 1.35 & Pa \\
LDC & Random-error-aware s2 & 1.02 & 9.07 & 17.8 & 18.9 & 24.2 & 24.6 & 24.6 & 1.38 & 1.30 & Pa \\
LDC & Random-error-aware s3 & 1.64 & 9.72 & 18.8 & 21.1 & 28.7 & 28.7 & 28.7 & 1.52 & 1.36 & A \\
LDC & Random-error-aware s4 & 0.948 & 11.9 & 21.8 & 23.5 & 33.5 & 33.5 & 33.5 & 1.53 & 1.43 & A \\
LDC & Box s0 & 0.820 & 0.860 & 1.17 & 1.36 & 2.55 & 2.60 & 2.60 & 2.22 & 1.91 & Pd \\
LDC & Box s1 & 0.673 & 0.677 & 0.807 & 0.862 & 1.43 & 1.51 & 1.51 & 1.87 & 1.75 & Pd \\
LDC & Box s2 & 0.793 & 0.778 & 1.28 & 1.60 & 3.09 & 3.78 & 3.78 & 2.95 & 2.37 & Pr \\
LDC & Box s3 & 1.71 & 1.63 & 2.58 & 2.62 & 3.77 & 4.10 & 4.10 & 1.59 & 1.57 & Pr \\
LDC & Box s4 & 0.626 & 0.702 & 1.05 & 1.19 & 2.44 & 3.23 & 3.23 & 3.07 & 2.71 & Pc \\
LDC & Smooth-mask box s0 & 0.724 & 0.722 & 1.28 & 1.44 & 2.90 & 3.04 & 3.04 & 2.36 & 2.11 & Pa \\
LDC & Smooth-mask box s1 & 1.33 & 1.36 & 2.10 & 2.42 & 3.80 & 3.97 & 3.97 & 1.89 & 1.64 & Pa \\
LDC & Smooth-mask box s2 & 0.688 & 0.781 & 1.42 & 1.42 & 2.34 & 2.51 & 2.51 & 1.77 & 1.76 & Pa \\
LDC & Smooth-mask box s3 & 1.94 & 1.97 & 2.90 & 3.20 & 5.17 & 5.39 & 5.39 & 1.86 & 1.69 & Pa \\
LDC & Smooth-mask box s4 & 0.700 & 0.699 & 1.00 & 1.41 & 2.90 & 2.90 & 2.90 & 2.89 & 2.05 & Pa \\
LDC & Box FT s0 & 0.750 & 0.729 & 0.957 & 1.00 & 1.85 & 1.99 & 1.99 & 2.08 & 1.99 & Pd \\
LDC & Box FT s1 & 0.534 & 0.528 & 0.595 & 0.623 & 0.887 & 0.908 & 0.908 & 1.53 & 1.46 & Pa \\
LDC & Box FT s2 & 0.657 & 0.641 & 1.01 & 1.12 & 2.22 & 2.80 & 2.80 & 2.77 & 2.49 & Pd \\
LDC & Box FT s3 & 1.53 & 1.39 & 2.25 & 2.25 & 3.35 & 3.52 & 3.52 & 1.57 & 1.57 & Pa \\
LDC & Box FT s4 & 0.578 & 0.624 & 0.850 & 0.995 & 2.36 & 2.64 & 2.64 & 3.10 & 2.65 & Pa \\
LDC & Mixed-PGD FT s0 & 0.725 & 0.724 & 0.842 & 0.844 & 1.18 & 1.22 & 1.22 & 1.45 & 1.44 & Pd \\
LDC & Mixed-PGD FT s1 & 0.484 & 0.483 & 0.500 & 0.512 & 0.531 & 0.541 & 0.541 & 1.08 & 1.06 & Pd \\
LDC & Mixed-PGD FT s2 & 0.659 & 0.634 & 0.837 & 0.837 & 1.26 & 1.53 & 1.53 & 1.83 & 1.83 & Pd \\
LDC & Mixed-PGD FT s3 & 1.40 & 1.32 & 1.73 & 1.73 & 2.25 & 2.41 & 2.41 & 1.39 & 1.39 & Pm \\
LDC & Mixed-PGD FT s4 & 0.553 & 0.570 & 0.687 & 0.687 & 0.872 & 0.990 & 0.990 & 1.44 & 1.44 & Pr \\
LDC & SAM FT s0 & 1.37 & 1.60 & 2.37 & 2.52 & 3.86 & 4.11 & 4.11 & 1.73 & 1.63 & Pc \\
LDC & SAM FT s1 & 1.15 & 1.14 & 1.29 & 1.48 & 2.16 & 2.05 & 2.05 & 1.59 & 1.39 & Pc \\
LDC & SAM FT s2 & 1.22 & 1.45 & 2.11 & 2.30 & 3.86 & 3.97 & 3.97 & 1.88 & 1.73 & Pc \\
LDC & SAM FT s3 & 2.13 & 2.50 & 3.50 & 3.81 & 5.41 & 5.44 & 5.44 & 1.55 & 1.43 & Pd \\
LDC & SAM FT s4 & 1.71 & 1.82 & 2.63 & 2.63 & 4.24 & 4.24 & 4.24 & 1.61 & 1.61 & A \\
\midrule
HX & Random-error-aware s0 & 0.240 & 2.30 & 3.62 & 3.93 & 4.85 & 4.85 & 4.85 & 1.34 & 1.23 & A \\
HX & Random-error-aware s1 & 0.239 & 2.22 & 3.46 & 3.84 & 4.64 & 4.64 & 4.64 & 1.34 & 1.21 & A \\
HX & Random-error-aware s2 & 0.237 & 2.30 & 3.55 & 4.05 & 4.94 & 4.94 & 4.94 & 1.39 & 1.22 & A \\
HX & Random-error-aware s3 & 0.258 & 2.32 & 3.57 & 3.95 & 4.81 & 4.81 & 4.81 & 1.35 & 1.22 & A \\
HX & Random-error-aware s4 & 0.240 & 2.27 & 3.58 & 3.94 & 4.84 & 4.84 & 4.84 & 1.35 & 1.23 & A \\
HX & Box s0 & 0.353 & 0.316 & 0.432 & 0.505 & 0.969 & 1.21 & 1.21 & 2.80 & 2.40 & Pm \\
HX & Box s1 & 0.350 & 0.318 & 0.445 & 0.508 & 0.949 & 1.20 & 1.20 & 2.70 & 2.36 & Pc \\
HX & Box s2 & 0.335 & 0.298 & 0.420 & 0.463 & 0.900 & 1.14 & 1.14 & 2.72 & 2.47 & Pc \\
HX & Box s3 & 0.356 & 0.327 & 0.464 & 0.533 & 1.02 & 1.23 & 1.23 & 2.64 & 2.30 & Pc \\
HX & Box s4 & 0.360 & 0.319 & 0.427 & 0.498 & 0.818 & 1.10 & 1.10 & 2.57 & 2.21 & Pm \\
HX & Smooth-mask box s0 & 0.482 & 0.434 & 0.617 & 0.629 & 0.948 & 1.09 & 1.09 & 1.77 & 1.74 & Pd \\
HX & Smooth-mask box s1 & 0.534 & 0.467 & 0.645 & 0.662 & 1.06 & 1.20 & 1.20 & 1.85 & 1.81 & Pc \\
HX & Smooth-mask box s2 & 0.517 & 0.476 & 0.645 & 0.692 & 1.14 & 1.22 & 1.22 & 1.90 & 1.77 & Pr \\
HX & Smooth-mask box s3 & 0.549 & 0.498 & 0.644 & 0.712 & 1.25 & 1.28 & 1.28 & 1.99 & 1.80 & Pc \\
HX & Smooth-mask box s4 & 0.531 & 0.496 & 0.624 & 0.717 & 1.31 & 1.31 & 1.31 & 2.10 & 1.83 & A \\
HX & Box FT s0 & 0.329 & 0.292 & 0.399 & 0.476 & 0.975 & 1.18 & 1.18 & 2.95 & 2.47 & Pm \\
HX & Box FT s1 & 0.317 & 0.299 & 0.420 & 0.470 & 0.851 & 1.10 & 1.10 & 2.62 & 2.34 & Pc \\
HX & Box FT s2 & 0.317 & 0.283 & 0.398 & 0.445 & 0.729 & 1.12 & 1.12 & 2.81 & 2.51 & Pc \\
HX & Box FT s3 & 0.340 & 0.307 & 0.443 & 0.513 & 0.934 & 1.21 & 1.21 & 2.73 & 2.36 & Pc \\
HX & Box FT s4 & 0.342 & 0.296 & 0.410 & 0.476 & 0.833 & 1.06 & 1.06 & 2.59 & 2.23 & Pm \\
HX & Mixed-PGD FT s0 & 0.319 & 0.297 & 0.402 & 0.469 & 0.783 & 0.979 & 0.979 & 2.43 & 2.09 & Pm \\
HX & Mixed-PGD FT s1 & 0.321 & 0.295 & 0.414 & 0.458 & 0.727 & 0.893 & 0.893 & 2.16 & 1.95 & Pd \\
HX & Mixed-PGD FT s2 & 0.321 & 0.291 & 0.390 & 0.430 & 0.694 & 0.942 & 0.942 & 2.41 & 2.19 & Pc \\
HX & Mixed-PGD FT s3 & 0.338 & 0.310 & 0.430 & 0.499 & 0.913 & 1.05 & 1.05 & 2.44 & 2.10 & Pd \\
HX & Mixed-PGD FT s4 & 0.335 & 0.311 & 0.396 & 0.456 & 0.664 & 0.895 & 0.895 & 2.26 & 1.96 & Pm \\
HX & SAM FT s0 & 0.449 & 0.397 & 0.523 & 0.613 & 1.19 & 1.36 & 1.36 & 2.59 & 2.21 & Pm \\
HX & SAM FT s1 & 0.421 & 0.396 & 0.519 & 0.578 & 0.993 & 1.33 & 1.33 & 2.55 & 2.29 & Pc \\
HX & SAM FT s2 & 0.439 & 0.397 & 0.528 & 0.599 & 1.14 & 1.37 & 1.37 & 2.59 & 2.29 & Pc \\
HX & SAM FT s3 & 0.442 & 0.404 & 0.551 & 0.636 & 1.22 & 1.44 & 1.44 & 2.61 & 2.26 & Pc \\
HX & SAM FT s4 & 0.435 & 0.397 & 0.531 & 0.607 & 1.15 & 1.39 & 1.39 & 2.61 & 2.29 & Pr \\

\bottomrule
\end{tabular}

%% file: tables/tabS_real_a.tex
\begin{tabular}{llrrrrrrrrrr}
\toprule
 & & \multicolumn{5}{c}{\textsc{real} device (\%)} & \multicolumn{5}{c}{\textsc{det}: devices of 1000 exceeding $c\times$ nominal} \\
\cmidrule(lr){3-7}\cmidrule(lr){8-12}
Task & Arm, seed & nominal & max$_{200}$ & $z_{\rm search}$ mean & $z_{\rm search}$ max & nested & $c{=}1.1$ & 1.25 & 1.5 & 2 & UCB $q_{0.99}$ \\
\midrule
Darcy & Random-error-aware s0 & 1.62 & 7.12 & 10.2 & 10.3 & 10.2 & 1000 & 1000 & 998 & 964 & 7.48 \\
Darcy & Random-error-aware s1 & 1.66 & 6.36 & 8.97 & 9.06 & 8.98 & 1000 & 1000 & 997 & 899 & 6.26 \\
Darcy & Random-error-aware s2 & 1.65 & 6.91 & 9.69 & 9.76 & 9.68 & 1000 & 1000 & 997 & 916 & 6.79 \\
Darcy & Random-error-aware s3 & 1.68 & 7.77 & 10.9 & 11.2 & 10.9 & 1000 & 1000 & 999 & 948 & 7.95 \\
Darcy & Random-error-aware s4 & 1.58 & 11.3 & 14.4 & 15.1 & 14.7 & 1000 & 1000 & 1000 & 974 & 10.7 \\
Darcy & Box s0 & 2.51 & 3.17 & 5.90 & 6.16 & 5.87 & 10 & 3 & 0 & 0 & 2.87 \\
Darcy & Box s1 & 2.56 & 3.38 & 6.20 & 6.59 & 6.13 & 16 & 3 & 0 & 0 & 2.96 \\
Darcy & Box s2 & 2.54 & 3.26 & 6.09 & 6.52 & 6.07 & 12 & 4 & 0 & 0 & 3.03 \\
Darcy & Box s3 & 2.43 & 3.16 & 6.20 & 6.83 & 6.31 & 17 & 3 & 0 & 0 & 2.84 \\
Darcy & Box s4 & 2.45 & 3.17 & 6.00 & 6.36 & 6.02 & 15 & 3 & 0 & 0 & 2.88 \\
Darcy & Smooth-mask box s0 & 2.97 & 3.50 & 7.07 & 7.80 & 6.93 & 73 & 3 & 0 & 0 & 3.51 \\
Darcy & Smooth-mask box s1 & 2.98 & 3.61 & 6.90 & 7.48 & 6.94 & 45 & 3 & 0 & 0 & 3.56 \\
Darcy & Smooth-mask box s2 & 3.12 & 3.83 & 7.21 & 7.94 & 7.29 & 13 & 3 & 0 & 0 & 3.69 \\
Darcy & Smooth-mask box s3 & 2.98 & 3.64 & 6.64 & 7.62 & 6.79 & 23 & 2 & 0 & 0 & 3.78 \\
Darcy & Smooth-mask box s4 & 3.09 & 3.89 & 6.69 & 7.17 & 6.63 & 76 & 4 & 0 & 0 & 3.80 \\
Darcy & Box FT s0 & 2.32 & 2.93 & 5.28 & 5.44 & 5.26 & 14 & 2 & 0 & 0 & 2.66 \\
Darcy & Box FT s1 & 2.41 & 3.14 & 5.98 & 6.42 & 5.90 & 18 & 3 & 0 & 0 & 2.79 \\
Darcy & Box FT s2 & 2.35 & 2.99 & 5.47 & 5.85 & 5.44 & 20 & 3 & 0 & 0 & 2.90 \\
Darcy & Box FT s3 & 2.29 & 2.95 & 5.63 & 6.01 & 5.67 & 20 & 3 & 0 & 0 & 2.70 \\
Darcy & Box FT s4 & 2.41 & 3.20 & 6.10 & 6.50 & 6.08 & 12 & 2 & 0 & 0 & 2.74 \\
Darcy & Mixed-PGD FT s0 & 2.06 & 2.51 & 4.04 & 4.19 & 4.06 & 45 & 2 & 0 & 0 & 2.45 \\
Darcy & Mixed-PGD FT s1 & 2.05 & 2.67 & 4.48 & 4.80 & 4.41 & 151 & 15 & 0 & 0 & 2.58 \\
Darcy & Mixed-PGD FT s2 & 2.09 & 2.70 & 4.54 & 5.02 & 4.48 & 64 & 7 & 0 & 0 & 2.60 \\
Darcy & Mixed-PGD FT s3 & 2.08 & 2.59 & 4.63 & 4.89 & 4.64 & 67 & 5 & 0 & 0 & 2.53 \\
Darcy & Mixed-PGD FT s4 & 2.11 & 2.66 & 4.51 & 4.75 & 4.50 & 62 & 3 & 0 & 0 & 2.48 \\
Darcy & SAM FT s0 & 2.19 & 3.05 & 4.96 & 5.12 & 4.97 & 179 & 33 & 1 & 0 & 2.93 \\
Darcy & SAM FT s1 & 2.28 & 3.02 & 5.29 & 5.60 & 5.22 & 185 & 34 & 1 & 0 & 3.01 \\
Darcy & SAM FT s2 & 2.24 & 3.26 & 4.90 & 4.94 & 4.90 & 157 & 30 & 1 & 0 & 3.09 \\
Darcy & SAM FT s3 & 2.20 & 3.07 & 4.89 & 4.98 & 5.18 & 298 & 67 & 1 & 0 & 3.06 \\
Darcy & SAM FT s4 & 2.34 & 3.15 & 5.81 & 6.10 & 5.82 & 140 & 18 & 0 & 0 & 2.99 \\
\midrule
NS & Random-error-aware s0 & 12.3 & 33.4 & 45.5 & 46.1 & 45.5 & 1000 & 990 & 904 & 471 & 34.9 \\
NS & Random-error-aware s1 & 12.5 & 36.1 & 50.9 & 51.1 & 50.9 & 1000 & 981 & 924 & 586 & 38.1 \\
NS & Random-error-aware s2 & 9.03 & 36.9 & 52.3 & 52.9 & 52.4 & 1000 & 1000 & 981 & 902 & 38.2 \\
NS & Random-error-aware s3 & 8.79 & 33.7 & 47.4 & 48.0 & 47.6 & 1000 & 1000 & 992 & 901 & 35.8 \\
NS & Random-error-aware s4 & 10.8 & 36.8 & 50.5 & 51.2 & 50.4 & 1000 & 998 & 968 & 784 & 38.3 \\
NS & Box s0 & 14.7 & 16.4 & 25.9 & 26.6 & 25.9 & 1 & 0 & 0 & 0 & 15.4 \\
NS & Box s1 & 14.3 & 16.1 & 23.9 & 24.4 & 23.8 & 1 & 0 & 0 & 0 & 15.0 \\
NS & Box s2 & 15.5 & 17.3 & 28.2 & 28.6 & 28.0 & 1 & 0 & 0 & 0 & 16.2 \\
NS & Box s3 & 13.0 & 14.8 & 24.9 & 25.7 & 25.1 & 6 & 0 & 0 & 0 & 14.4 \\
NS & Box s4 & 14.9 & 16.4 & 27.5 & 28.7 & 27.5 & 1 & 0 & 0 & 0 & 15.3 \\
NS & Smooth-mask box s0 & 19.3 & 20.7 & 35.1 & 37.0 & 35.6 & 2 & 0 & 0 & 0 & 20.9 \\
NS & Smooth-mask box s1 & 21.1 & 22.7 & 33.2 & 35.7 & 33.0 & 4 & 0 & 0 & 0 & 23.3 \\
NS & Smooth-mask box s2 & 19.7 & 21.2 & 33.4 & 34.4 & 33.2 & 2 & 0 & 0 & 0 & 21.0 \\
NS & Smooth-mask box s3 & 16.5 & 17.5 & 25.7 & 26.6 & 26.0 & 8 & 0 & 0 & 0 & 18.2 \\
NS & Smooth-mask box s4 & 18.3 & 20.8 & 31.8 & 33.6 & 32.0 & 1 & 0 & 0 & 0 & 19.7 \\
NS & Box FT s0 & 13.7 & 15.4 & 24.3 & 24.9 & 24.4 & 3 & 0 & 0 & 0 & 14.6 \\
NS & Box FT s1 & 13.9 & 15.4 & 22.8 & 23.1 & 22.5 & 1 & 0 & 0 & 0 & 14.5 \\
NS & Box FT s2 & 14.9 & 16.5 & 26.7 & 27.1 & 26.6 & 1 & 0 & 0 & 0 & 15.5 \\
NS & Box FT s3 & 11.3 & 13.6 & 24.9 & 25.8 & 25.2 & 13 & 1 & 0 & 0 & 12.9 \\
NS & Box FT s4 & 14.1 & 15.8 & 26.5 & 28.1 & 26.6 & 1 & 0 & 0 & 0 & 14.7 \\
NS & Mixed-PGD FT s0 & 13.4 & 14.8 & 21.4 & 22.0 & 21.4 & 2 & 0 & 0 & 0 & 14.4 \\
NS & Mixed-PGD FT s1 & 13.3 & 14.5 & 21.4 & 21.9 & 21.3 & 1 & 0 & 0 & 0 & 14.0 \\
NS & Mixed-PGD FT s2 & 14.2 & 15.5 & 22.7 & 23.2 & 22.6 & 1 & 0 & 0 & 0 & 14.8 \\
NS & Mixed-PGD FT s3 & 10.9 & 12.6 & 21.1 & 21.7 & 21.3 & 13 & 0 & 0 & 0 & 12.1 \\
NS & Mixed-PGD FT s4 & 13.7 & 14.8 & 22.0 & 22.7 & 21.9 & 0 & 0 & 0 & 0 & 14.4 \\
NS & SAM FT s0 & 14.4 & 16.2 & 24.6 & 25.1 & 24.8 & 1 & 0 & 0 & 0 & 15.1 \\
NS & SAM FT s1 & 14.1 & 15.8 & 23.0 & 23.4 & 23.1 & 1 & 0 & 0 & 0 & 14.7 \\
NS & SAM FT s2 & 15.1 & 16.8 & 27.1 & 27.4 & 27.1 & 2 & 0 & 0 & 0 & 15.8 \\
NS & SAM FT s3 & 12.0 & 14.1 & 24.8 & 25.5 & 25.1 & 11 & 0 & 0 & 0 & 13.7 \\
NS & SAM FT s4 & 14.7 & 16.2 & 25.8 & 26.8 & 25.8 & 1 & 0 & 0 & 0 & 15.1 \\

\bottomrule
\end{tabular}

%% file: tables/tabS_real_b.tex
\begin{tabular}{llrrrrrrrrrr}
\toprule
 & & \multicolumn{5}{c}{\textsc{real} device (\%)} & \multicolumn{5}{c}{\textsc{det}: devices of 1000 exceeding $c\times$ nominal} \\
\cmidrule(lr){3-7}\cmidrule(lr){8-12}
Task & Arm, seed & nominal & max$_{200}$ & $z_{\rm search}$ mean & $z_{\rm search}$ max & nested & $c{=}1.1$ & 1.25 & 1.5 & 2 & UCB $q_{0.99}$ \\
\midrule
LDC & Random-error-aware s0 & 1.20 & 25.4 & 35.3 & 35.9 & 35.4 & 995 & 989 & 987 & 975 & 25.6 \\
LDC & Random-error-aware s1 & 1.26 & 19.8 & 28.2 & 29.3 & 28.0 & 1000 & 1000 & 991 & 981 & 19.9 \\
LDC & Random-error-aware s2 & 1.29 & 17.2 & 25.2 & 26.0 & 25.2 & 991 & 987 & 971 & 953 & 17.8 \\
LDC & Random-error-aware s3 & 1.25 & 20.0 & 28.5 & 29.2 & 28.5 & 972 & 966 & 958 & 925 & 19.4 \\
LDC & Random-error-aware s4 & 1.25 & 21.9 & 33.7 & 34.3 & 33.8 & 994 & 989 & 983 & 969 & 22.2 \\
LDC & Box s0 & 1.20 & 1.86 & 2.11 & 3.60 & 2.31 & 231 & 49 & 7 & 0 & 1.25 \\
LDC & Box s1 & 0.897 & 1.42 & 1.49 & 1.82 & 1.63 & 46 & 2 & 0 & 0 & 0.807 \\
LDC & Box s2 & 1.03 & 2.22 & 2.57 & 3.91 & 2.64 & 190 & 73 & 20 & 1 & 1.36 \\
LDC & Box s3 & 1.15 & 2.15 & 3.08 & 4.08 & 3.17 & 219 & 71 & 2 & 0 & 2.42 \\
LDC & Box s4 & 0.962 & 1.71 & 2.44 & 3.65 & 2.32 & 470 & 141 & 24 & 0 & 1.05 \\
LDC & Smooth-mask box s0 & 1.05 & 1.95 & 3.01 & 3.88 & 2.91 & 275 & 146 & 27 & 0 & 1.24 \\
LDC & Smooth-mask box s1 & 0.983 & 2.15 & 3.60 & 4.63 & 3.53 & 383 & 201 & 33 & 0 & 2.35 \\
LDC & Smooth-mask box s2 & 0.917 & 2.01 & 2.31 & 3.56 & 2.13 & 444 & 277 & 96 & 3 & 1.32 \\
LDC & Smooth-mask box s3 & 0.997 & 2.29 & 4.07 & 5.43 & 4.18 & 334 & 105 & 9 & 0 & 2.95 \\
LDC & Smooth-mask box s4 & 0.938 & 1.94 & 3.25 & 4.02 & 3.15 & 166 & 55 & 3 & 1 & 1.04 \\
LDC & Box FT s0 & 1.05 & 1.57 & 1.86 & 2.77 & 1.87 & 68 & 4 & 0 & 0 & 0.931 \\
LDC & Box FT s1 & 0.621 & 0.880 & 0.923 & 1.07 & 0.987 & 8 & 0 & 0 & 0 & 0.595 \\
LDC & Box FT s2 & 0.908 & 1.84 & 1.94 & 2.94 & 2.16 & 148 & 48 & 8 & 0 & 1.01 \\
LDC & Box FT s3 & 1.05 & 1.78 & 2.54 & 3.51 & 2.54 & 132 & 26 & 0 & 0 & 2.07 \\
LDC & Box FT s4 & 0.820 & 1.37 & 2.09 & 3.15 & 2.03 & 334 & 70 & 7 & 0 & 0.938 \\
LDC & Mixed-PGD FT s0 & 0.918 & 1.31 & 1.04 & 1.66 & 1.16 & 31 & 0 & 0 & 0 & 0.836 \\
LDC & Mixed-PGD FT s1 & 0.531 & 0.602 & 0.611 & 0.746 & 0.610 & 0 & 0 & 0 & 0 & 0.500 \\
LDC & Mixed-PGD FT s2 & 0.794 & 1.49 & 1.00 & 1.65 & 1.32 & 31 & 1 & 0 & 0 & 0.792 \\
LDC & Mixed-PGD FT s3 & 0.973 & 1.58 & 1.57 & 2.40 & 1.57 & 52 & 0 & 0 & 0 & 1.65 \\
LDC & Mixed-PGD FT s4 & 0.748 & 1.06 & 0.829 & 1.31 & 0.820 & 61 & 0 & 0 & 0 & 0.642 \\
LDC & SAM FT s0 & 1.62 & 3.17 & 4.28 & 4.98 & 4.23 & 615 & 252 & 34 & 0 & 2.38 \\
LDC & SAM FT s1 & 1.21 & 1.44 & 1.95 & 2.17 & 1.94 & 24 & 1 & 0 & 0 & 1.37 \\
LDC & SAM FT s2 & 1.61 & 3.11 & 3.25 & 4.19 & 3.71 & 678 & 322 & 49 & 0 & 2.09 \\
LDC & SAM FT s3 & 1.77 & 3.34 & 4.76 & 5.45 & 4.78 & 635 & 335 & 39 & 0 & 3.50 \\
LDC & SAM FT s4 & 1.88 & 3.00 & 4.50 & 4.89 & 4.50 & 364 & 82 & 1 & 0 & 2.45 \\
\midrule
HX & Random-error-aware s0 & 0.259 & 3.76 & 4.85 & 4.88 & 4.84 & 1000 & 1000 & 1000 & 1000 & 3.69 \\
HX & Random-error-aware s1 & 0.261 & 3.63 & 4.66 & 4.68 & 4.66 & 1000 & 1000 & 1000 & 1000 & 3.50 \\
HX & Random-error-aware s2 & 0.239 & 3.64 & 4.95 & 4.96 & 4.95 & 1000 & 1000 & 1000 & 1000 & 3.66 \\
HX & Random-error-aware s3 & 0.275 & 3.67 & 4.82 & 4.84 & 4.82 & 1000 & 1000 & 1000 & 1000 & 3.65 \\
HX & Random-error-aware s4 & 0.253 & 3.68 & 4.87 & 4.89 & 4.87 & 1000 & 1000 & 1000 & 1000 & 3.64 \\
HX & Box s0 & 0.357 & 0.483 & 1.23 & 1.27 & 1.23 & 24 & 4 & 0 & 0 & 0.432 \\
HX & Box s1 & 0.353 & 0.459 & 1.15 & 1.22 & 1.15 & 31 & 4 & 0 & 0 & 0.433 \\
HX & Box s2 & 0.330 & 0.420 & 1.11 & 1.18 & 1.11 & 28 & 5 & 0 & 0 & 0.420 \\
HX & Box s3 & 0.361 & 0.499 & 1.19 & 1.32 & 1.19 & 35 & 7 & 0 & 0 & 0.457 \\
HX & Box s4 & 0.349 & 0.457 & 1.11 & 1.15 & 1.12 & 33 & 2 & 0 & 0 & 0.438 \\
HX & Smooth-mask box s0 & 0.465 & 0.590 & 1.07 & 1.11 & 1.09 & 33 & 5 & 0 & 0 & 0.610 \\
HX & Smooth-mask box s1 & 0.515 & 0.618 & 1.18 & 1.27 & 1.18 & 19 & 0 & 0 & 0 & 0.630 \\
HX & Smooth-mask box s2 & 0.513 & 0.665 & 1.19 & 1.24 & 1.20 & 34 & 2 & 0 & 0 & 0.645 \\
HX & Smooth-mask box s3 & 0.561 & 0.703 & 1.29 & 1.38 & 1.28 & 23 & 5 & 0 & 0 & 0.689 \\
HX & Smooth-mask box s4 & 0.568 & 0.667 & 1.28 & 1.33 & 1.28 & 45 & 5 & 0 & 0 & 0.670 \\
HX & Box FT s0 & 0.335 & 0.443 & 1.19 & 1.24 & 1.19 & 26 & 5 & 0 & 0 & 0.412 \\
HX & Box FT s1 & 0.320 & 0.417 & 1.05 & 1.11 & 1.06 & 85 & 10 & 0 & 0 & 0.415 \\
HX & Box FT s2 & 0.310 & 0.403 & 1.09 & 1.14 & 1.09 & 29 & 7 & 0 & 0 & 0.408 \\
HX & Box FT s3 & 0.342 & 0.478 & 1.17 & 1.30 & 1.18 & 35 & 6 & 1 & 0 & 0.443 \\
HX & Box FT s4 & 0.329 & 0.432 & 1.06 & 1.10 & 1.07 & 19 & 3 & 0 & 0 & 0.416 \\
HX & Mixed-PGD FT s0 & 0.324 & 0.441 & 0.997 & 1.04 & 0.989 & 37 & 5 & 0 & 0 & 0.402 \\
HX & Mixed-PGD FT s1 & 0.317 & 0.399 & 0.847 & 0.884 & 0.903 & 31 & 4 & 0 & 0 & 0.401 \\
HX & Mixed-PGD FT s2 & 0.312 & 0.385 & 0.907 & 0.959 & 0.909 & 23 & 3 & 0 & 0 & 0.386 \\
HX & Mixed-PGD FT s3 & 0.339 & 0.461 & 1.01 & 1.14 & 1.02 & 26 & 5 & 0 & 0 & 0.423 \\
HX & Mixed-PGD FT s4 & 0.327 & 0.415 & 0.904 & 0.950 & 0.934 & 44 & 2 & 0 & 0 & 0.403 \\
HX & SAM FT s0 & 0.451 & 0.609 & 1.39 & 1.43 & 1.39 & 10 & 2 & 0 & 0 & 0.524 \\
HX & SAM FT s1 & 0.422 & 0.533 & 1.30 & 1.34 & 1.31 & 18 & 2 & 0 & 0 & 0.516 \\
HX & SAM FT s2 & 0.435 & 0.556 & 1.35 & 1.39 & 1.35 & 13 & 2 & 0 & 0 & 0.528 \\
HX & SAM FT s3 & 0.441 & 0.624 & 1.41 & 1.52 & 1.42 & 17 & 3 & 0 & 0 & 0.532 \\
HX & SAM FT s4 & 0.445 & 0.602 & 1.43 & 1.48 & 1.44 & 17 & 2 & 0 & 0 & 0.527 \\

\bottomrule
\end{tabular}

%% file: tables/tabS_v3_e1.tex
\begin{tabular}{llrrrrrrrrr}
\toprule
 & & & \multicolumn{2}{c}{uniform} & \multicolumn{2}{c}{Sobol'} & \multicolumn{2}{c}{corners} & \multicolumn{2}{c}{corners + uniform} \\
\cmidrule(lr){4-5}\cmidrule(lr){6-7}\cmidrule(lr){8-9}\cmidrule(lr){10-11}
Task & Arm (s0) & searched & error & $R_{\rm match}$ & error & $R_{\rm match}$ & error & $R_{\rm match}$ & error & $R_{\rm match}$ \\
\midrule
Darcy & Random-error-aware & 10.0 & 8.30 & 1.21 & 8.24 & 1.22 & 10.0 & 1.00 & 9.83 & 1.02 \\
 & Box & 5.77 & 3.52 & 1.64 & 3.73 & 1.55 & 4.75 & 1.21 & 4.75 & 1.21 \\
 & Smooth-mask box & 6.64 & 4.11 & 1.62 & 3.96 & 1.68 & 5.13 & 1.30 & 5.11 & 1.30 \\
 & Box FT & 5.17 & 2.96 & 1.74 & 3.35 & 1.54 & 4.62 & 1.12 & 4.62 & 1.12 \\
 & Mixed-PGD FT & 4.05 & 2.63 & 1.54 & 2.69 & 1.50 & 3.90 & 1.04 & 3.90 & 1.04 \\
 & SAM FT & 4.90 & 3.24 & 1.51 & 3.40 & 1.44 & 4.78 & 1.03 & 4.78 & 1.03 \\
\addlinespace
NS & Random-error-aware & 45.5 & 37.0 & 1.23 & 37.9 & 1.20 & 44.9 & 1.01 & 45.1 & 1.01 \\
 & Box & 26.1 & 17.5 & 1.49 & 17.8 & 1.47 & 21.8 & 1.20 & 22.0 & 1.19 \\
 & Smooth-mask box & 35.0 & 23.3 & 1.50 & 23.8 & 1.47 & 28.0 & 1.25 & 28.0 & 1.25 \\
 & Box FT & 24.6 & 16.2 & 1.52 & 16.7 & 1.48 & 21.5 & 1.14 & 21.5 & 1.14 \\
 & Mixed-PGD FT & 21.8 & 15.4 & 1.42 & 15.9 & 1.37 & 18.6 & 1.17 & 18.6 & 1.17 \\
 & SAM FT & 24.7 & 16.9 & 1.46 & 17.1 & 1.44 & 21.6 & 1.14 & 21.8 & 1.13 \\
\addlinespace
LDC & Random-error-aware & 35.8 & 29.1 & 1.23 & 28.0 & 1.28 & 34.7 & 1.03 & 34.6 & 1.03 \\
 & Box & 2.60 & 1.52 & 1.71 & 1.68 & 1.55 & 2.40 & 1.08 & 2.40 & 1.08 \\
 & Smooth-mask box & 3.04 & 1.54 & 1.97 & 1.64 & 1.85 & 2.58 & 1.18 & 2.48 & 1.22 \\
 & Box FT & 1.99 & 1.14 & 1.74 & 1.30 & 1.53 & 1.73 & 1.15 & 1.73 & 1.15 \\
 & Mixed-PGD FT & 1.22 & 0.936 & 1.30 & 0.916 & 1.33 & 1.12 & 1.09 & 1.11 & 1.10 \\
 & SAM FT & 4.11 & 2.77 & 1.48 & 2.76 & 1.49 & 3.30 & 1.24 & 3.17 & 1.30 \\
\addlinespace
HX & Random-error-aware & 4.85 & 3.93 & 1.23 & 4.07 & 1.19 & 4.80 & 1.01 & 4.78 & 1.02 \\
 & Box & 1.21 & 0.544 & 2.23 & 0.562 & 2.16 & 0.764 & 1.59 & 0.764 & 1.59 \\
 & Smooth-mask box & 1.09 & 0.664 & 1.64 & 0.653 & 1.67 & 0.892 & 1.22 & 0.892 & 1.22 \\
 & Box FT & 1.18 & 0.521 & 2.26 & 0.550 & 2.14 & 0.771 & 1.52 & 0.771 & 1.52 \\
 & Mixed-PGD FT & 0.979 & 0.465 & 2.11 & 0.492 & 1.99 & 0.520 & 1.88 & 0.520 & 1.88 \\
 & SAM FT & 1.36 & 0.636 & 2.13 & 0.715 & 1.90 & 0.712 & 1.90 & 0.712 & 1.90 \\
\bottomrule
\end{tabular}

%% file: tables/tabS_v3_mc.tex
\begin{tabular}{lrrrrrrrrrrr}
\toprule
 & \multicolumn{3}{c}{best development error (\%)} & & & & \multicolumn{5}{c}{generalised Pareto tail} \\
\cmidrule(lr){2-4}\cmidrule(lr){8-12}
Task (Box s0) & $10^3$ & $10^4$ & $10^5$ & searched & $\ge$ searched & $R$ at $10^5$ & $\xi$ & $N^*$ ($q{=}0.98$) & $N^*$ (0.99) & $N^*$ (0.995) & bootstrap 5\% \\
\midrule
Darcy & 3.03 & 3.27 & 3.41 & 5.41 & 0 & 1.59 & $-0.030$ & $3.8\times10^{16}$ & $3.6\times10^{17}$ & $5.4\times10^{11}$ & $9.5\times10^{9}$ \\
NS & 16.7 & 17.4 & 17.9 & 26.2 & 0 & 1.45 & $-0.080$ & $\infty$ & $\infty$ & $5.0\times10^{17}$ & $8.2\times10^{19}$ \\
LDC & 1.28 & 1.39 & 1.72 & 2.54 & 0 & 1.45 & $-0.104$ & $\infty$ & $\infty$ & $\infty$ & $2.9\times10^{14}$ \\
HX & 0.526 & 0.587 & 0.599 & 1.27 & 0 & 2.12 & $+0.010$ & $\infty$ & $9.0\times10^{15}$ & $2.0\times10^{10}$ & $7.0\times10^{10}$ \\
\bottomrule
\end{tabular}

%% file: tables/tabS_v3_e2.tex
\begin{tabular}{llrrrrrrrrrrl}
\toprule
 & & & & & & & & \multicolumn{3}{c}{$R_{\rm match}$} & & \\
\cmidrule(lr){9-11}
Model & Arm, seed & nominal & MC mean & max$_{200}$ & max$_{1000}$ & searched & $R_{200}$ & uniform & corners & mixed & nom./MC mean & searched device \\
\midrule
HX & Random-error-aware s0 & 0.262 & 1.01 & 1.47 & 1.57 & 2.58 & 1.75 & 1.48 & 1.06 & 1.09 & 0.26 & 0/0/0 \\
 & Random-error-aware s1 & 0.263 & 1.04 & 1.49 & 1.64 & 2.68 & 1.80 & 1.56 & 1.09 & 1.09 & 0.25 & 0/0/0 \\
 & Random-error-aware s2 & 0.245 & 1.15 & 1.59 & 1.77 & 2.74 & 1.73 & 1.42 & 1.05 & 1.05 & 0.21 & 0/0/0 \\
 & Box s0 & 0.510 & 0.332 & 0.493 & 0.634 & 1.11 & 2.25 & 1.68 & 1.44 & 1.44 & 1.54 & 4/4/8 \\
 & Box s1 & 0.558 & 0.352 & 0.515 & 0.671 & 1.15 & 2.23 & 1.60 & 1.33 & 1.34 & 1.58 & 4/4/6 \\
 & Box s2 & 0.500 & 0.330 & 0.462 & 0.601 & 1.12 & 2.42 & 1.73 & 1.24 & 1.25 & 1.51 & 4/4/8 \\
 & Mixed-PGD FT s0 & 0.347 & 0.310 & 0.400 & 0.441 & 0.636 & 1.59 & 1.44 & 1.41 & 1.39 & 1.12 & 4/4/8 \\
 & Mixed-PGD FT s1 & 0.354 & 0.319 & 0.392 & 0.447 & 0.635 & 1.62 & 1.42 & 1.22 & 1.23 & 1.11 & 4/4/6 \\
 & Mixed-PGD FT s2 & 0.339 & 0.311 & 0.385 & 0.436 & 0.629 & 1.64 & 1.44 & 1.34 & 1.34 & 1.09 & 4/4/7 \\
\addlinespace
Darcy & Random-error-aware s0 & 1.87 & 2.47 & 3.28 & 3.32 & 5.40 & 1.65 & 1.51 & 1.01 & 1.01 & 0.76 & 0/0/0 \\
 & Random-error-aware s1 & 2.01 & 2.63 & 3.27 & 3.29 & 4.62 & 1.42 & 1.29 & 1.05 & 1.05 & 0.77 & 0/0/0 \\
 & Random-error-aware s2 & 1.90 & 2.46 & 2.84 & 3.03 & 4.50 & 1.58 & 1.43 & 1.07 & 1.07 & 0.77 & 2/4/3 \\
 & Box s0 & 2.02 & 1.88 & 2.04 & 2.10 & 2.96 & 1.45 & 1.35 & 1.04 & 1.04 & 1.08 & 0/0/0 \\
 & Box s1 & 2.01 & 1.95 & 2.14 & 2.23 & 3.05 & 1.42 & 1.33 & 1.06 & 1.06 & 1.03 & 0/0/0 \\
 & Box s2 & 2.04 & 1.93 & 2.14 & 2.29 & 3.23 & 1.51 & 1.35 & 1.19 & 1.19 & 1.06 & 4/4/7 \\
 & Mixed-PGD FT s0 & 1.93 & 1.82 & 1.94 & 1.94 & 2.59 & 1.34 & 1.29 & 1.02 & 1.05 & 1.06 & 0/0/0 \\
 & Mixed-PGD FT s1 & 1.96 & 1.89 & 2.04 & 2.08 & 2.72 & 1.33 & 1.27 & 1.06 & 1.06 & 1.04 & 0/0/0 \\
 & Mixed-PGD FT s2 & 1.95 & 1.88 & 2.02 & 2.09 & 2.76 & 1.36 & 1.34 & 1.11 & 1.10 & 1.04 & 4/4/8 \\
\addlinespace
Darcy, sinc & Random-error-aware s0 & 1.87 & 2.14 & 2.67 & 2.73 & 4.53 & 1.69 & 1.46 & 1.02 & 1.02 & 0.87 & 0/0/0 \\
 & Random-error-aware s1 & 2.01 & 2.28 & 2.68 & 2.69 & 3.94 & 1.47 & 1.34 & 1.06 & 1.06 & 0.88 & 0/0/0 \\
 & Random-error-aware s2 & 1.90 & 2.28 & 2.78 & 2.90 & 4.55 & 1.64 & 1.45 & 1.18 & 1.18 & 0.83 & 3/4/0 \\
 & Box s0 & 1.86 & 1.92 & 2.03 & 2.10 & 2.86 & 1.40 & 1.30 & 1.04 & 1.04 & 0.97 & 0/0/0 \\
 & Box s1 & 1.95 & 1.99 & 2.17 & 2.25 & 2.90 & 1.33 & 1.26 & 1.03 & 1.03 & 0.98 & 0/0/0 \\
 & Box s2 & 1.83 & 1.92 & 2.11 & 2.20 & 2.82 & 1.34 & 1.24 & 1.06 & 1.06 & 0.95 & 4/4/4 \\
 & Mixed-PGD FT s0 & 1.82 & 1.86 & 1.96 & 1.96 & 2.54 & 1.29 & 1.23 & 1.02 & 1.03 & 0.98 & 0/0/0 \\
 & Mixed-PGD FT s1 & 1.92 & 1.94 & 2.05 & 2.12 & 2.64 & 1.29 & 1.24 & 1.05 & 1.05 & 0.99 & 0/0/0 \\
 & Mixed-PGD FT s2 & 1.79 & 1.88 & 2.03 & 2.05 & 2.54 & 1.25 & 1.22 & 1.02 & 1.02 & 0.95 & 0/0/0 \\
\addlinespace
Transfer & Darcy Box s0 & 5.97 & 5.11 & 5.67 & 5.85 & 6.68 & 1.18 & 1.12 & 1.26 & 1.12 & 1.17 & 4/4/8 \\
 & Darcy Mixed-PGD FT s0 & 4.07 & 3.55 & 3.89 & 3.98 & 4.64 & 1.19 & 1.14 & 1.29 & 1.14 & 1.15 & 4/4/8 \\
 & HX Box s0 & 1.17 & 0.878 & 1.05 & 1.10 & 1.55 & 1.48 & 1.41 & 1.63 & 1.41 & 1.33 & 4/4/8 \\
 & HX Mixed-PGD FT s0 & 0.934 & 0.715 & 0.851 & 0.875 & 1.26 & 1.49 & 1.42 & 1.62 & 1.45 & 1.30 & 4/4/8 \\
\bottomrule
\end{tabular}

%% file: tables/tabS_shared.tex
\begin{tabular}{llrrrrrrrrl}
\toprule
Task & Arm (s0) & nominal & MC mean & max$_{200}$ & max$_{1000}$ & all 256 corners & searched & $R_{200}$ & $R_{1000}$ & method \\
\midrule
Darcy & Random-error-aware & 1.58 & 5.05 & 8.55 & 8.77 & 9.85 & 9.85 & 1.15 & 1.12 & K \\
Darcy & Box & 2.46 & 2.38 & 4.30 & 4.62 & 4.75 & 5.88 & 1.37 & 1.27 & Pd \\
Darcy & Smooth-mask box & 2.85 & 3.01 & 4.65 & 5.10 & 5.18 & 6.50 & 1.40 & 1.27 & Pd \\
Darcy & Box FT & 2.27 & 2.25 & 3.88 & 4.13 & 4.65 & 5.25 & 1.35 & 1.27 & Pm \\
Darcy & Mixed-PGD FT & 2.00 & 2.16 & 3.16 & 3.26 & 3.95 & 4.12 & 1.31 & 1.26 & Pd \\
Darcy & SAM FT & 2.15 & 2.43 & 3.92 & 4.07 & 4.83 & 4.94 & 1.26 & 1.22 & Pd \\
NS & Random-error-aware & 12.3 & 24.5 & 40.0 & 40.9 & 44.5 & 44.5 & 1.11 & 1.09 & K \\
NS & Box & 14.7 & 14.0 & 19.9 & 21.1 & 22.4 & 26.3 & 1.32 & 1.24 & C0 \\
NS & Smooth-mask box & 19.2 & 19.0 & 26.4 & 29.0 & 27.7 & 34.8 & 1.32 & 1.20 & Pd \\
NS & Box FT & 13.7 & 13.2 & 18.6 & 19.6 & 22.0 & 25.2 & 1.36 & 1.28 & Pd \\
NS & Mixed-PGD FT & 13.5 & 13.3 & 17.2 & 17.7 & 19.2 & 22.0 & 1.28 & 1.25 & Pd \\
NS & SAM FT & 14.3 & 13.9 & 19.2 & 20.0 & 22.3 & 24.8 & 1.30 & 1.24 & C0 \\
LDC & Random-error-aware & 1.08 & 13.8 & 28.6 & 29.1 & 30.7 & 30.7 & 1.07 & 1.06 & K \\
LDC & Box & 0.820 & 0.847 & 1.33 & 1.34 & 1.53 & 1.53 & 1.15 & 1.13 & Pa \\
LDC & Smooth-mask box & 0.724 & 0.800 & 1.53 & 1.69 & 2.21 & 2.21 & 1.44 & 1.31 & K \\
LDC & Box FT & 0.750 & 0.731 & 0.982 & 0.996 & 1.18 & 1.18 & 1.20 & 1.18 & K \\
LDC & Mixed-PGD FT & 0.725 & 0.722 & 0.792 & 0.798 & 0.892 & 0.892 & 1.13 & 1.12 & K \\
LDC & SAM FT & 1.37 & 1.56 & 2.03 & 2.04 & 2.20 & 2.25 & 1.11 & 1.10 & C1 \\
\bottomrule
\end{tabular}

%% file: tables/tabS_sweep.tex
\begin{tabular}{llrrrrrrl}
\toprule
Task & Arm (s0) & $\kappa$ & MC mean & max$_{200}$ & corners & unseen & searched & $R_{200}$ \\
\midrule
Darcy & Box & 0.25 & 2.11 & 2.56 & 2.20 & 2.26 & 3.17 & 1.24 \\
 &  & 0.5 & 1.90 & 2.66 & 2.10 & 2.26 & 3.95 & 1.48 \\
 &  & 1 & 2.01 & 2.90 & 4.27 & 4.88 & 5.77 & 1.99 \\
 &  & 1.5 & 2.65 & 4.31 & 6.75 & 7.52 & 7.52 & 1.75 \\
 &  & 2 & 3.58 & 5.98 & 9.04 & 9.91 & 9.91 & 1.66 \\
\addlinespace
Darcy & Mixed-PGD FT & 0.25 & 1.84 & 2.05 & 1.89 & 1.90 & 2.38 & 1.16 \\
 &  & 0.5 & 1.78 & 2.12 & 1.98 & 2.06 & 2.84 & 1.34 \\
 &  & 1 & 1.95 & 2.34 & 3.54 & 3.93 & 4.05 & 1.73 \\
 &  & 1.5 & 2.45 & 3.54 & 5.81 & 6.52 & 6.52 & 1.84 \\
 &  & 2 & 3.23 & 5.01 & 8.17 & 8.95 & 8.95 & 1.79 \\
\addlinespace
Darcy & Smooth-mask box & 0.25 & 2.59 & 2.92 & 2.63 & 2.76 & 3.55 & 1.21 \\
 &  & 0.5 & 2.50 & 3.01 & 2.95 & 3.12 & 4.41 & 1.47 \\
 &  & 1 & 2.75 & 3.45 & 4.81 & 5.29 & 6.64 & 1.92 \\
 &  & 1.5 & 3.36 & 4.62 & 6.76 & 7.29 & 9.28 & 2.01 \\
 &  & 2 & 4.17 & 6.03 & 8.55 & 9.60 & 11.9 & 1.97 \\
\addlinespace
NS & Box & 0.25 & 13.7 & 14.8 & 13.9 & 14.2 & 16.7 & 1.13 \\
 &  & 0.5 & 13.1 & 14.9 & 13.3 & 13.7 & 19.3 & 1.29 \\
 &  & 1 & 13.0 & 15.4 & 20.1 & 22.3 & 26.1 & 1.70 \\
 &  & 1.5 & 14.7 & 20.1 & 28.6 & 31.5 & 35.2 & 1.75 \\
 &  & 2 & 17.8 & 26.2 & 39.4 & 48.6 & 49.2 & 1.88 \\
\addlinespace
NS & Mixed-PGD FT & 0.25 & 13.0 & 13.6 & 13.1 & 13.3 & 14.7 & 1.08 \\
 &  & 0.5 & 12.7 & 13.8 & 12.9 & 13.3 & 16.6 & 1.20 \\
 &  & 1 & 12.8 & 14.4 & 17.6 & 19.1 & 21.8 & 1.52 \\
 &  & 1.5 & 14.0 & 17.5 & 25.8 & 28.0 & 29.0 & 1.66 \\
 &  & 2 & 16.6 & 23.2 & 41.2 & 54.4 & 54.4 & 2.35 \\
\addlinespace
NS & Smooth-mask box & 0.25 & 18.2 & 19.4 & 18.5 & 19.1 & 21.9 & 1.13 \\
 &  & 0.5 & 17.7 & 19.7 & 18.2 & 19.5 & 25.7 & 1.30 \\
 &  & 1 & 18.2 & 20.5 & 25.6 & 28.4 & 35.0 & 1.71 \\
 &  & 1.5 & 20.1 & 25.0 & 33.6 & 38.3 & 46.1 & 1.85 \\
 &  & 2 & 22.9 & 30.3 & 40.8 & 48.0 & 57.4 & 1.90 \\
\addlinespace
LDC & Box & 0.25 & 0.823 & 0.883 & 0.873 & 0.936 & 1.00 & 1.14 \\
 &  & 0.5 & 0.830 & 0.957 & 1.00 & 1.21 & 1.36 & 1.43 \\
 &  & 1 & 0.860 & 1.17 & 1.88 & 2.55 & 2.65 & 2.26 \\
 &  & 1.5 & 0.929 & 1.73 & 4.30 & 4.94 & 4.95 & 2.86 \\
 &  & 2 & 1.07 & 2.76 & 7.91 & 8.12 & 8.12 & 2.94 \\
\addlinespace
LDC & Mixed-PGD FT & 0.25 & 0.723 & 0.755 & 0.755 & 0.771 & 0.784 & 1.04 \\
 &  & 0.5 & 0.723 & 0.786 & 0.799 & 0.826 & 0.847 & 1.08 \\
 &  & 1 & 0.726 & 0.842 & 0.997 & 1.18 & 1.20 & 1.43 \\
 &  & 1.5 & 0.735 & 0.920 & 1.45 & 1.80 & 1.80 & 1.96 \\
 &  & 2 & 0.754 & 1.10 & 2.38 & 2.92 & 2.93 & 2.67 \\
\addlinespace
LDC & Smooth-mask box & 0.25 & 0.674 & 0.813 & 0.820 & 0.920 & 1.07 & 1.31 \\
 &  & 0.5 & 0.661 & 0.901 & 1.20 & 1.32 & 1.51 & 1.68 \\
 &  & 1 & 0.728 & 1.28 & 2.33 & 2.90 & 3.04 & 2.36 \\
 &  & 1.5 & 0.900 & 1.92 & 3.53 & 4.82 & 5.17 & 2.69 \\
 &  & 2 & 1.15 & 2.62 & 5.18 & 7.08 & 7.08 & 2.70 \\
\addlinespace
\bottomrule
\end{tabular}

%% file: tables/tabS_cp.tex
\begin{tabular}{llrrrrrrrl}
\toprule
Task & Arm, seed & nominal & MC mean & max$_{200}$ & max$_{1000}$ & unseen & searched & $R_{200}$ & method \\
\midrule
Darcy & Random-error-aware s0 & 1.86 & 7.07 & 9.77 & 10.8 & 14.1 & 14.7 & 1.50 & Pa \\
Darcy & Random-error-aware s1 & 1.91 & 5.51 & 7.27 & 7.92 & 10.6 & 10.6 & 1.45 & Pa \\
Darcy & Random-error-aware s2 & 1.86 & 6.05 & 8.38 & 8.93 & 12.7 & 12.9 & 1.54 & Pa \\
Darcy & Box s0 & 3.16 & 2.05 & 2.45 & 2.82 & 4.36 & 4.38 & 1.78 & Pa \\
Darcy & Box s1 & 3.50 & 2.08 & 2.67 & 3.02 & 4.34 & 4.38 & 1.64 & Pa \\
Darcy & Box s2 & 3.75 & 2.15 & 3.20 & 3.21 & 4.60 & 5.32 & 1.66 & Pc \\
Darcy & Mixed-PGD FT s0 & 2.80 & 1.99 & 2.35 & 2.55 & 3.69 & 3.70 & 1.57 & Pa \\
Darcy & Mixed-PGD FT s1 & 3.00 & 2.00 & 2.47 & 2.67 & 3.78 & 3.81 & 1.54 & Pa \\
Darcy & Mixed-PGD FT s2 & 3.36 & 2.07 & 2.72 & 2.78 & 3.93 & 4.23 & 1.56 & Pc \\
NS & Random-error-aware s0 & 10.7 & 31.4 & 39.1 & 42.9 & 55.7 & 55.7 & 1.42 & A \\
NS & Random-error-aware s1 & 13.7 & 34.6 & 45.0 & 46.7 & 61.1 & 61.1 & 1.36 & A \\
NS & Random-error-aware s2 & 13.4 & 34.0 & 42.2 & 46.2 & 61.0 & 61.0 & 1.44 & A \\
NS & Box s0 & 17.2 & 12.8 & 15.4 & 16.1 & 20.4 & 23.0 & 1.50 & Pc \\
NS & Box s1 & 18.2 & 13.0 & 15.5 & 15.9 & 21.8 & 23.6 & 1.52 & Pc \\
NS & Box s2 & 17.6 & 13.4 & 15.2 & 15.8 & 20.9 & 22.6 & 1.48 & Pc \\
NS & Mixed-PGD FT s0 & 15.6 & 12.7 & 14.6 & 15.1 & 17.2 & 19.0 & 1.30 & Pc \\
NS & Mixed-PGD FT s1 & 16.7 & 12.9 & 14.7 & 14.9 & 18.5 & 19.9 & 1.36 & Pc \\
NS & Mixed-PGD FT s2 & 16.7 & 12.9 & 14.5 & 14.6 & 18.7 & 19.8 & 1.37 & Pd \\
LDC & Random-error-aware s0 & 1.16 & 25.0 & 35.1 & 37.1 & 47.9 & 47.9 & 1.37 & Pa \\
LDC & Random-error-aware s1 & 1.01 & 22.1 & 35.0 & 35.0 & 43.4 & 43.4 & 1.24 & A \\
LDC & Random-error-aware s2 & 0.988 & 17.7 & 26.9 & 31.6 & 37.3 & 37.4 & 1.39 & Pa \\
LDC & Box s0 & 0.727 & 0.844 & 1.26 & 1.43 & 3.09 & 3.13 & 2.48 & Pd \\
LDC & Box s1 & 0.766 & 0.887 & 1.37 & 1.52 & 2.29 & 2.57 & 1.88 & Pk \\
LDC & Box s2 & 0.661 & 0.829 & 1.24 & 1.43 & 3.73 & 3.76 & 3.03 & Pd \\
LDC & Mixed-PGD FT s0 & 0.629 & 0.608 & 0.689 & 0.718 & 0.899 & 0.984 & 1.43 & Pd \\
LDC & Mixed-PGD FT s1 & 0.616 & 0.618 & 0.689 & 0.704 & 0.801 & 0.834 & 1.21 & Pd \\
LDC & Mixed-PGD FT s2 & 0.569 & 0.623 & 0.758 & 0.791 & 1.64 & 1.67 & 2.21 & Pa \\
\bottomrule
\end{tabular}

%% file: tables/tabS_paired.tex
\begin{tabular}{llrrrrrr}
\toprule
 & & \multicolumn{3}{c}{searched (\%)} & \multicolumn{3}{c}{unseen attacks (\%)} \\
\cmidrule(lr){3-5}\cmidrule(lr){6-8}
Task & seed & Box FT & Mixed-PGD FT & change & Box FT & Mixed-PGD FT & change \\
\midrule
Darcy & s0 & 5.17 & 4.05 & $-$22\% & 4.76 & 3.93 & $-$17\% \\
 & s1 & 5.95 & 4.43 & $-$26\% & 5.33 & 4.07 & $-$24\% \\
 & s2 & 5.37 & 4.43 & $-$18\% & 4.75 & 4.05 & $-$15\% \\
 & s3 & 5.53 & 4.55 & $-$18\% & 4.71 & 4.03 & $-$15\% \\
 & s4 & 5.82 & 4.29 & $-$26\% & 4.67 & 4.03 & $-$14\% \\
\addlinespace
NS & s0 & 24.6 & 21.8 & $-$11\% & 22.5 & 19.9 & $-$12\% \\
 & s1 & 22.7 & 21.7 & $-$4\% & 21.5 & 19.1 & $-$11\% \\
 & s2 & 26.9 & 22.8 & $-$15\% & 25.0 & 21.0 & $-$16\% \\
 & s3 & 25.4 & 21.7 & $-$15\% & 21.2 & 18.4 & $-$13\% \\
 & s4 & 26.0 & 21.5 & $-$17\% & 22.9 & 19.6 & $-$15\% \\
\addlinespace
LDC & s0 & 1.99 & 1.22 & $-$39\% & 1.85 & 1.18 & $-$36\% \\
 & s1 & 0.908 & 0.541 & $-$40\% & 0.887 & 0.531 & $-$40\% \\
 & s2 & 2.80 & 1.53 & $-$45\% & 2.22 & 1.26 & $-$43\% \\
 & s3 & 3.52 & 2.41 & $-$32\% & 3.35 & 2.25 & $-$33\% \\
 & s4 & 2.64 & 0.990 & $-$62\% & 2.36 & 0.872 & $-$63\% \\
\addlinespace
HX & s0 & 1.18 & 0.979 & $-$17\% & 0.975 & 0.783 & $-$20\% \\
 & s1 & 1.10 & 0.893 & $-$19\% & 0.851 & 0.727 & $-$15\% \\
 & s2 & 1.12 & 0.942 & $-$16\% & 0.729 & 0.694 & $-$5\% \\
 & s3 & 1.21 & 1.05 & $-$13\% & 0.934 & 0.913 & $-$2\% \\
 & s4 & 1.06 & 0.895 & $-$16\% & 0.833 & 0.664 & $-$20\% \\
\bottomrule
\end{tabular}

%% file: tables/tabS_hx.tex
\begin{tabular}{llrrrrr}
\toprule
Arm & Device & primary (mean of channels) & $p$ & $u_1$ & $u_2$ & joint median \\
\midrule
Random-error-aware & nominal & 0.243 & 0.314 & 0.209 & 0.206 & 0.275 \\
 & MC (dev-selected) & 3.94 & 5.65 & 3.14 & 3.04 & 5.66 \\
 & searched & 4.82 & 6.87 & 3.87 & 3.72 & 6.88 \\
\addlinespace
Box & nominal & 0.351 & 0.449 & 0.305 & 0.299 & 0.406 \\
 & MC (dev-selected) & 0.501 & 0.636 & 0.432 & 0.436 & 0.576 \\
 & searched & 1.18 & 1.46 & 1.06 & 1.01 & 1.31 \\
\addlinespace
Smooth-mask box & nominal & 0.523 & 0.687 & 0.439 & 0.442 & 0.564 \\
 & MC (dev-selected) & 0.679 & 0.908 & 0.563 & 0.567 & 0.816 \\
 & searched & 1.22 & 1.66 & 0.999 & 0.999 & 1.50 \\
\addlinespace
Box FT & nominal & 0.329 & 0.424 & 0.286 & 0.277 & 0.378 \\
 & MC (dev-selected) & 0.475 & 0.604 & 0.410 & 0.412 & 0.546 \\
 & searched & 1.13 & 1.39 & 1.05 & 0.961 & 1.24 \\
\addlinespace
Mixed-PGD FT & nominal & 0.327 & 0.420 & 0.283 & 0.277 & 0.378 \\
 & MC (dev-selected) & 0.462 & 0.608 & 0.390 & 0.389 & 0.569 \\
 & searched & 0.952 & 1.23 & 0.840 & 0.783 & 1.10 \\
\addlinespace
SAM FT & nominal & 0.437 & 0.533 & 0.393 & 0.386 & 0.497 \\
 & MC (dev-selected) & 0.606 & 0.751 & 0.540 & 0.528 & 0.713 \\
 & searched & 1.38 & 1.58 & 1.39 & 1.16 & 1.46 \\
\addlinespace
Digital FNO & ideal hardware & 0.162 & 0.236 & 0.134 & 0.117 & 0.211 \\
\addlinespace
\bottomrule
\end{tabular}

%% file: tables/tabS_cert.tex
\begin{tabular}{lllrrrrr}
\toprule
Model & Knobs & $\kappa$ & expansion-point error (\%) & worst found (\%) & bound (\%) & bound / worst & float slack \\
\midrule
Box & 32 & 0.001 & 2.396 &2.400 & 2.416 & 1.01 & -- \\
Box & 32 & 0.003 & 2.393 &2.405 & 2.448 & 1.02 & -- \\
Box & 32 & 0.005 & 2.391 &2.410 & 2.578 & 1.07 & -- \\
Box & 32 & 0.007 & 2.388 &2.416 & 2.978 & 1.23 & -- \\
Box & 32 & 0.01 & 2.385 &2.424 & 4.915 & 2.03 & -- \\
Box & 32 & 0.03 & 2.360 &2.478 & 315.7 & 127 & -- \\
Box & 32 & 0.1 & 2.278 &2.679 & $1.31\times10^{4}$ & $4.88\times10^{3}$ & -- \\
Box & 32 & 1 & 1.710 &6.719 & $4.05\times10^{9}$ & $6.02\times10^{8}$ & -- \\
Box & 28 (ext.\ fixed) & 0.001 & 2.396 &2.399 & 2.410 & 1.00 & -- \\
Box & 28 (ext.\ fixed) & 0.003 & 2.393 &2.404 & 2.429 & 1.01 & -- \\
Box & 28 (ext.\ fixed) & 0.005 & 2.391 &2.409 & 2.518 & 1.05 & -- \\
Box & 28 (ext.\ fixed) & 0.007 & 2.388 &2.414 & 2.797 & 1.16 & -- \\
Box & 28 (ext.\ fixed) & 0.01 & 2.385 &2.421 & 4.143 & 1.71 & -- \\
Box & 28 (ext.\ fixed) & 0.03 & 2.360 &2.469 & 255.3 & 103 & -- \\
Box & 28 (ext.\ fixed) & 0.1 & 2.278 &2.646 & $1.03\times10^{4}$ & $3.90\times10^{3}$ & -- \\
Box & 28 (ext.\ fixed) & 1 & 1.668 &6.128 & $2.55\times10^{9}$ & $4.16\times10^{8}$ & -- \\
Mixed-PGD FT & 32 & 0.001 & 2.039 &2.042 & 2.059 & 1.01 & -- \\
Mixed-PGD FT & 32 & 0.003 & 2.038 &2.046 & 2.092 & 1.02 & -- \\
Mixed-PGD FT & 32 & 0.005 & 2.036 &2.050 & 2.238 & 1.09 & -- \\
Mixed-PGD FT & 32 & 0.007 & 2.035 &2.054 & 2.705 & 1.32 & -- \\
Mixed-PGD FT & 32 & 0.01 & 2.033 &2.061 & 5.064 & 2.46 & -- \\
Mixed-PGD FT & 32 & 0.03 & 2.018 &2.102 & 336.1 & 160 & -- \\
Mixed-PGD FT & 32 & 0.1 & 1.969 &2.261 & $1.38\times10^{4}$ & $6.12\times10^{3}$ & -- \\
Mixed-PGD FT & 32 & 1 & 1.789 &5.340 & $4.53\times10^{9}$ & $8.48\times10^{8}$ & -- \\
Mixed-PGD FT & 28 (ext.\ fixed) & 0.001 & 2.039 &2.042 & 2.052 & 1.01 & -- \\
Mixed-PGD FT & 28 (ext.\ fixed) & 0.003 & 2.038 &2.046 & 2.072 & 1.01 & -- \\
Mixed-PGD FT & 28 (ext.\ fixed) & 0.005 & 2.036 &2.049 & 2.170 & 1.06 & -- \\
Mixed-PGD FT & 28 (ext.\ fixed) & 0.007 & 2.035 &2.053 & 2.493 & 1.21 & -- \\
Mixed-PGD FT & 28 (ext.\ fixed) & 0.01 & 2.033 &2.058 & 4.114 & 2.00 & -- \\
Mixed-PGD FT & 28 (ext.\ fixed) & 0.03 & 2.018 &2.096 & 271.7 & 130 & -- \\
Mixed-PGD FT & 28 (ext.\ fixed) & 0.1 & 1.968 &2.238 & $1.09\times10^{4}$ & $4.88\times10^{3}$ & -- \\
Mixed-PGD FT & 28 (ext.\ fixed) & 1 & 1.755 &4.930 & $2.84\times10^{9}$ & $5.76\times10^{8}$ & -- \\
Box & 32 & 0.001 & 2.396 &2.400 & 4.122 & 1.72 & $10^{-5}$ \\
Box & 32 & 0.003 & 2.393 &2.405 & 4.574 & 1.90 & $10^{-5}$ \\
Box & 32 & 0.005 & 2.391 &2.410 & 5.555 & 2.30 & $10^{-5}$ \\
Box & 32 & 0.01 & 2.385 &2.424 & 13.84 & 5.71 & $10^{-5}$ \\
Box & 28 (ext.\ fixed) & 0.001 & 2.396 &2.399 & 4.012 & 1.67 & $10^{-5}$ \\
Box & 28 (ext.\ fixed) & 0.003 & 2.393 &2.404 & 4.393 & 1.83 & $10^{-5}$ \\
Box & 28 (ext.\ fixed) & 0.005 & 2.391 &2.409 & 5.195 & 2.16 & $10^{-5}$ \\
Box & 28 (ext.\ fixed) & 0.01 & 2.385 &2.421 & 11.52 & 4.76 & $10^{-5}$ \\
Mixed-PGD FT & 32 & 0.001 & 2.039 &2.042 & 4.129 & 2.02 & $10^{-5}$ \\
Mixed-PGD FT & 32 & 0.003 & 2.038 &2.046 & 4.684 & 2.29 & $10^{-5}$ \\
Mixed-PGD FT & 32 & 0.005 & 2.036 &2.050 & 5.911 & 2.88 & $10^{-5}$ \\
Mixed-PGD FT & 32 & 0.01 & 2.033 &2.061 & 16.69 & 8.10 & $10^{-5}$ \\
Mixed-PGD FT & 28 (ext.\ fixed) & 0.001 & 2.039 &2.042 & 3.993 & 1.96 & $10^{-5}$ \\
Mixed-PGD FT & 28 (ext.\ fixed) & 0.003 & 2.038 &2.046 & 4.458 & 2.18 & $10^{-5}$ \\
Mixed-PGD FT & 28 (ext.\ fixed) & 0.005 & 2.036 &2.049 & 5.458 & 2.66 & $10^{-5}$ \\
Mixed-PGD FT & 28 (ext.\ fixed) & 0.01 & 2.033 &2.058 & 13.63 & 6.62 & $10^{-5}$ \\
\bottomrule
\end{tabular}

%% file: tables/tabS_decomp.tex
\begin{tabular}{lrrrr}
\toprule
Effect alone & passive s0 & passive s1 & ideal (calibrated) & mean ratio \\
\midrule
none (ideal passive) & 1.64 & 1.73 & 2.04 & 1.00 \\
input DAC, 8 bits & 1.65 & 1.73 & 2.05 & 1.00 \\
mask phase, 8 bits & 1.65 & 1.73 & 2.10 & 1.01 \\
mask amplitude, 8 bits & 1.65 & 1.75 & 2.05 & 1.00 \\
output ADC, 10 bits & 1.64 & 1.73 & 2.04 & 1.00 \\
extinction ratio, 30\,dB & 8.78 & 8.03 & 12.04 & 5.29 \\
static phase error, $\sigma=0.05$\,rad & 1.91 & 1.97 & 3.64 & 1.36 \\
static amplitude error, $\sigma=0.02$ & 1.69 & 1.79 & 2.30 & 1.06 \\
crosstalk width, 0.5\,pixel & 16.49 & 17.52 & 18.69 & 9.76 \\
path piston, $\sigma=0.05$\,rad & 1.88 & 1.97 & 3.51 & 1.33 \\
shot noise, $10^4$ photons & 1.66 & 1.74 & 2.07 & 1.01 \\
read noise, $10^{-3}$ of full scale & 1.65 & 1.73 & 2.05 & 1.00 \\
registration shift, (0.3, 0)\,pixel & 7.14 & 6.97 & 17.46 & 5.64 \\
registration shift, (0.3, 0.3)\,pixel & 9.82 & 8.38 & 20.09 & 6.88 \\
defocus, 0.2\,rad & 1.68 & 1.77 & 2.13 & 1.03 \\
tilt, (0.2, 0.2)\,rad & 1.93 & 1.99 & 2.64 & 1.20 \\
reference phase, 0.1\,rad & 2.32 & 2.38 & 4.11 & 1.60 \\
all (v1 \textsc{realistic}) & 16.91 & 17.73 & 18.65 & 9.89 \\
\bottomrule
\end{tabular}

%% file: tables/tabS_portability.tex
\begin{tabular}{lrrrrrr}
\toprule
Setting & \multicolumn{3}{c}{Frozen learned library (\%)} & \multicolumn{3}{c}{Frozen random optics (\%)} \\
\cmidrule(lr){2-4}\cmidrule(lr){5-7}
 & calibration chip & other chips & penalty & calibration chip & other chips & penalty \\
\midrule
main ($N_\mathrm{train}=1000$) & 20.35 & 21.50 & 1.16 & 24.12 & 25.88 & 1.76 \\
pre-mix adapter & 16.57 & 18.63 & 2.06 & 19.60 & 22.03 & 2.43 \\
$N_\mathrm{train}=250$ & 25.07 & 25.81 & 0.74 & 27.48 & 29.99 & 2.51 \\
$N_\mathrm{train}=100$ & 27.83 & 28.46 & 0.64 & 30.17 & 32.62 & 2.46 \\
slim adapter & 25.51 & 28.12 & 2.62 & 24.15 & 26.25 & 2.10 \\
16 blocks & 20.41 & 22.07 & 1.66 & 21.44 & 22.53 & 1.09 \\
phase-only & 20.33 & 26.99 & 6.67 & 18.88 & 29.44 & 10.56 \\
intensity detection & 57.08 & 56.67 & $-$0.42 & 32.14 & 32.76 & 0.63 \\
advection–diffusion ($N_\mathrm{train}=1000$) & 26.91 & 28.72 & 1.81 & 32.42 & 35.10 & 2.68 \\
library, adapter trained with device randomisation & 22.07 & 22.04 & $-$0.03 & -- & -- & -- \\
\bottomrule
\end{tabular}

%% file: tables/tabS_ldc_fields.tex
\begin{tabular}{l ccc ccc ccc}
\toprule
 & \multicolumn{3}{c}{pressure $p$} & \multicolumn{3}{c}{TKE $k$} & \multicolumn{3}{c}{speed $|\mathbf{u}|$} \\
\cmidrule(lr){2-4}\cmidrule(lr){5-7}\cmidrule(lr){8-10}
Arm & nom. & MC & search & nom. & MC & search & nom. & MC & search \\
\midrule
Random-error-aware & 1.08 & 21.0 & 30.1 & 10.7 & 31.4 & 43.4 & 5.97 & 16.9 & 24.1 \\
Box & 0.92 & 1.38 & 3.04 & 14.6 & 15.0 & 19.3 & 8.85 & 7.92 & 12.6 \\
Smooth-mask box & 1.08 & 1.74 & 3.56 & 12.3 & 12.7 & 15.6 & 6.77 & 8.09 & 8.14 \\
Box FT & 0.81 & 1.13 & 2.37 & 15.2 & 15.6 & 18.4 & 8.70 & 9.07 & 11.6 \\
Mixed-PGD FT & 0.76 & 0.92 & 1.34 & 15.1 & 15.6 & 15.9 & 8.76 & 8.97 & 10.8 \\
SAM FT & 1.52 & 2.38 & 3.96 & 20.4 & 23.4 & 23.5 & 12.1 & 11.9 & 13.8 \\
Digital & 0.63 & -- & -- & 13.5 & -- & -- & 8.20 & -- & -- \\
\bottomrule
\end{tabular}

%% file: tables/tabS_hybrid.tex
\begin{tabular}{llrrrrrrrrrrl}
\toprule
 & & & & & & & & \multicolumn{2}{c}{shared at $z_{\rm search}$} & \multicolumn{2}{c}{$R_{200}$, other boxes} & \\
\cmidrule(lr){9-10}\cmidrule(lr){11-12}
Task & Arm (s0) & nominal & max$_{200}$ & max$_{1000}$ & searched & $R_{200}$ & $R_{1000}$ & crosstalk & extinction & 32-D & 8-D & method \\
\midrule
Darcy & Random-error-aware & 1.58 & 8.67 & 8.67 & 10.0 & 1.16 & 1.16 & 0.6 & 33 & 1.38 & 1.15 & A \\
 & Box & 2.46 & 3.29 & 3.60 & 5.77 & 1.75 & 1.60 & 0.4 & 27 & 1.99 & 1.37 & Pc \\
 & Smooth-mask box & 2.85 & 3.95 & 4.25 & 6.68 & 1.69 & 1.57 & 0.4 & 27 & 1.92 & 1.40 & Pc \\
 & Box FT & 2.27 & 3.16 & 3.27 & 5.19 & 1.64 & 1.59 & 0.4 & 27 & 1.95 & 1.35 & Pc \\
 & Mixed-PGD FT & 2.00 & 2.73 & 2.81 & 4.06 & 1.49 & 1.44 & 0.4 & 27 & 1.73 & 1.31 & Pc \\
 & SAM FT & 2.15 & 3.44 & 3.55 & 4.92 & 1.43 & 1.39 & 0.4 & 27 & 1.67 & 1.26 & Pc \\
\addlinespace
NS & Random-error-aware & 12.3 & 38.0 & 39.8 & 45.5 & 1.19 & 1.14 & 0.6 & 33 & 1.29 & 1.11 & A \\
 & Box & 14.7 & 17.1 & 18.2 & 26.2 & 1.54 & 1.43 & 0.4 & 27 & 1.70 & 1.32 & Pc \\
 & Smooth-mask box & 19.2 & 23.2 & 24.6 & 34.9 & 1.50 & 1.42 & 0.4 & 27 & 1.71 & 1.32 & Pd \\
 & Box FT & 13.7 & 16.0 & 17.0 & 24.7 & 1.55 & 1.45 & 0.4 & 27 & 1.70 & 1.36 & Pc \\
 & Mixed-PGD FT & 13.5 & 15.2 & 16.0 & 22.0 & 1.44 & 1.37 & 0.4 & 27 & 1.52 & 1.28 & Pd \\
 & SAM FT & 14.3 & 16.5 & 17.4 & 24.8 & 1.50 & 1.43 & 0.4 & 27 & 1.65 & 1.30 & Pc \\
\addlinespace
LDC & Random-error-aware & 1.08 & 25.4 & 28.6 & 35.6 & 1.40 & 1.24 & 0.6 & 33 & 1.36 & 1.07 & A \\
 & Box & 0.820 & 1.13 & 1.30 & 2.47 & 2.18 & 1.90 & 0.4 & 27 & 2.22 & 1.15 & Pk \\
 & Smooth-mask box & 0.724 & 1.52 & 1.66 & 2.78 & 1.83 & 1.67 & 0.6 & 33 & 2.36 & 1.44 & Pk \\
 & Box FT & 0.750 & 1.01 & 1.24 & 1.92 & 1.90 & 1.56 & 0.4 & 27 & 2.08 & 1.20 & Pk \\
 & Mixed-PGD FT & 0.725 & 0.850 & 0.900 & 1.21 & 1.43 & 1.35 & 0.4 & 27 & 1.45 & 1.13 & Pd \\
 & SAM FT & 1.37 & 2.12 & 2.21 & 3.60 & 1.70 & 1.63 & 0.4 & 27 & 1.73 & 1.11 & Pa \\
\addlinespace
HX & Random-error-aware & 0.240 & 4.11 & 4.16 & 4.85 & 1.18 & 1.17 & 0.6 & 33 & 1.34 & -- & A \\
 & Box & 0.353 & 0.500 & 0.574 & 1.21 & 2.42 & 2.10 & 0.4 & 27 & 2.80 & -- & Pd \\
 & Smooth-mask box & 0.482 & 0.716 & 0.737 & 1.09 & 1.52 & 1.48 & 0.4 & 27 & 1.77 & -- & Pc \\
 & Box FT & 0.329 & 0.473 & 0.542 & 1.19 & 2.51 & 2.19 & 0.4 & 27 & 2.95 & -- & Pc \\
 & Mixed-PGD FT & 0.319 & 0.445 & 0.525 & 0.950 & 2.14 & 1.81 & 0.4 & 27 & 2.43 & -- & Pc \\
 & SAM FT & 0.449 & 0.612 & 0.663 & 1.36 & 2.22 & 2.05 & 0.4 & 27 & 2.59 & -- & Pr \\
\bottomrule
\end{tabular}

%% file: tables/tabS_hx_qoi.tex
\begin{tabular}{l rrrrr rrrrr rrrrr}
\toprule
 & \multicolumn{5}{c}{$Q_p$} & \multicolumn{5}{c}{$Q_s$} & \multicolumn{5}{c}{$Q_{\max}$} \\
\cmidrule(lr){2-6}\cmidrule(lr){7-11}\cmidrule(lr){12-16}
Arm (seeds) & nom. & MC max & search & ratio & $n_>$ & nom. & MC max & search & ratio & $n_>$ & nom. & MC max & search & ratio & $n_>$ \\
\midrule
\multicolumn{16}{l}{\emph{Main round, mode-level device}} \\
Random-error-aware (5) & 0.125 & 5.37 & 6.54 & 1.18--1.24 & 5/5 & 0.0391 & 2.95 & 3.61 & 1.21--1.24 & 5/5 & 0.160 & 3.06 & 3.66 & 1.15--1.26 & 5/5 \\
Box (5) & 0.132 & 0.367 & 0.539 & 0.46--2.21 & 3/5 & 0.0597 & 0.212 & 0.212 & 0.30--1.75 & 1/5 & 0.210 & 0.414 & 0.537 & 0.69--1.90 & 4/5 \\
Smooth-mask box (5) & 0.371 & 0.734 & 1.28 & 1.53--2.01 & 5/5 & 0.146 & 0.354 & 0.535 & 1.28--2.04 & 5/5 & 0.283 & 0.583 & 0.693 & 0.81--1.58 & 3/5 \\
Box FT (5) & 0.121 & 0.407 & 0.506 & 0.28--2.02 & 3/5 & 0.0516 & 0.228 & 0.180 & 0.27--1.51 & 1/5 & 0.222 & 0.403 & 0.486 & 0.52--1.84 & 3/5 \\
Mixed-PGD FT (5) & 0.115 & 0.328 & 0.506 & 0.73--2.75 & 4/5 & 0.0460 & 0.174 & 0.187 & 0.29--2.25 & 3/5 & 0.211 & 0.396 & 0.452 & 0.57--1.70 & 3/5 \\
SAM FT (5) & 0.142 & 0.370 & 0.618 & 0.81--2.26 & 4/5 & 0.0518 & 0.190 & 0.395 & 1.18--2.59 & 5/5 & 0.215 & 0.489 & 0.702 & 1.10--1.85 & 5/5 \\
All (30) &  &  &  & med.\ 1.45 & 24/30 &  &  &  & med.\ 1.23 & 20/30 &  &  &  & med.\ 1.22 & 23/30 \\
\addlinespace
\multicolumn{16}{l}{\emph{Pixel-level device}} \\
Random-error-aware (3) & 0.144 & 2.21 & 3.36 & 1.49--1.56 & 3/3 & 0.0546 & 1.24 & 1.88 & 1.48--1.57 & 3/3 & 0.166 & 1.27 & 1.71 & 1.23--1.40 & 3/3 \\
Box (3) & 0.604 & 0.803 & 1.46 & 1.70--2.00 & 3/3 & 0.304 & 0.374 & 0.702 & 1.74--2.03 & 3/3 & 0.333 & 0.400 & 0.801 & 1.88--2.09 & 3/3 \\
Mixed-PGD FT (3) & 0.151 & 0.453 & 0.617 & 1.22--1.47 & 3/3 & 0.0354 & 0.191 & 0.278 & 1.34--1.51 & 3/3 & 0.126 & 0.330 & 0.317 & 0.94--0.98 & 0/3 \\
All (9) &  &  &  & med.\ 1.52 & 9/9 &  &  &  & med.\ 1.51 & 9/9 &  &  &  & med.\ 1.40 & 6/9 \\
\bottomrule
\end{tabular}

%% file: tables/tabS_ldc_meanp.tex
\begin{tabular}{lrrrrrrrrrl}
\toprule
 & & \multicolumn{3}{c}{mean removed} & \multicolumn{2}{c}{$>1$} & & & \multicolumn{2}{c}{seed 0, $R_{1000}^{\rm mr}$} \\
\cmidrule(lr){3-5}\cmidrule(lr){6-7}\cmidrule(lr){10-11}
Arm (seeds) & $R_{200}$ & $R_{200}^{\rm mr}$ & $R_{1000}^{\rm mr}$ & $R_{200}^{p,\rm mr}$ & $R_{200}^{\rm mr}$ & $R_{1000}^{\rm mr}$ & offset share & $e^{\rm mr}/e$ & searched & re-searched \\
\midrule
Random-error-aware (5) & 1.40 & 1.39 & 1.35 & 1.39 & 5/5 & 5/5 & 0.94 & 0.98 & 1.35 & 1.36 (Pk) \\
Box (5) & 2.22 & 1.63 & 1.59 & 1.63 & 5/5 & 5/5 & 0.93 & 1.00 & 1.59 & 1.73 (Pm) \\
Smooth-mask box (5) & 1.89 & 1.76 & 1.65 & 1.76 & 5/5 & 5/5 & 0.92 & 1.05 & 1.80 & 1.83 (Pa) \\
Box FT (5) & 2.08 & 1.56 & 1.53 & 1.56 & 5/5 & 5/5 & 0.92 & 1.03 & 1.48 & 1.52 (Pz) \\
Mixed-PGD FT (5) & 1.44 & 1.15 & 1.15 & 1.15 & 5/5 & 5/5 & 0.84 & 1.32 & 1.15 & 1.25 (Pc) \\
SAM FT (5) & 1.61 & 1.21 & 1.16 & 1.21 & 5/5 & 4/5 & 0.84 & 1.30 & 1.17 & 1.51 (Pd) \\
\addlinespace
All (30) & 1.67 & 1.47 & 1.39 & 1.47 & 30/30 & 29/30 & 0.92 & 1.03 & & \\
Range (30) & 1.08--3.10 & 1.07--2.15 & 0.96--1.93 & 1.07--2.15 & & & 0.48--0.94 & 0.93--2.51 & & \\
\bottomrule
\end{tabular}

%% file: tables/tabS_hx_qoi_search.tex
\begin{tabular}{l rrrrrc rrrrrc}
\toprule
 & \multicolumn{6}{c}{$Q_p$} & \multicolumn{6}{c}{$Q_s$} \\
\cmidrule(lr){2-7}\cmidrule(lr){8-13}
Arm (seeds) & $R_{1000}$ & $n_{>\rm MC}$ & search$/W_1$ & $n_{>W_1}$ & $R_{\rm field}$ & 1/2/5\% & $R_{1000}$ & $n_{>\rm MC}$ & search$/W_1$ & $n_{>W_1}$ & $R_{\rm field}$ & 1/2/5\% \\
\midrule
\multicolumn{13}{l}{\emph{Main round, mode-level device}} \\
Random-error-aware (5) & 1.24 & 5/5 & 1.40 & 5/5 & 1.01 & 0/0/5 & 1.24 & 5/5 & 1.40 & 5/5 & 1.01 & 0/0/0 \\
Box (5) & 2.45 & 5/5 & 3.33 & 5/5 & 1.43 & 1/0/0 & 2.34 & 5/5 & 3.29 & 5/5 & 2.66 & 0/0/0 \\
Smooth-mask box (5) & 1.91 & 5/5 & 2.99 & 5/5 & 1.09 & 5/0/0 & 1.83 & 5/5 & 3.07 & 5/5 & 1.32 & 0/0/0 \\
Box FT (5) & 2.34 & 5/5 & 3.32 & 5/5 & 1.75 & 1/0/0 & 2.27 & 5/5 & 3.54 & 5/5 & 3.00 & 0/0/0 \\
Mixed-PGD FT (5) & 2.41 & 5/5 & 3.25 & 5/5 & 1.72 & 0/0/0 & 2.47 & 5/5 & 2.98 & 5/5 & 2.07 & 0/0/0 \\
SAM FT (5) & 2.32 & 5/5 & 3.06 & 5/5 & 1.40 & 1/0/0 & 2.76 & 5/5 & 5.01 & 5/5 & 1.31 & 0/0/0 \\
All (30) & 2.25 & 30/30 & 3.03 & 30/30 & 1.14 & 8/0/5 & 2.20 & 30/30 & 3.04 & 30/30 & 1.35 & 0/0/0 \\
\addlinespace
\multicolumn{13}{l}{\emph{Pixel-level device}} \\
Random-error-aware (3) & 1.57 & 3/3 & 1.89 & 3/3 & 1.04 & 0/3/0 & 1.55 & 3/3 & 1.83 & 3/3 & 1.01 & 0/1/0 \\
Box (3) & 1.80 & 3/3 & 3.70 & 3/3 & 1.01 & 3/0/0 & 2.02 & 3/3 & 3.77 & 3/3 & 1.03 & 0/0/0 \\
Mixed-PGD FT (3) & 1.57 & 3/3 & 2.18 & 3/3 & 1.15 & 0/0/0 & 1.60 & 3/3 & 1.86 & 3/3 & 1.07 & 0/0/0 \\
All (9) & 1.58 & 9/9 & 2.18 & 9/9 & 1.03 & 3/3/0 & 1.60 & 9/9 & 1.86 & 9/9 & 1.03 & 0/1/0 \\
\bottomrule
\end{tabular}